\documentclass[10pt,twocolumn,letterpaper]{article}
\usepackage[pagenumbers]{cvpr}
\def\confName{CVPR}
\def\confYear{2027}
\usepackage[T1]{fontenc}
\usepackage{mathtools}
\usepackage{microtype}
\usepackage{colortbl}
\DeclareMathOperator{\Concat}{Concat}
\DeclareMathOperator{\SiLU}{SiLU}
\DeclareMathOperator{\clip}{clip}
\DeclareMathOperator{\Corr}{Corr}
\DeclareMathOperator{\BCE}{BCE}
\newcommand{\E}{\mathbb{E}}
\newcommand{\R}{\mathbb{R}}

\DeclareRobustCommand{\metarefinetitle}{MetaRefine}

\DeclareRobustCommand{\captiongreen}[1]{{\setlength{\fboxsep}{0.6pt}\colorbox[RGB]{206,234,212}{#1}}}
\DeclareRobustCommand{\captionred}[1]{{\setlength{\fboxsep}{0.6pt}\colorbox[RGB]{246,211,211}{#1}}}
\usepackage{cuted}
\usepackage{stfloats}
\usepackage[nocheckfootnote]{flushend}
\definecolor{cvprblue}{rgb}{0.21,0.49,0.74}

\usepackage[breaklinks,colorlinks,allcolors=cvprblue]{hyperref}
\usepackage[capitalize]{cleveref}
\newcommand{\method}{MetaRefine\xspace}
\title{\texorpdfstring{\metarefinetitle}{MetaRefine}: Score Refinement with Multi-Level Features for Image Assessment}
\author{
{\normalsize Chuanzhi Xu\textsuperscript{1,*} \quad Chengkun Yue\textsuperscript{2} \quad Haolan Guo\textsuperscript{1} \quad Xuanhua Yin\textsuperscript{1}}\\
{\normalsize Boyu Wei\textsuperscript{1} \quad Zihan Deng\textsuperscript{3} \quad Qiang Qu\textsuperscript{1} \quad Weidong Cai\textsuperscript{1}}\\[3pt]
{\normalsize \textsuperscript{1}The University of Sydney \quad \textsuperscript{2}Indiana University \quad \textsuperscript{3}The University of Hong Kong \quad \href{https://github.com/LouckXu/MetaRefine}{\raisebox{-0.12em}{\includegraphics[height=0.9em]{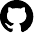}}\,Project Page}}
}
\hypersetup{
  pdftitle={MetaRefine: Score Refinement with Multi-Level Features for Image Assessment},
  pdfauthor={Chuanzhi Xu, Chengkun Yue, Haolan Guo, Xuanhua Yin, Boyu Wei, Zihan Deng, Qiang Qu, Weidong Cai},
  pdfsubject={Preprint}
}
\makeatletter
\patchcmd{\@maketitle}{\vskip .375in}{\vskip 0pt}{}{}
\patchcmd{\@maketitle}{\vspace*{24pt}}{\vspace*{10pt}}{}{}
\patchcmd{\@maketitle}{\vspace*{12pt}}{\vspace*{0pt}}{}{}
\makeatother
\begin{document}
\ifdefined\pdfminorversion
  \pdfminorversion=7
\fi
\raggedbottom
\maketitle
\begingroup
\renewcommand{\thefootnote}{*}
\footnotetext{Corresponding author: \href{mailto:chuanzhi.xu@sydney.edu.au}{\texttt{chuanzhi.xu@sydney.edu.au}}.}
\endgroup
\begin{strip}
\centering
\includegraphics[width=\linewidth]{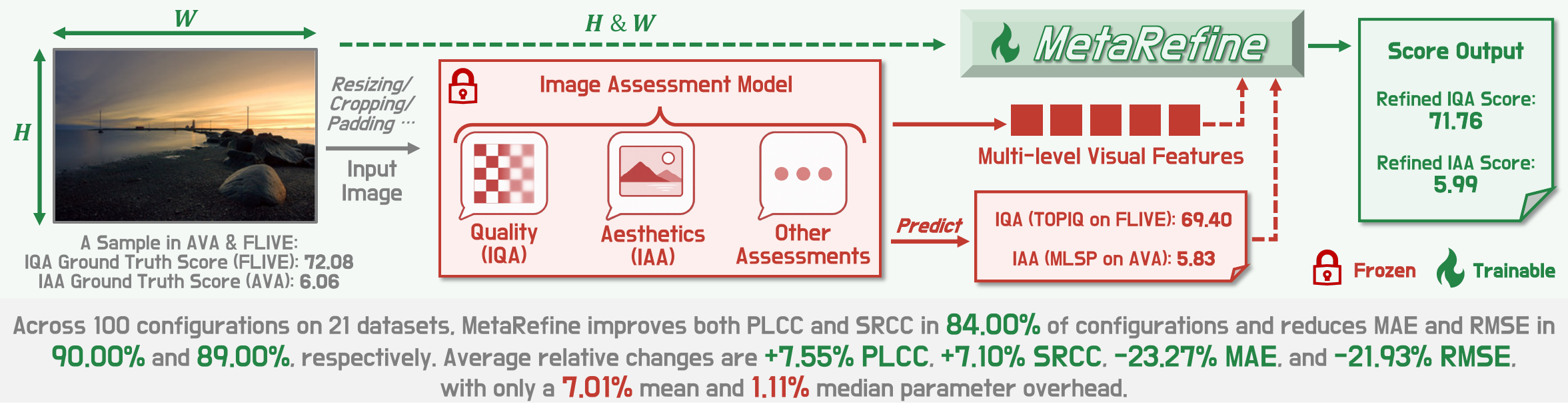}
\captionsetup{hypcap=false}
\captionof{figure}{\textbf{MetaRefine.} A metadata-aware score-refinement architecture for image assessment with higher correlations and lower errors.}
\label{fig:teaser}
\captionsetup{hypcap=true}
\end{strip}
\begin{abstract}
\interlinepenalty=10000
An image's original dimensions, aspect ratio, and composition provide important cues that can affect its technical and aesthetic quality. Image assessment methods use resizing, cropping, padding, or patch sampling to accommodate different input sizes. However, these operations can alter or obscure original geometric and compositional cues. Existing assessment models also encode rich multi-level visual evidence that can be further extracted and exploited to improve existing models. This motivates us to combine visual evidence with original-image geometry for score refinement without changing the original scoring pipelines. In this paper, we introduce \method, a metadata-aware score refinement architecture whose lightweight residual heads predict score corrections from geometry, multi-level visual features, and the frozen model's original score. Across 100 configurations on 21 datasets, \method improves PLCC in 88 configurations and SRCC in 85, with average relative gains of 7.55\% and 7.10\%, respectively. It reduces MAE in 90 configurations and RMSE in 89, with average relative reductions of 23.27\% and 21.93\%, respectively. The median parameter overhead is only 1.11\%. \method can further improve state-of-the-art (SOTA) assessment models on multiple benchmarks.
\par
\end{abstract}
\section{Introduction}
\label{sec:intro}
\begin{figure*}[t]
\centering
\includegraphics[width=\textwidth]{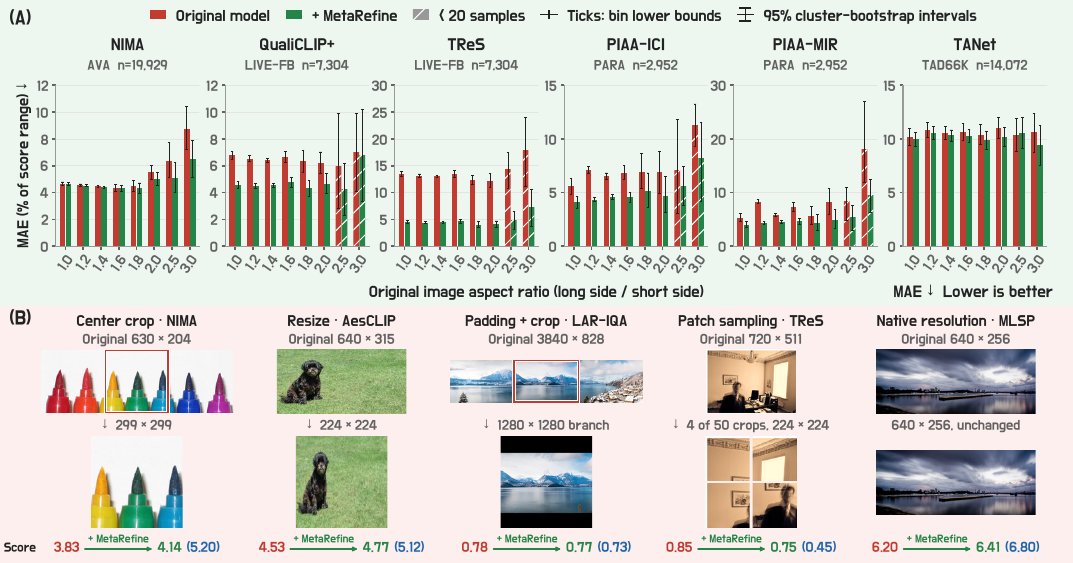}
\caption{\textbf{Refinement across image geometries and input pipelines.}
\textbf{(A)} Original and refined MAE across aspect-ratio bins for six assessment models~\cite{talebi2018nima,agnolucci2024qualiclip,golestaneh2022tres,chen2025individual,he2022tanet}.
\textbf{(B)} Illustrative score corrections under the models' original cropping, resizing, padding, patch sampling, and native-resolution inputs~\cite{talebi2018nima,sheng2023aesclip,avanaki2024lar,golestaneh2022tres,hosu2019mlsp}. Red/green: original/refined scores. Blue parentheses: ground truth.}
\label{fig:aspect-ratio-bars}
\end{figure*}
Image assessment predicts human judgments of images across several tasks. Two central tasks are image quality assessment (IQA), which evaluates technical quality affected by noise, blur, or compression artifacts, and image aesthetic assessment (IAA), which concerns photographic appeal shaped by composition, lighting, and style~\cite{talebi2018nima,deng2017survey}.

Despite their different targets, many models for these tasks share a common approach: a visual encoder extracts image features, and a scoring head maps those features to a score or rating distribution~\cite{talebi2018nima}. Research has improved this approach along several complementary directions. Multi-level feature fusion combines local appearance with high-level image content~\cite{hosu2019mlsp,chen2024topiq}. Multi-scale and aspect-ratio-preserving processing retains image detail and spatial structure~\cite{wang2019aspect,ke2021musiq,behrad2025charm}. Efficient adaptation makes pretrained representations more sensitive to quality with limited trainable parameters~\cite{xu2024loda}. Together, these directions strengthen the visual evidence available for assessment.

These advances also offer an opportunity to improve predictions by further exploiting the visual representations of existing image assessment models. Beyond the final predicted score, their multi-level intermediate features contain complementary information about local appearance and high-level image content~\cite{chen2024topiq,hosu2019mlsp}. This motivates us to extract and aggregate these visual features to improve score prediction without adding another image encoder.

Alongside this visual evidence, original-image geometry provides spatial context that may not remain explicit after input preprocessing. Resizing, cropping, padding, and patch sampling accommodate different image sizes and computational budgets (\Cref{fig:aspect-ratio-bars}(B)). Resizing to a common height and width removes original dimensions from the input shape and can alter image proportions. Cropping or patch sampling can remove global composition context~\cite{ke2021musiq,behrad2025charm}. Padding, native-resolution input, and multi-scale processing can preserve more of the original visual evidence~\cite{hosu2019mlsp,ke2021musiq,behrad2025charm}. However, original-image geometry, such as pixel area and aspect ratio, is not necessarily supplied explicitly to score prediction. Perceptual studies further show that presentation size can affect aesthetic judgments differently across images, while presentation resolution can influence perceived technical quality~\cite{chu2013size,wiedemann2023konx,hosu2025iisa}. Image aspect ratio is also associated with aesthetic preference~\cite{kuzinas2024creative} and can influence preferred subject placement within the frame~\cite{forman2021center}.

We evaluate existing image assessment models and examine how their prediction errors vary with original image geometry. \Cref{fig:aspect-ratio-bars}(A) shows different patterns of prediction error across original-image aspect ratios in several of the models presented. These observations further motivate us to consider combining original geometric context with the multi-level visual evidence already encoded by the models to improve their score predictions.

In this paper, we propose MetaRefine, a lightweight score refinement framework that combines multi-level features and original predictions from frozen image assessment models with geometric context (\Cref{fig:teaser,fig:aspect-ratio-bars}). The architecture extracts and aggregates these features during the original forward pass and explicitly encodes original-image geometry. Lightweight residual heads learn score corrections from target-dataset training labels.

Our contributions can be summarized as follows:
\begin{itemize}
  \item We introduce a common score refinement framework that applies across pretrained IQA and IAA architectures.
  \item We combine aggregated multi-level visual features and the original score with log pixel area and log aspect ratio as auxiliary context in an ensemble of residual heads without a magnitude bound. The design is also highly lightweight.
  \item \method achieves mean relative gains of 7.55\% in Pearson linear correlation coefficient (PLCC) and 7.10\% in Spearman rank correlation coefficient (SRCC), and reductions of 23.27\% in mean absolute error (MAE) and 21.93\% in root mean squared error (RMSE), with 1.64M added parameters on average (\Cref{fig:teaser}). Gains extend to state-of-the-art IQA and IAA models.
\end{itemize}
\section{Related Work}
\label{sec:related}
\method is effective across many of the image quality assessment and image aesthetic assessment methods below.

\noindent\textbf{Image Quality Assessment.}
IQA predicts technical quality affected by image distortions. We study no-reference IQA, without corresponding undistorted reference images. Common datasets include KADID-10k for synthetic distortions and KonIQ-10k, SPAQ, and LIVE-FB (FLIVE) for authentic distortions~\cite{lin2019kadid,hosu2020koniq,fang2020spaq,ying2020paq}. WaDIQaM-NR and RankIQA learn quality-sensitive representations through local quality weighting and ranking supervision~\cite{bosse2018wadiqam,liu2017rankiqa}. MUSIQ, TReS, and TOPIQ use multi-scale inputs or multi-level features to combine local distortions with global image content~\cite{ke2021musiq,golestaneh2022tres,chen2024topiq}. CLIP-IQA+ and QualiCLIP+ connect visual representations to quality descriptions through vision-language models~\cite{wang2023clipiqa,agnolucci2024qualiclip}.

\noindent\textbf{Other Image Quality Assessment.}
Beyond general-purpose quality assessment of natural photographs, specialized IQA considers panoramic, facial, and AI-generated images. Representative datasets include CVIQ for panoramic images~\cite{sun2018cviq}, CGFIQA-40k for faces~\cite{chen2024dslfiqa}, and AGIQA-3K for generated images~\cite{li2023agiqa}. Assessor360 and DSL-FIQA model viewport sequences and landmark-guided facial features, respectively~\cite{wu2023assessor360,chen2024dslfiqa}. IP-IQA and AMFF-Net combine visual information with text prompts to assess generated images~\cite{qu2024ipiqa,zhou2024amff}. These studies highlight how viewing behavior, facial structure, and image-prompt correspondence provide context for quality prediction.

\noindent\textbf{Image Aesthetic Assessment.}
IAA evaluates visual appeal~\cite{deng2017survey}, commonly benchmarked on AVA, AADB, TAD66K, and PARA~\cite{murray2012ava,kong2016aadb,he2022tanet,yang2022para}. Learning strategies exploit rating distributions or relative rankings (NIMA, FGAesQ)~\cite{talebi2018nima,yang2026fgaesq}, alongside image themes, aesthetic attributes, or observer characteristics (TANet, AesCLIP, PIAA-MIR and PIAA-ICI)~\cite{he2022tanet,sheng2023aesclip,chen2025individual}. MLSP, EAT, and Charm improve spatial representations through pooling, multi-scale processing, or attention~\cite{hosu2019mlsp,he2023eat,behrad2025charm}.
\section{Methodology}
\label{sec:method}
\begin{figure*}[t]
\centering
\includegraphics[width=\linewidth]{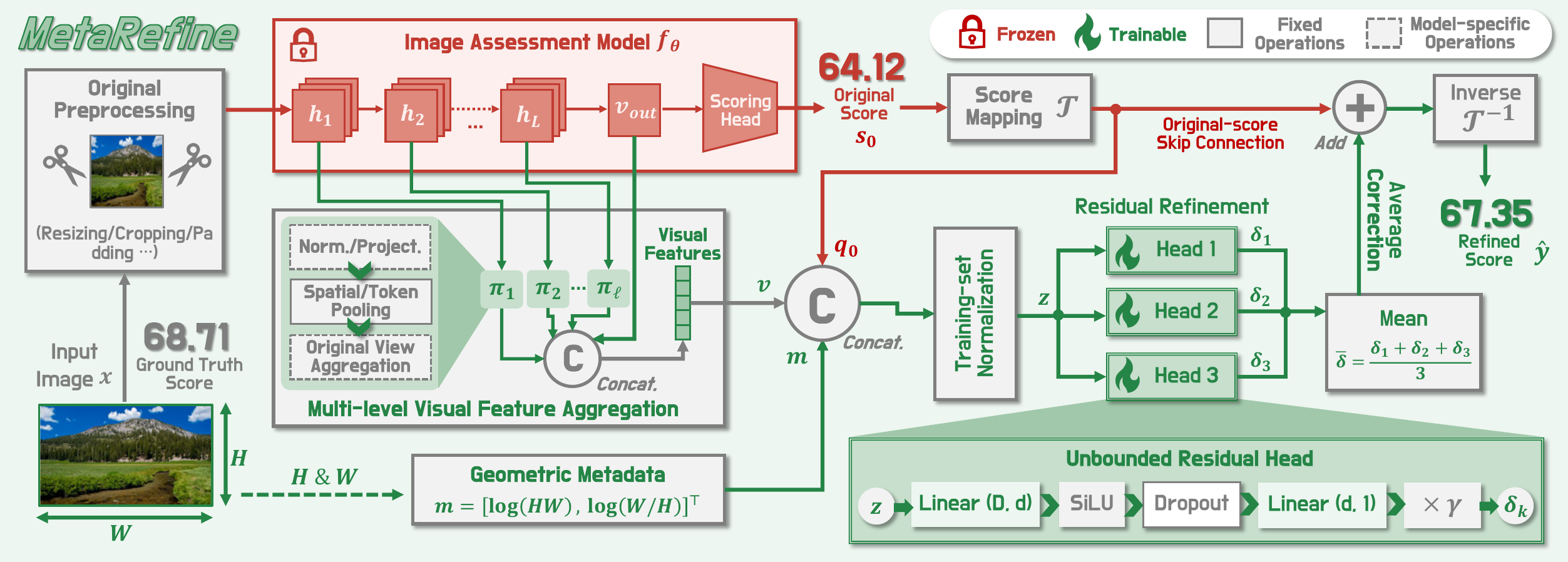}
\caption{\textbf{Overview of \method.} Metadata-aware score refinement using an existing frozen assessment model, with the residual head detailed at the bottom right. Only the residual heads are optimized.}
\label{fig:architecture}
\end{figure*}

\method is a metadata-aware score refinement architecture that requires an existing pretrained image assessment model, which remains frozen throughout refinement training and inference. As shown in \Cref{fig:architecture}, the frozen model provides its original score and multi-level visual features through its existing preprocessing and forward computation. In parallel, the metadata branch computes log pixel area and log aspect ratio from image dimensions recorded before preprocessing. After score mapping and joint input normalization, the residual heads use the original score, visual features, and geometric metadata to predict corrections without a fixed magnitude limit. These corrections are averaged and added to the original score on the refinement score scale, before the result is mapped back to the evaluation score scale.

\subsection{Problem Formulation}
\label{sec:formulation}
Let $x$ be an image with height $H$ and width $W$ measured in pixels before preprocessing, and let $y$ denote the ground-truth score provided by the dataset. A pretrained image assessment model $f_\theta$, including all of its feature extractors and scoring components, uses its original input configuration $P$ to predict:
\begin{equation}
 s_0(x)=f_{\theta}(P(x)),
 \label{eq:base}
\end{equation}
where $P(x)$ denotes the complete input configuration used to score $x$, including image preprocessing, original views, and any fixed auxiliary inputs or prescribed evaluation context. We freeze all learned parameters $\theta$ in the complete assessment model and retain its evaluation mode and fixed scoring state. With the original preprocessing and multi-view evaluation settings fixed, we train only the added residual heads using the training set $\{(x_i,y_i)\}_{i=1}^{n}$, where $n$ is the number of training images and subscript $i$ indexes image-specific quantities. This preserves the pretrained model's scoring computation while targeting its remaining prediction errors.

The frozen assessment model can be replaced by another pretrained model without modifying that model's architecture. For each chosen model, we configure feature extraction and the residual heads' input dimension, then train a separate set of \method residual heads with the same metadata definition, refinement architecture, and objective.

\subsection{Multi-Level Visual Feature Aggregation}
\label{sec:features}
We aggregate features from multiple stages of the frozen pretrained assessment model to provide visual information for score refinement. Let $h_{\ell}(x)$ denote the visual representation at extraction location $\ell$ in the frozen assessment model's original forward computation. Pooling and concatenation give:
\begin{equation}
 \mathbf v(x)=\Concat\!\left(
 \mathbf v_{\mathrm{out}}(x),\{\pi_{\ell}(h_{\ell}(x))\}_{\ell\in\mathcal S}
 \right),
 \label{eq:features}
\end{equation}
where $\Concat$ concatenates vectors in a fixed order, $\mathbf v_{\mathrm{out}}$ is the final visual representation used for scoring and $\mathcal S$ identifies the additional feature extraction locations. These locations may span multiple branches and include final-stage token summaries. The operator $\pi_\ell$ denotes the fixed aggregation needed to obtain a feature vector from each location. Aggregation details appear in Appendix~\ref{app:aggregation}.

\subsection{Geometric Context Modeling}
\label{sec:geometry}
The metadata branch uses the stored dimensions $H$ and $W$ recorded before the frozen assessment model's preprocessing. We encode the original image dimensions as a geometric context vector:
\begin{equation}
 \mathbf m(x)=\left[\log(HW),\;\log(W/H)\right]^{\top}.
 \label{eq:metadata}
\end{equation}
Log pixel area describes image size. In contrast, log aspect ratio describes shape and landscape or portrait orientation. Logarithms map multiplicative changes in dimensions to additive shifts and reciprocal aspect ratios to opposite signs.

Before forming the joint input, we map the original score $s_0$ and label $y$ from the evaluation score scale to the refinement score scale through a fixed affine map $\mathcal T(s)=as+b$, with $a>0$. The coefficients $a$ and $b$ are set by the task's score scale and do not depend on test predictions. Define $q_0=\mathcal T(s_0)$ and $\tilde y=\mathcal T(y)$.

We form the joint input from visual features $\mathbf v$, metadata $\mathbf m$, and mapped score $q_0$:
\begin{equation}
 \mathbf u=\Concat(\mathbf v,\mathbf m,q_0),\qquad
 \mathbf z=\mathcal N_{\mathrm{train}}(\mathbf u),
 \label{eq:input}
\end{equation}
where $\mathbf u,\mathbf z\in\R^D$ and $D=\dim(\mathbf v)+3$, accounting for two metadata features and one original score in addition to the visual features. The input normalization operator $\mathcal N_{\mathrm{train}}$ standardizes each input feature using training-set statistics, then clips the standardized values. Its detailed definition is provided in Appendix~\ref{app:normalization}.

\subsection{Unbounded Residual Refinement}
\label{sec:head}
Residual learning targets errors that remain predictable from the frozen model's visual representations and original score. Geometric metadata can provide additional context when it explains residual variation beyond these inputs. We use the joint input $\mathbf z$ to learn an image-dependent correction. Appendix~\ref{app:residual-interpretation} provides a supporting squared-error analysis.

The normalized joint input $\mathbf z$ feeds an unbounded residual head. For input dimension $D$ and hidden width $d$, each residual head predicts at inference:
\begin{equation}
 \delta_{\phi}(\mathbf z)=\gamma\left(
 \mathbf w^{\top}\SiLU(\mathbf W\mathbf z+\mathbf b)+c
 \right),
 \label{eq:head}
\end{equation}
where $\phi=\{\mathbf W,\mathbf b,\mathbf w,c\}$ is the set of trainable parameters, with $\mathbf W\in\R^{d\times D}$, $\mathbf b,\mathbf w\in\R^d$, and $c\in\R$. Here, $\SiLU$ denotes the sigmoid linear unit. The fixed output scale $\gamma>0$ multiplies the scalar output on the refinement score scale without imposing a magnitude limit. We initialize $\mathbf w$ and $c$ to zero, so every residual head initially predicts zero correction and the refined score equals the original score.

We independently train $K$ residual heads, with $\phi_k$ denoting the parameters of head $k$, and average their corrections at inference:
\begin{equation}
 \bar\delta(x)=\frac{1}{K}\sum_{k=1}^{K}\delta_{\phi_k}(\mathbf z(x)).
 \label{eq:ensemble}
\end{equation}
All residual heads share the normalized joint input $\mathbf z$ and reuse the frozen assessment model's original forward computation. We add the average correction $\bar\delta$ to the original score $q_0$ on the refinement score scale, then apply $\mathcal T^{-1}$ to obtain the final refined score $\hat y$ on the evaluation score scale:
\begin{equation}
 \hat y(x)=\mathcal T^{-1}\!\left(q_0(x)+\bar\delta(x)\right).
 \label{eq:prediction}
\end{equation}
Averaging can reduce training variability when the learned corrections are not perfectly correlated. Each correction depends on the image's visual and geometric context as well as its original score. Therefore, refinement can adjust both score values and image rankings.

The two geometric metadata features contribute only $2Kd$ weights to the residual heads' input layers. The parameter-count derivation is in Appendix~\ref{app:parameter-count}.

\subsection{Joint Regression and Ranking Objective}
\label{sec:objective}
We train each residual head to account for both human scores and relative preferences between images. Each update uses all training images for regression and correlation, with sampled image pairs for ranking. Fixed features and original scores can be cached, avoiding repeated computation through the frozen model during residual-head training. For a single residual head with parameters $\phi$, we omit the head index $k$ and let $\delta_i=\delta_\phi(\mathbf z_i)$ denote its correction, $q_i=q_{0,i}+\delta_i$ its refined score, and $e_i=q_i-\tilde y_i$ its error.

To supervise relative image preferences, let $\mathcal P$ denote the multiset of sampled training-image index pairs. For each $(i,j)\in\mathcal P$, we define soft pairwise preference targets and predictions:
\begin{equation}
 p_{ij}=\sigma\!\left(\frac{\tilde y_i-\tilde y_j}{T_{\mathrm{rank}}}\right),\qquad
 \hat p_{ij}=\sigma\!\left(\frac{q_i-q_j}{T_{\mathrm{rank}}}\right),
 \label{eq:preferences}
\end{equation}
where $\sigma$ is the logistic sigmoid and $T_{\mathrm{rank}}>0$ is the ranking temperature on the refinement score scale. The pairwise objective is:
\begin{equation}
 \mathcal L_{\mathrm{pair}}=\frac{1}{|\mathcal P|}
 \sum_{(i,j)\in\mathcal P}\BCE(p_{ij},\hat p_{ij}).
 \label{eq:pairloss}
\end{equation}
Here, $\BCE$ denotes binary cross-entropy. The soft targets encode the direction and strength of score differences, guiding how the residual heads adjust comparisons between images.

The joint objective combines robust regression, correlation, pairwise ranking, and residual regularization. Let $H_\kappa$ be the Huber loss with transition $\kappa>0$. We use $\epsilon>0$ as the smoothing constant for the absolute-error term and $\Corr$ for Pearson correlation. The vectors $\mathbf q$ and $\tilde{\mathbf y}$ collect the single residual head's refined training scores and corresponding ground-truth scores on the refinement score scale. We minimize:
\begin{align}
 \mathcal L={}&\frac{1}{n}\sum_{i=1}^{n}\frac{H_{\kappa}(e_i)}{\kappa}
 +\lambda_{\mathrm{abs}}\frac{1}{n}\sum_{i=1}^{n}\sqrt{e_i^2+\epsilon^2}
 \nonumber\\
 &+\lambda_{\mathrm{corr}}\big[1-\Corr(\mathbf q,\tilde{\mathbf y})\big]
 \nonumber\\
 &+\lambda_{\mathrm{pair}}\mathcal L_{\mathrm{pair}}
 +\lambda_{\mathrm{res}}\frac{1}{n}\sum_{i=1}^{n}\delta_i^2.
 \label{eq:loss}
\end{align}
\begin{figure}[!t]
\centering
\includegraphics[width=\columnwidth]{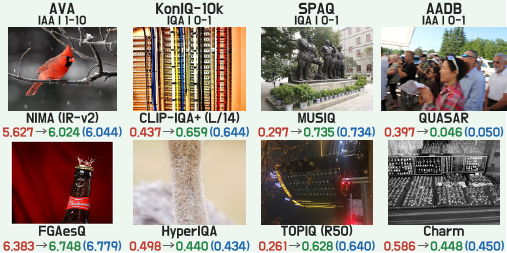}
\caption{Visual examples of score refinement across datasets. Original (red), \method (green), and GT (blue).}
\label{fig:qualitative-main}
\end{figure}
The nonnegative weights $\lambda_{\mathrm{abs}}$, $\lambda_{\mathrm{corr}}$, $\lambda_{\mathrm{pair}}$, and $\lambda_{\mathrm{res}}$ balance the smoothed absolute error, correlation, pairwise ranking, and residual penalty, respectively. The pointwise terms penalize score errors. By contrast, the correlation term encourages agreement with the variation in human ratings across the training set. Pearson correlation is unchanged by a positive affine transformation of the predictions, so the pointwise terms also anchor predictions to the target score scale. The residual penalty discourages large adjustments through a soft quadratic cost without imposing a hard magnitude limit. The same objective and residual-head design are used across assessment models. Pair sampling and loss-function definitions are provided in Appendix~\ref{app:loss-definitions}. Hyperparameters and model-specific feature extraction details are described in Appendices~\ref{app:optimization}-\ref{app:feature-extraction}.
\section{Experiments and Results}
\label{sec:experiments}

\subsection{Experimental Setup}
\noindent\textbf{Training and Inference.} We use three independently initialized residual heads with hidden width 64, output scale $\gamma=0.375$, and dropout probability $0.5$. Each head uses target-dataset training labels for 85 full-batch AdamW updates (learning rate $10^{-3}$, weight decay $0.02$), with cached features and scores. These head and optimization settings are shared across configurations. Model-specific feature extraction, data partitions, and checkpoint protocols appear in Appendices~\ref{app:feature-extraction} and \ref{app:data-protocols}-\ref{app:retrained-models}.

\noindent\textbf{Evaluation Protocol.} We evaluate \method with 51 IQA and IAA methods across 100 model-dataset configurations on 21 datasets. For each configuration, we compare the frozen assessment model's original scores with the refined scores on the same evaluation images, retaining its configured preprocessing and evaluation context. We use PLCC, SRCC, MAE, and RMSE as evaluation metrics.

\subsection{Comparison with Existing Methods}
\renewcommand{\dblfloatpagefraction}{0.6}
\begin{table*}[p]
\centering
\caption{\textbf{Complete score-refinement results, part 1 of 2 (52 configurations).} Base: frozen baseline assessment model. Ref.: the baseline model after \method refinement. Gain is the relative change from Base to Ref., computed by dividing the increase in SRCC/PLCC or the reduction in MAE/RMSE by the absolute Base value and expressing the result as a percentage. Params $\uparrow$: added trainable parameters relative to the complete frozen baseline (\%). Darker \captiongreen{green}/\captionred{red} indicates larger relative improvement/decline, respectively, with color intensity capped at a 30\% change. Improvements correspond to higher SRCC/PLCC or lower MAE/RMSE. Rows are sorted by refined SRCC within each dataset. Test/Total gives the default evaluation count and total sample count used in that dataset protocol. Data partitions and evaluation protocols are detailed in Appendices~\ref{app:data-protocols}-\ref{app:score-scales}.}
\label{tab:complete-results-1}
\begingroup
\fontsize{8}{9}\selectfont
\setlength{\tabcolsep}{1.3pt}
\renewcommand{\arraystretch}{0.98}
\begin{tabular*}{\textwidth}{@{\extracolsep{\fill}}l*{4}{rr>{\fontsize{7}{9}\selectfont}r}@{\extracolsep{0pt}\hspace{3pt}}r@{}}
\toprule
 & \multicolumn{3}{c}{SRCC {\fontsize{7.5}{9}\selectfont$\uparrow$}} & \multicolumn{3}{c}{PLCC {\fontsize{7.5}{9}\selectfont$\uparrow$}} & \multicolumn{3}{c}{MAE {\fontsize{7.5}{9}\selectfont$\downarrow$}} & \multicolumn{3}{c}{RMSE {\fontsize{7.5}{9}\selectfont$\downarrow$}} & \\
\cmidrule(lr){2-4}\cmidrule(lr){5-7}\cmidrule(lr){8-10}\cmidrule(lr){11-13}
Assessment model & \multicolumn{1}{c}{Base} & \multicolumn{1}{c}{Ref.} & \multicolumn{1}{c}{Gain} & \multicolumn{1}{c}{Base} & \multicolumn{1}{c}{Ref.} & \multicolumn{1}{c}{Gain} & \multicolumn{1}{c}{Base} & \multicolumn{1}{c}{Ref.} & \multicolumn{1}{c}{Gain} & \multicolumn{1}{c}{Base} & \multicolumn{1}{c}{Ref.} & \multicolumn{1}{c}{Gain} & Params {\fontsize{7.5}{9}\selectfont$\uparrow$} \\
\midrule
\addlinespace[1pt]\rowcolor[gray]{0.92}\multicolumn{14}{@{}l}{\textbf{KonIQ-10k (IQA)}~\cite{hosu2020koniq} - Test/Total: 2,015 / 10,073} \\
DP-IQA (Student)~\cite{fu2024dpiqa} & 0.9752 & 0.9781 & \cellcolor[RGB]{229,244,232}+0.30\% & 0.9808 & 0.9828 & \cellcolor[RGB]{231,244,233}+0.21\% & 0.0312 & 0.0203 & \cellcolor[RGB]{142,205,157}+34.93\% & 0.0396 & 0.0285 & \cellcolor[RGB]{145,206,160}+27.97\% & 0.26\% \\
MANIQA~\cite{yang2022maniqa} & 0.9022 & 0.9505 & \cellcolor[RGB]{198,230,205}+5.35\% & 0.9280 & 0.9639 & \cellcolor[RGB]{204,233,210}+3.87\% & 0.1521 & 0.0292 & \cellcolor[RGB]{142,205,157}+80.79\% & 0.1614 & 0.0377 & \cellcolor[RGB]{142,205,157}+76.63\% & 0.71\% \\
RALI~\cite{zhao2026rali} & 0.9248 & 0.9440 & \cellcolor[RGB]{213,237,218}+2.08\% & 0.9410 & 0.9552 & \cellcolor[RGB]{217,238,221}+1.51\% & 0.0494 & 0.0354 & \cellcolor[RGB]{145,206,159}+28.36\% & 0.0632 & 0.0458 & \cellcolor[RGB]{146,207,161}+27.50\% & 0.24\% \\
TReS~\cite{golestaneh2022tres} & 0.9362 & 0.9406 & \cellcolor[RGB]{227,243,230}+0.48\% & 0.9461 & 0.9554 & \cellcolor[RGB]{221,240,225}+0.99\% & 0.1602 & 0.0348 & \cellcolor[RGB]{142,205,157}+78.29\% & 0.1736 & 0.0465 & \cellcolor[RGB]{142,205,157}+73.20\% & 0.74\% \\
UNIQUE (mix)~\cite{zhang2021unique} & 0.9288 & 0.9377 & \cellcolor[RGB]{222,240,225}+0.95\% & 0.9334 & 0.9463 & \cellcolor[RGB]{218,239,222}+1.38\% & 0.6644 & 0.0568 & \cellcolor[RGB]{142,205,157}+91.45\% & 0.7752 & 0.0727 & \cellcolor[RGB]{142,205,157}+90.62\% & 225.87\% \\
QCN~\cite{shin2024qcn} & 0.9341 & 0.9349 & \cellcolor[RGB]{234,246,236}+0.08\% & 0.9438 & 0.9451 & \cellcolor[RGB]{233,245,234}+0.13\% & 0.0379 & 0.0374 & \cellcolor[RGB]{218,239,222}+1.39\% & 0.0509 & 0.0501 & \cellcolor[RGB]{216,238,220}+1.69\% & 1.23\% \\
TOPIQ~\cite{chen2024topiq} & 0.9299 & 0.9300 & \cellcolor[RGB]{237,247,238}+0.02\% & 0.9436 & 0.9444 & \cellcolor[RGB]{234,246,236}+0.09\% & 0.0388 & 0.0383 & \cellcolor[RGB]{218,239,222}+1.37\% & 0.0511 & 0.0507 & \cellcolor[RGB]{224,242,228}+0.68\% & 0.87\% \\
TRIQA~\cite{sureddi2025triqa} & 0.8023 & 0.9259 & \cellcolor[RGB]{169,217,181}+15.41\% & 0.8412 & 0.9376 & \cellcolor[RGB]{179,221,189}+11.46\% & 0.0972 & 0.0415 & \cellcolor[RGB]{142,205,157}+57.25\% & 0.1133 & 0.0536 & \cellcolor[RGB]{142,205,157}+52.68\% & 1.53\% \\
Charm~\cite{behrad2025charm} & 0.8901 & 0.9225 & \cellcolor[RGB]{205,233,211}+3.64\% & 0.9120 & 0.9387 & \cellcolor[RGB]{209,235,214}+2.93\% & 0.0619 & 0.0407 & \cellcolor[RGB]{142,205,157}+34.25\% & 0.0802 & 0.0534 & \cellcolor[RGB]{142,205,157}+33.47\% & 2.74\% \\
QualiCLIP+~\cite{agnolucci2024qualiclip} & 0.8966 & 0.9191 & \cellcolor[RGB]{211,236,216}+2.51\% & 0.8995 & 0.9264 & \cellcolor[RGB]{208,234,214}+3.00\% & 0.0608 & 0.0442 & \cellcolor[RGB]{147,207,161}+27.20\% & 0.0769 & 0.0587 & \cellcolor[RGB]{153,210,166}+23.62\% & 0.53\% \\
KonCept512~\cite{hosu2020koniq} & 0.9178 & 0.9178 & \cellcolor[RGB]{238,248,239}+0.00\% & 0.9314 & 0.9332 & \cellcolor[RGB]{231,245,233}+0.19\% & 0.0930 & 0.0420 & \cellcolor[RGB]{142,205,157}+54.78\% & 0.1039 & 0.0556 & \cellcolor[RGB]{142,205,157}+46.52\% & 0.37\% \\
CLIP-IQA+~\cite{wang2023clipiqa} & 0.8847 & 0.9121 & \cellcolor[RGB]{208,234,213}+3.09\% & 0.9012 & 0.9250 & \cellcolor[RGB]{210,235,215}+2.64\% & 0.0649 & 0.0452 & \cellcolor[RGB]{142,205,157}+30.43\% & 0.0813 & 0.0590 & \cellcolor[RGB]{146,207,161}+27.46\% & 0.53\% \\
CLIP-IQA+ (L/14)~\cite{wang2023clipiqa} & 0.8610 & 0.9114 & \cellcolor[RGB]{196,229,203}+5.85\% & 0.8747 & 0.9285 & \cellcolor[RGB]{195,229,202}+6.15\% & 0.0602 & 0.0444 & \cellcolor[RGB]{148,208,162}+26.25\% & 0.0770 & 0.0578 & \cellcolor[RGB]{151,209,164}+24.94\% & 0.24\% \\
MUSIQ~\cite{ke2021musiq} & 0.8699 & 0.9108 & \cellcolor[RGB]{201,231,207}+4.71\% & 0.8977 & 0.9304 & \cellcolor[RGB]{205,233,211}+3.64\% & 0.0615 & 0.0427 & \cellcolor[RGB]{142,205,157}+30.59\% & 0.0815 & 0.0567 & \cellcolor[RGB]{142,205,157}+30.36\% & 2.18\% \\
HyperIQA~\cite{su2020hyperiqa} & 0.9031 & 0.9099 & \cellcolor[RGB]{224,241,227}+0.76\% & 0.9227 & 0.9295 & \cellcolor[RGB]{224,241,227}+0.74\% & 0.0452 & 0.0433 & \cellcolor[RGB]{203,232,210}+4.04\% & 0.0599 & 0.0570 & \cellcolor[RGB]{200,231,207}+4.85\% & 1.42\% \\
DBCNN~\cite{zhang2020dbcnn} & 0.9034 & 0.9062 & \cellcolor[RGB]{229,244,232}+0.31\% & 0.9197 & 0.9226 & \cellcolor[RGB]{229,244,231}+0.32\% & 0.0472 & 0.0451 & \cellcolor[RGB]{201,231,208}+4.51\% & 0.0619 & 0.0600 & \cellcolor[RGB]{208,234,213}+3.06\% & 83.31\% \\
LIQE (mix)~\cite{zhang2023liqe} & 0.9242 & 0.8901 & \cellcolor[RGB]{247,209,209}-3.69\% & 0.9073 & 0.9075 & \cellcolor[RGB]{236,247,238}+0.02\% & 0.1867 & 0.0542 & \cellcolor[RGB]{142,205,157}+70.96\% & 0.2121 & 0.0696 & \cellcolor[RGB]{142,205,157}+67.16\% & 0.52\% \\
NIMA~\cite{talebi2018nima} & 0.8802 & 0.8875 & \cellcolor[RGB]{223,241,226}+0.82\% & 0.9107 & 0.9158 & \cellcolor[RGB]{226,242,229}+0.56\% & 0.0495 & 0.0469 & \cellcolor[RGB]{198,230,205}+5.31\% & 0.0642 & 0.0620 & \cellcolor[RGB]{206,234,212}+3.39\% & 1.11\% \\
ARNIQA (SPAQ)~\cite{agnolucci2024arniqa} & 0.7342 & 0.8817 & \cellcolor[RGB]{160,213,172}+20.09\% & 0.7720 & 0.9007 & \cellcolor[RGB]{167,216,178}+16.68\% & 0.0830 & 0.0511 & \cellcolor[RGB]{142,205,157}+38.48\% & 0.1072 & 0.0675 & \cellcolor[RGB]{142,205,157}+37.04\% & 6.28\% \\
LIQE~\cite{zhang2023liqe} & 0.8967 & 0.8663 & \cellcolor[RGB]{247,210,210}-3.40\% & 0.9023 & 0.9063 & \cellcolor[RGB]{227,243,230}+0.45\% & 0.1855 & 0.0556 & \cellcolor[RGB]{142,205,157}+70.01\% & 0.2112 & 0.0707 & \cellcolor[RGB]{142,205,157}+66.54\% & 0.52\% \\
QPT V2 (IQA)\textsuperscript{r}~\cite{xie2024qpt} & 0.8637 & 0.8648 & \cellcolor[RGB]{232,245,234}+0.14\% & 0.8848 & 0.8854 & \cellcolor[RGB]{235,246,236}+0.06\% & 0.0616 & 0.0539 & \cellcolor[RGB]{177,220,187}+12.38\% & 0.0772 & 0.0717 & \cellcolor[RGB]{192,227,200}+7.08\% & 1.08\% \\
WaDIQaM-NR~\cite{bosse2018wadiqam} & 0.8046 & 0.3557 & \cellcolor[RGB]{235,151,151}-55.79\% & 0.8300 & 0.3591 & \cellcolor[RGB]{235,151,151}-56.73\% & 0.8095 & 0.2740 & \cellcolor[RGB]{142,205,157}+66.15\% & 0.8381 & 0.3472 & \cellcolor[RGB]{142,205,157}+58.58\% & 3.54\% \\
\addlinespace[1pt]\rowcolor[gray]{0.92}\multicolumn{14}{@{}l}{\textbf{KADID-10k (IQA)}~\cite{lin2019kadid} - Test/Total: 2,125 / 10,125} \\
MANIQA~\cite{yang2022maniqa} & 0.9777 & 0.9782 & \cellcolor[RGB]{235,246,237}+0.05\% & 0.9784 & 0.9789 & \cellcolor[RGB]{235,246,237}+0.05\% & 0.0551 & 0.0422 & \cellcolor[RGB]{153,210,167}+23.49\% & 0.0712 & 0.0564 & \cellcolor[RGB]{158,212,171}+20.71\% & 0.71\% \\
ARNIQA (KonIQ)~\cite{agnolucci2024arniqa} & 0.7599 & 0.9194 & \cellcolor[RGB]{158,212,171}+20.99\% & 0.7369 & 0.9192 & \cellcolor[RGB]{151,209,165}+24.74\% & 0.1988 & 0.0822 & \cellcolor[RGB]{142,205,157}+58.67\% & 0.2400 & 0.1081 & \cellcolor[RGB]{142,205,157}+54.97\% & 6.28\% \\
NLNet~\cite{jia2022nlnet} & 0.8418 & 0.8701 & \cellcolor[RGB]{207,234,212}+3.36\% & 0.8459 & 0.8683 & \cellcolor[RGB]{210,235,215}+2.65\% & 0.1369 & 0.1149 & \cellcolor[RGB]{168,216,179}+16.13\% & 0.1791 & 0.1532 & \cellcolor[RGB]{172,218,182}+14.47\% & 1.34\% \\
HyperIQA (KonIQ)~\cite{su2020hyperiqa} & 0.5125 & 0.8052 & \cellcolor[RGB]{142,205,157}+57.12\% & 0.5317 & 0.7959 & \cellcolor[RGB]{142,205,157}+49.67\% & 0.2082 & 0.1328 & \cellcolor[RGB]{142,205,157}+36.23\% & 0.2564 & 0.1708 & \cellcolor[RGB]{142,205,157}+33.40\% & 1.42\% \\
MUSIQ (KonIQ)~\cite{ke2021musiq} & 0.5767 & 0.7853 & \cellcolor[RGB]{142,205,157}+36.17\% & 0.5782 & 0.7730 & \cellcolor[RGB]{142,205,157}+33.69\% & 0.2233 & 0.1569 & \cellcolor[RGB]{142,205,157}+29.75\% & 0.2761 & 0.1993 & \cellcolor[RGB]{146,207,160}+27.81\% & 2.18\% \\
SIND\textsuperscript{r}~\cite{huang2025sind} & 0.8606 & 0.7634 & \cellcolor[RGB]{242,185,185}-11.29\% & 0.8428 & 0.7676 & \cellcolor[RGB]{243,191,191}-8.92\% & 1.5902 & 0.2740 & \cellcolor[RGB]{142,205,157}+82.77\% & 1.8116 & 0.3269 & \cellcolor[RGB]{142,205,157}+81.95\% & 0.43\% \\
DFSS-IQA\textsuperscript{r}~\cite{chen2024dfss} & 0.7764 & 0.7553 & \cellcolor[RGB]{248,213,213}-2.71\% & 0.3597 & 0.5109 & \cellcolor[RGB]{142,205,157}+42.04\% & 0.4948 & 0.1830 & \cellcolor[RGB]{142,205,157}+63.02\% & 1.6334 & 0.4215 & \cellcolor[RGB]{142,205,157}+74.20\% & 1.15\% \\
PAUQA\textsuperscript{r}~\cite{lin2025pauqa} & 0.6200 & 0.6713 & \cellcolor[RGB]{188,225,196}+8.27\% & 0.6100 & 0.6545 & \cellcolor[RGB]{191,227,199}+7.28\% & 0.2088 & 0.2124 & \cellcolor[RGB]{249,219,219}-1.74\% & 0.2599 & 0.2631 & \cellcolor[RGB]{249,222,222}-1.23\% & 1.01\% \\
\addlinespace[1pt]\rowcolor[gray]{0.92}\multicolumn{14}{@{}l}{\textbf{SPAQ (IQA)}~\cite{fang2020spaq} - Test/Total: 2,129 / 11,125} \\
DP-IQA (Student)~\cite{fu2024dpiqa} & 0.9339 & 0.9333 & \cellcolor[RGB]{252,236,236}-0.07\% & 0.9381 & 0.9381 & \cellcolor[RGB]{253,239,239}-0.01\% & 0.0589 & 0.0569 & \cellcolor[RGB]{206,233,212}+3.46\% & 0.0745 & 0.0720 & \cellcolor[RGB]{207,234,213}+3.27\% & 0.26\% \\
QualiCLIP+~\cite{agnolucci2024qualiclip} & 0.9149 & 0.9204 & \cellcolor[RGB]{225,242,228}+0.60\% & 0.9186 & 0.9254 & \cellcolor[RGB]{224,241,227}+0.74\% & 0.0757 & 0.0632 & \cellcolor[RGB]{167,216,178}+16.56\% & 0.0948 & 0.0811 & \cellcolor[RGB]{172,218,182}+14.48\% & 0.53\% \\
QCN (384)~\cite{shin2024qcn} & 0.9092 & 0.9167 & \cellcolor[RGB]{223,241,226}+0.83\% & 0.9062 & 0.9195 & \cellcolor[RGB]{218,238,222}+1.47\% & 0.0836 & 0.0648 & \cellcolor[RGB]{155,211,168}+22.45\% & 0.1048 & 0.0820 & \cellcolor[RGB]{156,211,169}+21.78\% & 1.23\% \\
NIMA (IR-v2)~\cite{talebi2018nima} & 0.9108 & 0.9133 & \cellcolor[RGB]{230,244,232}+0.28\% & 0.9157 & 0.9183 & \cellcolor[RGB]{229,244,232}+0.29\% & 0.0672 & 0.0659 & \cellcolor[RGB]{214,237,218}+2.03\% & 0.0856 & 0.0838 & \cellcolor[RGB]{213,236,218}+2.17\% & 1.11\% \\
NIMA (KonIQ)~\cite{talebi2018nima} & 0.8572 & 0.9064 & \cellcolor[RGB]{197,229,204}+5.74\% & 0.8578 & 0.9135 & \cellcolor[RGB]{194,228,201}+6.49\% & 0.0915 & 0.0665 & \cellcolor[RGB]{146,207,161}+27.31\% & 0.1146 & 0.0847 & \cellcolor[RGB]{149,208,163}+26.09\% & 1.11\% \\
MUSIQ (SPAQ)~\cite{ke2021musiq} & 0.6984 & 0.9009 & \cellcolor[RGB]{144,206,158}+28.99\% & 0.6954 & 0.9057 & \cellcolor[RGB]{142,205,157}+30.24\% & 0.1824 & 0.0695 & \cellcolor[RGB]{142,205,157}+61.88\% & 0.2210 & 0.0893 & \cellcolor[RGB]{142,205,157}+59.62\% & 2.18\% \\
MUSIQ (KonIQ)~\cite{ke2021musiq} & 0.6040 & 0.8892 & \cellcolor[RGB]{142,205,157}+47.22\% & 0.5437 & 0.8944 & \cellcolor[RGB]{142,205,157}+64.48\% & 0.2526 & 0.0741 & \cellcolor[RGB]{142,205,157}+70.67\% & 0.2985 & 0.0949 & \cellcolor[RGB]{142,205,157}+68.23\% & 2.18\% \\
TOPIQ (R50)~\cite{chen2024topiq} & 0.7491 & 0.8874 & \cellcolor[RGB]{163,214,175}+18.46\% & 0.7423 & 0.8914 & \cellcolor[RGB]{160,213,172}+20.09\% & 0.2015 & 0.0740 & \cellcolor[RGB]{142,205,157}+63.28\% & 0.2339 & 0.0954 & \cellcolor[RGB]{142,205,157}+59.20\% & 0.87\% \\
\addlinespace[1pt]\rowcolor[gray]{0.92}\multicolumn{14}{@{}l}{\textbf{LIVE Challenge (IQA)}~\cite{ghadiyaram2016challenge} - Test/Total: 231 / 1,162} \\
DBCNN~\cite{zhang2020dbcnn} & 0.9618 & 0.9594 & \cellcolor[RGB]{251,232,232}-0.24\% & 0.9683 & 0.9649 & \cellcolor[RGB]{251,230,230}-0.36\% & 0.0324 & 0.0349 & \cellcolor[RGB]{244,195,195}-7.80\% & 0.0525 & 0.0532 & \cellcolor[RGB]{249,221,221}-1.40\% & 83.31\% \\
MetaIQA~\cite{zhu2020metaiqa} & 0.8921 & 0.9015 & \cellcolor[RGB]{221,240,224}+1.06\% & 0.8982 & 0.9234 & \cellcolor[RGB]{209,235,215}+2.81\% & 0.0915 & 0.0600 & \cellcolor[RGB]{142,205,157}+34.48\% & 0.1201 & 0.0778 & \cellcolor[RGB]{142,205,157}+35.27\% & 1.40\% \\
HyperIQA (baseline)~\cite{su2020hyperiqa} & 0.8886 & 0.8900 & \cellcolor[RGB]{232,245,234}+0.15\% & 0.9077 & 0.9100 & \cellcolor[RGB]{230,244,232}+0.26\% & 0.0767 & 0.0757 & \cellcolor[RGB]{218,239,222}+1.37\% & 0.0956 & 0.0955 & \cellcolor[RGB]{232,245,234}+0.14\% & 1.42\% \\
HyperIQA+NT~\cite{liu2024defense} & 0.8665 & 0.8621 & \cellcolor[RGB]{251,228,228}-0.51\% & 0.8760 & 0.8831 & \cellcolor[RGB]{223,241,226}+0.82\% & 0.0882 & 0.0806 & \cellcolor[RGB]{187,225,196}+8.57\% & 0.1097 & 0.1013 & \cellcolor[RGB]{190,226,198}+7.64\% & 1.42\% \\
QualiCLIP+~\cite{agnolucci2024qualiclip} & 0.8382 & 0.8572 & \cellcolor[RGB]{212,236,217}+2.27\% & 0.8604 & 0.8779 & \cellcolor[RGB]{214,237,218}+2.03\% & 0.0825 & 0.0702 & \cellcolor[RGB]{170,218,181}+14.96\% & 0.1023 & 0.0931 & \cellcolor[RGB]{186,224,195}+8.99\% & 0.53\% \\
MUSIQ (KonIQ)~\cite{ke2021musiq} & 0.7600 & 0.8209 & \cellcolor[RGB]{189,226,197}+8.01\% & 0.8115 & 0.8539 & \cellcolor[RGB]{199,230,205}+5.22\% & 0.0968 & 0.0806 & \cellcolor[RGB]{166,216,178}+16.77\% & 0.1235 & 0.1063 & \cellcolor[RGB]{173,219,183}+13.95\% & 2.18\% \\
NIMA (KonIQ)~\cite{talebi2018nima} & 0.7847 & 0.7894 & \cellcolor[RGB]{225,242,228}+0.60\% & 0.8275 & 0.8483 & \cellcolor[RGB]{211,236,216}+2.52\% & 0.0955 & 0.0829 & \cellcolor[RGB]{175,219,185}+13.22\% & 0.1190 & 0.1074 & \cellcolor[RGB]{184,223,193}+9.81\% & 1.11\% \\
ARNIQA (KonIQ)~\cite{agnolucci2024arniqa} & 0.6219 & 0.7779 & \cellcolor[RGB]{150,209,164}+25.08\% & 0.6889 & 0.8259 & \cellcolor[RGB]{160,213,172}+19.88\% & 0.1239 & 0.0907 & \cellcolor[RGB]{147,207,161}+26.84\% & 0.1593 & 0.1141 & \cellcolor[RGB]{145,206,159}+28.40\% & 6.28\% \\
\addlinespace[1pt]\rowcolor[gray]{0.92}\multicolumn{14}{@{}l}{\textbf{FLIVE (IQA)}~\cite{ying2020paq} - Test/Total: 7,304 / 159,240 (images and crops)} \\
DP-IQA (Student)~\cite{fu2024dpiqa} & 0.7588 & 0.7827 & \cellcolor[RGB]{208,234,213}+3.15\% & 0.8366 & 0.8572 & \cellcolor[RGB]{211,236,216}+2.46\% & 0.0319 & 0.0299 & \cellcolor[RGB]{195,228,202}+6.28\% & 0.0430 & 0.0404 & \cellcolor[RGB]{195,229,203}+6.07\% & 0.26\% \\
MUSIQ~\cite{ke2021musiq} & 0.7443 & 0.7498 & \cellcolor[RGB]{224,241,227}+0.74\% & 0.7829 & 0.7887 & \cellcolor[RGB]{224,241,227}+0.75\% & 0.0410 & 0.0408 & \cellcolor[RGB]{228,243,231}+0.37\% & 0.0549 & 0.0545 & \cellcolor[RGB]{223,241,227}+0.79\% & 2.18\% \\
TOPIQ~\cite{chen2024topiq} & 0.7261 & 0.7429 & \cellcolor[RGB]{212,236,217}+2.32\% & 0.7564 & 0.7766 & \cellcolor[RGB]{210,235,215}+2.67\% & 0.0446 & 0.0418 & \cellcolor[RGB]{194,228,202}+6.35\% & 0.0590 & 0.0556 & \cellcolor[RGB]{197,229,204}+5.75\% & 0.87\% \\
QualiCLIP+~\cite{agnolucci2024qualiclip} & 0.6537 & 0.7199 & \cellcolor[RGB]{183,223,192}+10.12\% & 0.6781 & 0.7408 & \cellcolor[RGB]{185,224,194}+9.25\% & 0.0651 & 0.0455 & \cellcolor[RGB]{142,205,157}+30.08\% & 0.0813 & 0.0606 & \cellcolor[RGB]{150,208,164}+25.46\% & 0.53\% \\
TReS~\cite{golestaneh2022tres} & 0.6293 & 0.7109 & \cellcolor[RGB]{175,220,185}+12.96\% & 0.6398 & 0.7472 & \cellcolor[RGB]{166,216,178}+16.77\% & 0.1317 & 0.0444 & \cellcolor[RGB]{142,205,157}+66.27\% & 0.1470 & 0.0588 & \cellcolor[RGB]{142,205,157}+60.02\% & 0.74\% \\
PaQ-2-PiQ~\cite{ying2020paq} & 0.5772 & 0.6122 & \cellcolor[RGB]{195,229,203}+6.07\% & 0.6606 & 0.6894 & \cellcolor[RGB]{202,232,208}+4.36\% & 0.0441 & 0.0413 & \cellcolor[RGB]{194,228,202}+6.38\% & 0.0603 & 0.0572 & \cellcolor[RGB]{199,230,205}+5.21\% & 1.58\% \\
\bottomrule
\end{tabular*}
\par\vspace{3pt}
\begin{minipage}{\textwidth}\fontsize{7.5}{8.5}\selectfont
\textsuperscript{r}Retrained image assessment model.
\end{minipage}
\endgroup
\end{table*}
\begin{table*}[p]
\centering
\caption{\textbf{Complete score-refinement results, part 2 of 2 (48 configurations).} Base: frozen baseline assessment model. Ref.: the baseline model after \method refinement. Gain is the relative change from Base to Ref., computed by dividing the increase in SRCC/PLCC or the reduction in MAE/RMSE by the absolute Base value and expressing the result as a percentage. Params $\uparrow$: added trainable parameters relative to the complete frozen baseline (\%). Darker \captiongreen{green}/\captionred{red} indicates larger relative improvement/decline, respectively, with color intensity capped at a 30\% change. Improvements correspond to higher SRCC/PLCC or lower MAE/RMSE. Rows are sorted by refined SRCC within each dataset. Test/Total gives the default evaluation count and total sample count used in that dataset protocol. Data partitions and evaluation protocols are detailed in Appendices~\ref{app:data-protocols}-\ref{app:score-scales}.}
\label{tab:complete-results-2}
\begingroup
\fontsize{7.7}{8.2}\selectfont
\setlength{\tabcolsep}{1.3pt}
\renewcommand{\arraystretch}{0.90}
\begin{tabular*}{\textwidth}{@{\extracolsep{\fill}}l*{4}{rr>{\fontsize{7}{8.2}\selectfont}r}@{\extracolsep{0pt}\hspace{3pt}}r@{}}
\toprule
 & \multicolumn{3}{c}{SRCC {\fontsize{7.5}{9}\selectfont$\uparrow$}} & \multicolumn{3}{c}{PLCC {\fontsize{7.5}{9}\selectfont$\uparrow$}} & \multicolumn{3}{c}{MAE {\fontsize{7.5}{9}\selectfont$\downarrow$}} & \multicolumn{3}{c}{RMSE {\fontsize{7.5}{9}\selectfont$\downarrow$}} & \\
\cmidrule(lr){2-4}\cmidrule(lr){5-7}\cmidrule(lr){8-10}\cmidrule(lr){11-13}
Assessment model & \multicolumn{1}{c}{Base} & \multicolumn{1}{c}{Ref.} & \multicolumn{1}{c}{Gain} & \multicolumn{1}{c}{Base} & \multicolumn{1}{c}{Ref.} & \multicolumn{1}{c}{Gain} & \multicolumn{1}{c}{Base} & \multicolumn{1}{c}{Ref.} & \multicolumn{1}{c}{Gain} & \multicolumn{1}{c}{Base} & \multicolumn{1}{c}{Ref.} & \multicolumn{1}{c}{Gain} & Params {\fontsize{7.5}{9}\selectfont$\uparrow$} \\
\midrule
\addlinespace[1pt]\rowcolor[gray]{0.92}\multicolumn{14}{@{}l}{\textbf{AVA (IAA)}~\cite{murray2012ava} - Test/Total: 19,929 / 255,508} \\
FGAesQ~\cite{yang2026fgaesq} & 0.8720 & 0.8838 & \cellcolor[RGB]{218,239,222}+1.35\% & 0.8676 & 0.8791 & \cellcolor[RGB]{219,239,223}+1.32\% & 0.2885 & 0.2719 & \cellcolor[RGB]{197,229,204}+5.75\% & 0.3785 & 0.3593 & \cellcolor[RGB]{199,230,206}+5.07\% & 0.53\% \\
ELTA~\cite{liu2024elta} & 0.8420 & 0.8569 & \cellcolor[RGB]{215,238,220}+1.77\% & 0.8437 & 0.8581 & \cellcolor[RGB]{216,238,220}+1.71\% & 0.3110 & 0.2959 & \cellcolor[RGB]{200,231,207}+4.87\% & 0.4037 & 0.3864 & \cellcolor[RGB]{202,232,209}+4.27\% & 0.62\% \\
QPT V2 (IAA)\textsuperscript{r}~\cite{xie2024qpt} & 0.8551 & 0.8555 & \cellcolor[RGB]{235,246,237}+0.05\% & 0.8649 & 0.8658 & \cellcolor[RGB]{233,245,235}+0.11\% & 0.0307 & 0.0299 & \cellcolor[RGB]{211,235,216}+2.54\% & 0.0389 & 0.0381 & \cellcolor[RGB]{214,237,218}+2.03\% & 1.08\% \\
ArtiMuse~\cite{cao2026artimuse} & 0.8265 & 0.8261 & \cellcolor[RGB]{252,236,236}-0.05\% & 0.8250 & 0.8249 & \cellcolor[RGB]{253,238,238}-0.01\% & 0.3265 & 0.3263 & \cellcolor[RGB]{234,246,236}+0.07\% & 0.4249 & 0.4249 & \cellcolor[RGB]{253,239,239}-0.00\% & 0.01\% \\
Q-Align~\cite{wu2024qalign} & 0.8216 & 0.8240 & \cellcolor[RGB]{229,244,232}+0.29\% & 0.8170 & 0.8211 & \cellcolor[RGB]{226,242,229}+0.50\% & 0.3326 & 0.3291 & \cellcolor[RGB]{221,240,224}+1.05\% & 0.4334 & 0.4291 & \cellcolor[RGB]{221,240,225}+0.99\% & 0.04\% \\
MUSIQ~\cite{ke2021musiq} & 0.8177 & 0.8184 & \cellcolor[RGB]{234,246,236}+0.09\% & 0.8191 & 0.8196 & \cellcolor[RGB]{235,246,236}+0.05\% & 0.3368 & 0.3341 & \cellcolor[RGB]{223,241,226}+0.80\% & 0.4353 & 0.4315 & \cellcolor[RGB]{222,241,226}+0.88\% & 2.18\% \\
AesCLIP~\cite{sheng2023aesclip} & 0.8045 & 0.8091 & \cellcolor[RGB]{226,242,228}+0.58\% & 0.8119 & 0.8157 & \cellcolor[RGB]{227,243,230}+0.47\% & 0.3141 & 0.3084 & \cellcolor[RGB]{215,237,220}+1.81\% & 0.4061 & 0.3997 & \cellcolor[RGB]{217,238,221}+1.58\% & 0.69\% \\
TOPIQ (Swin)~\cite{chen2024topiq} & 0.7910 & 0.7941 & \cellcolor[RGB]{228,243,230}+0.39\% & 0.7901 & 0.7940 & \cellcolor[RGB]{227,243,229}+0.49\% & 0.3566 & 0.3524 & \cellcolor[RGB]{220,240,224}+1.17\% & 0.4614 & 0.4569 & \cellcolor[RGB]{222,240,225}+0.96\% & 0.36\% \\
Charm~\cite{behrad2025charm} & 0.7831 & 0.7934 & \cellcolor[RGB]{219,239,223}+1.32\% & 0.7840 & 0.7938 & \cellcolor[RGB]{219,239,223}+1.24\% & 0.3606 & 0.3517 & \cellcolor[RGB]{211,236,216}+2.47\% & 0.4682 & 0.4575 & \cellcolor[RGB]{212,236,217}+2.27\% & 0.52\% \\
MLSP~\cite{hosu2019mlsp} & 0.7525 & 0.7487 & \cellcolor[RGB]{251,228,228}-0.50\% & 0.7546 & 0.7513 & \cellcolor[RGB]{251,229,229}-0.44\% & 0.3830 & 0.3855 & \cellcolor[RGB]{250,227,227}-0.67\% & 0.4942 & 0.4969 & \cellcolor[RGB]{251,228,228}-0.53\% & 0.67\% \\
TOPIQ (R50)~\cite{chen2024topiq} & 0.7359 & 0.7415 & \cellcolor[RGB]{224,241,227}+0.76\% & 0.7371 & 0.7418 & \cellcolor[RGB]{225,242,228}+0.65\% & 0.3937 & 0.3905 & \cellcolor[RGB]{223,241,226}+0.81\% & 0.5080 & 0.5043 & \cellcolor[RGB]{224,241,227}+0.73\% & 0.60\% \\
NIMA (IR-v2)~\cite{talebi2018nima} & 0.7164 & 0.7216 & \cellcolor[RGB]{224,241,227}+0.73\% & 0.7185 & 0.7236 & \cellcolor[RGB]{224,241,227}+0.71\% & 0.4092 & 0.4044 & \cellcolor[RGB]{220,239,224}+1.17\% & 0.5264 & 0.5196 & \cellcolor[RGB]{219,239,223}+1.29\% & 1.11\% \\
NIMA (VGG-16)~\cite{talebi2018nima} & 0.6606 & 0.6612 & \cellcolor[RGB]{234,246,236}+0.08\% & 0.6640 & 0.6652 & \cellcolor[RGB]{231,245,233}+0.19\% & 0.4425 & 0.4398 & \cellcolor[RGB]{225,242,228}+0.61\% & 0.5660 & 0.5628 & \cellcolor[RGB]{226,242,229}+0.55\% & 2.34\% \\
\addlinespace[1pt]\rowcolor[gray]{0.92}\multicolumn{14}{@{}l}{\textbf{PARA GIAA (IAA)}~\cite{yang2022para} - Test/Total: 2,952 / 31,220} \\
Charm~\cite{behrad2025charm} & 0.8933 & 0.9085 & \cellcolor[RGB]{216,238,220}+1.71\% & 0.9294 & 0.9390 & \cellcolor[RGB]{221,240,225}+1.03\% & 0.0377 & 0.0351 & \cellcolor[RGB]{192,227,200}+6.90\% & 0.0492 & 0.0459 & \cellcolor[RGB]{193,228,201}+6.66\% & 2.74\% \\
PIAA-MIR (Swin-T)~\cite{chen2025individual} & 0.7236 & 0.8561 & \cellcolor[RGB]{163,214,175}+18.31\% & 0.7931 & 0.8977 & \cellcolor[RGB]{175,219,185}+13.18\% & 0.0702 & 0.0446 & \cellcolor[RGB]{142,205,157}+36.48\% & 0.0925 & 0.0589 & \cellcolor[RGB]{142,205,157}+36.35\% & 0.99\% \\
PIAA-ICI (Swin-T)~\cite{chen2025individual} & 0.7454 & 0.8524 & \cellcolor[RGB]{172,218,183}+14.35\% & 0.8243 & 0.8943 & \cellcolor[RGB]{187,225,196}+8.49\% & 0.0678 & 0.0447 & \cellcolor[RGB]{142,205,157}+34.00\% & 0.0859 & 0.0597 & \cellcolor[RGB]{142,205,157}+30.41\% & 0.99\% \\
PIAA-MIR (ViT-S)~\cite{chen2025individual} & 0.6076 & 0.8253 & \cellcolor[RGB]{142,205,157}+35.83\% & 0.6964 & 0.8747 & \cellcolor[RGB]{149,208,163}+25.61\% & 0.0908 & 0.0490 & \cellcolor[RGB]{142,205,157}+46.06\% & 0.1128 & 0.0647 & \cellcolor[RGB]{142,205,157}+42.66\% & 2.32\% \\
PIAA-ICI (ViT-S)~\cite{chen2025individual} & 0.5726 & 0.8220 & \cellcolor[RGB]{142,205,157}+43.56\% & 0.6777 & 0.8725 & \cellcolor[RGB]{144,206,159}+28.75\% & 0.0782 & 0.0491 & \cellcolor[RGB]{142,205,157}+37.22\% & 0.1002 & 0.0651 & \cellcolor[RGB]{142,205,157}+34.98\% & 2.34\% \\
\addlinespace[1pt]\rowcolor[gray]{0.92}\multicolumn{14}{@{}l}{\textbf{TAD66K (IAA)}~\cite{he2022tanet} - Test/Total: 14,072 / 66,327} \\
EAT~\cite{he2023eat} & 0.5170 & 0.5186 & \cellcolor[RGB]{229,244,232}+0.30\% & 0.5459 & 0.5467 & \cellcolor[RGB]{232,245,234}+0.14\% & 0.1022 & 0.1012 & \cellcolor[RGB]{222,240,225}+0.96\% & 0.1232 & 0.1229 & \cellcolor[RGB]{230,244,232}+0.27\% & 0.43\% \\
TANet~\cite{he2022tanet} & 0.5062 & 0.5084 & \cellcolor[RGB]{227,243,230}+0.43\% & 0.5260 & 0.5350 & \cellcolor[RGB]{216,238,220}+1.70\% & 0.1060 & 0.1033 & \cellcolor[RGB]{211,235,216}+2.57\% & 0.1254 & 0.1234 & \cellcolor[RGB]{217,238,221}+1.53\% & 1.01\% \\
QUASAR~\cite{kastryulin2024quasar} & 0.3480 & 0.4846 & \cellcolor[RGB]{142,205,157}+39.23\% & 0.3720 & 0.5124 & \cellcolor[RGB]{142,205,157}+37.75\% & 0.1162 & 0.1038 & \cellcolor[RGB]{181,222,191}+10.62\% & 0.1388 & 0.1246 & \cellcolor[RGB]{182,223,192}+10.21\% & 0.53\% \\
Charm~\cite{behrad2025charm} & 0.4503 & 0.4588 & \cellcolor[RGB]{215,237,219}+1.88\% & 0.4807 & 0.4898 & \cellcolor[RGB]{215,237,219}+1.89\% & 1.0698 & 1.0592 & \cellcolor[RGB]{221,240,225}+0.99\% & 1.2763 & 1.2784 & \cellcolor[RGB]{252,233,233}-0.17\% & 2.74\% \\
\addlinespace[1pt]\rowcolor[gray]{0.92}\multicolumn{14}{@{}l}{\textbf{AADB (IAA)}~\cite{kong2016aadb} - Test/Total: 1,000 / 9,830} \\
QUASAR~\cite{kastryulin2024quasar} & 0.4844 & 0.7690 & \cellcolor[RGB]{142,205,157}+58.76\% & 0.4921 & 0.7686 & \cellcolor[RGB]{142,205,157}+56.18\% & 0.1620 & 0.1054 & \cellcolor[RGB]{142,205,157}+34.92\% & 0.1966 & 0.1308 & \cellcolor[RGB]{142,205,157}+33.48\% & 0.53\% \\
Charm~\cite{behrad2025charm} & 0.7555 & 0.7612 & \cellcolor[RGB]{224,241,227}+0.74\% & 0.7665 & 0.7673 & \cellcolor[RGB]{233,245,235}+0.11\% & 0.1038 & 0.1045 & \cellcolor[RGB]{250,227,227}-0.69\% & 0.1308 & 0.1306 & \cellcolor[RGB]{234,246,235}+0.09\% & 2.74\% \\
TAVAR\textsuperscript{r}~\cite{li2023tavar} & 0.7102 & 0.7066 & \cellcolor[RGB]{251,228,228}-0.50\% & 0.7103 & 0.7096 & \cellcolor[RGB]{252,235,235}-0.09\% & 0.1244 & 0.1157 & \cellcolor[RGB]{192,227,200}+7.01\% & 0.1544 & 0.1434 & \cellcolor[RGB]{192,227,200}+7.14\% & 6.98\% \\
\addlinespace[1pt]\rowcolor[gray]{0.92}\multicolumn{14}{@{}l}{\textbf{Flickr-AES (IAA)}~\cite{ren2017personalized} - Test/Total: 4,737 / 40,499} \\
EAT~\cite{he2023eat} & 0.7324 & 0.8179 & \cellcolor[RGB]{179,221,188}+11.67\% & 0.7420 & 0.8394 & \cellcolor[RGB]{175,220,185}+13.13\% & 0.1246 & 0.0680 & \cellcolor[RGB]{142,205,157}+45.42\% & 0.1548 & 0.0872 & \cellcolor[RGB]{142,205,157}+43.67\% & 0.43\% \\
PAM~\cite{ren2017personalized} & 0.7073 & 0.7384 & \cellcolor[RGB]{202,232,208}+4.40\% & 0.7378 & 0.7675 & \cellcolor[RGB]{203,232,210}+4.03\% & 0.0873 & 0.0783 & \cellcolor[RGB]{182,223,191}+10.38\% & 0.1103 & 0.0989 & \cellcolor[RGB]{182,223,191}+10.40\% & 2.80\% \\
BLG-PIAA (ResNet-18)~\cite{zhu2022blgpiaa} & 0.6979 & 0.7236 & \cellcolor[RGB]{205,233,211}+3.68\% & 0.7208 & 0.7515 & \cellcolor[RGB]{202,232,209}+4.25\% & 0.0873 & 0.0803 & \cellcolor[RGB]{189,226,197}+8.02\% & 0.1109 & 0.1018 & \cellcolor[RGB]{188,226,197}+8.21\% & 1.40\% \\
\addlinespace[1pt]\rowcolor[gray]{0.92}\multicolumn{14}{@{}l}{\textbf{BAID (IAA)}~\cite{yi2023saan} - Test/Total: 5,625 / 60,337} \\
Charm~\cite{behrad2025charm} & 0.3731 & 0.3778 & \cellcolor[RGB]{219,239,223}+1.27\% & 0.3833 & 0.3944 & \cellcolor[RGB]{209,235,214}+2.88\% & 0.0463 & 0.0723 & \cellcolor[RGB]{235,151,151}-56.17\% & 0.0633 & 0.0880 & \cellcolor[RGB]{235,151,151}-38.94\% & 2.74\% \\
\addlinespace[1pt]\rowcolor[gray]{0.92}\multicolumn{14}{@{}l}{\textbf{UHD-IQA (IQA)}~\cite{hosu2024uhd} - Test/Total: 900 / 6,073} \\
UIQA~\cite{sun2025uiqa} & 0.8522 & 0.8448 & \cellcolor[RGB]{250,225,225}-0.86\% & 0.8014 & 0.7918 & \cellcolor[RGB]{249,222,222}-1.20\% & 0.0414 & 0.0416 & \cellcolor[RGB]{251,228,228}-0.51\% & 0.0624 & 0.0621 & \cellcolor[RGB]{227,242,229}+0.49\% & 1.00\% \\
LAR-IQA (MLP)~\cite{avanaki2024lar} & 0.8094 & 0.8185 & \cellcolor[RGB]{220,240,224}+1.12\% & 0.7744 & 0.7767 & \cellcolor[RGB]{229,244,232}+0.30\% & 0.0416 & 0.0473 & \cellcolor[RGB]{241,180,180}-13.64\% & 0.0576 & 0.0673 & \cellcolor[RGB]{239,173,173}-16.97\% & 1.29\% \\
LAR-IQA (KAN)~\cite{avanaki2024lar} & 0.7514 & 0.6497 & \cellcolor[RGB]{241,180,180}-13.54\% & 0.7314 & 0.6319 & \cellcolor[RGB]{241,180,180}-13.61\% & 0.0476 & 0.0540 & \cellcolor[RGB]{241,180,180}-13.53\% & 0.0633 & 0.0739 & \cellcolor[RGB]{240,173,173}-16.85\% & 1.24\% \\
GLUQ\textsuperscript{r}~\cite{zhu2026gluq} & 0.1666 & 0.2714 & \cellcolor[RGB]{142,205,157}+62.89\% & 0.1446 & 0.2710 & \cellcolor[RGB]{142,205,157}+87.49\% & 0.0695 & 0.0690 & \cellcolor[RGB]{224,241,227}+0.72\% & 0.0911 & 0.0912 & \cellcolor[RGB]{252,234,234}-0.14\% & 1.73\% \\
\addlinespace[1pt]\rowcolor[gray]{0.92}\multicolumn{14}{@{}l}{\textbf{AGIQA3K (IQA)}~\cite{li2023agiqa} - Test/Total: 605 / 2,982} \\
IP-IQA\textsuperscript{r}~\cite{qu2024ipiqa} & 0.8117 & 0.8202 & \cellcolor[RGB]{221,240,224}+1.05\% & 0.8555 & 0.8715 & \cellcolor[RGB]{215,237,219}+1.87\% & 0.0844 & 0.0791 & \cellcolor[RGB]{195,228,202}+6.26\% & 0.1067 & 0.1012 & \cellcolor[RGB]{199,230,206}+5.16\% & 0.62\% \\
AMFF-Net\textsuperscript{r}~\cite{zhou2024amff} & 0.5642 & 0.5815 & \cellcolor[RGB]{208,234,213}+3.06\% & 0.6233 & 0.6346 & \cellcolor[RGB]{215,237,220}+1.80\% & 0.1288 & 0.1218 & \cellcolor[RGB]{198,230,205}+5.44\% & 0.1602 & 0.1578 & \cellcolor[RGB]{217,238,221}+1.54\% & 0.38\% \\
\addlinespace[1pt]\rowcolor[gray]{0.92}\multicolumn{14}{@{}l}{\textbf{TID2013 (IQA)}~\cite{ponomarenko2015tid} - Test/Total: 600 / 3,000} \\
NLNet~\cite{jia2022nlnet} & 0.8322 & 0.8449 & \cellcolor[RGB]{217,238,221}+1.53\% & 0.8500 & 0.8644 & \cellcolor[RGB]{216,238,220}+1.69\% & 0.0609 & 0.0566 & \cellcolor[RGB]{192,227,200}+7.01\% & 0.0828 & 0.0792 & \cellcolor[RGB]{202,232,209}+4.29\% & 1.34\% \\
RankIQA~\cite{liu2017rankiqa} & 0.8422 & 0.7639 & \cellcolor[RGB]{243,190,190}-9.30\% & 0.8447 & 0.7939 & \cellcolor[RGB]{245,200,200}-6.02\% & 0.1319 & 0.0686 & \cellcolor[RGB]{142,205,157}+48.00\% & 0.1494 & 0.0920 & \cellcolor[RGB]{142,205,157}+38.43\% & 0.65\% \\
NIMA (MobileNet, technical)~\cite{talebi2018nima} & 0.5064 & 0.5385 & \cellcolor[RGB]{194,228,202}+6.34\% & 0.5605 & 0.5739 & \cellcolor[RGB]{212,236,217}+2.39\% & 0.9421 & 1.1015 & \cellcolor[RGB]{239,173,173}-16.92\% & 1.1654 & 1.3637 & \cellcolor[RGB]{239,173,173}-17.02\% & 13.02\% \\
\addlinespace[1pt]\rowcolor[gray]{0.92}\multicolumn{14}{@{}l}{\textbf{LIVE (IQA)}~\cite{sheikh2006live} - Test/Total: 160 / 779} \\
DBCNN~\cite{zhang2020dbcnn} & 0.9402 & 0.9761 & \cellcolor[RGB]{204,233,210}+3.82\% & 0.9354 & 0.9774 & \cellcolor[RGB]{201,231,208}+4.49\% & 0.1119 & 0.0499 & \cellcolor[RGB]{142,205,157}+55.40\% & 0.1447 & 0.0615 & \cellcolor[RGB]{142,205,157}+57.50\% & 83.31\% \\
RankIQA~\cite{liu2017rankiqa} & 0.9410 & 0.9744 & \cellcolor[RGB]{206,233,211}+3.55\% & 0.9368 & 0.9778 & \cellcolor[RGB]{202,232,208}+4.37\% & 0.1057 & 0.0446 & \cellcolor[RGB]{142,205,157}+57.81\% & 0.1388 & 0.0598 & \cellcolor[RGB]{142,205,157}+56.92\% & 0.65\% \\
NLNet~\cite{jia2022nlnet} & 0.9606 & 0.9691 & \cellcolor[RGB]{222,241,226}+0.88\% & 0.9634 & 0.9747 & \cellcolor[RGB]{220,239,224}+1.18\% & 0.0669 & 0.0503 & \cellcolor[RGB]{151,209,165}+24.71\% & 0.0878 & 0.0663 & \cellcolor[RGB]{151,209,165}+24.54\% & 1.34\% \\
\addlinespace[1pt]\rowcolor[gray]{0.92}\multicolumn{14}{@{}l}{\textbf{LIVE-MD (IQA)}~\cite{jayaraman2012livemd} - Test/Total: 90 / 450} \\
DBCNN~\cite{zhang2020dbcnn} & 0.9634 & 0.9637 & \cellcolor[RGB]{236,247,238}+0.03\% & 0.9603 & 0.9624 & \cellcolor[RGB]{231,244,233}+0.22\% & 0.0412 & 0.0375 & \cellcolor[RGB]{186,224,195}+8.97\% & 0.0530 & 0.0511 & \cellcolor[RGB]{205,233,211}+3.63\% & 83.31\% \\
\addlinespace[1pt]\rowcolor[gray]{0.92}\multicolumn{14}{@{}l}{\textbf{CSIQ (IQA)}~\cite{larson2010csiq} - Test/Total: 173 / 866} \\
NLNet~\cite{jia2022nlnet} & 0.9302 & 0.9294 & \cellcolor[RGB]{252,235,235}-0.08\% & 0.9387 & 0.9377 & \cellcolor[RGB]{252,235,235}-0.10\% & 0.0884 & 0.0878 & \cellcolor[RGB]{225,242,228}+0.60\% & 0.1194 & 0.1185 & \cellcolor[RGB]{223,241,227}+0.77\% & 1.34\% \\
\addlinespace[1pt]\rowcolor[gray]{0.92}\multicolumn{14}{@{}l}{\textbf{CVIQ (IQA)}~\cite{sun2018cviq} - Test/Total: 99 / 528} \\
Assessor360~\cite{wu2023assessor360} & 0.9761 & 0.9772 & \cellcolor[RGB]{233,245,235}+0.11\% & 0.9875 & 0.9864 & \cellcolor[RGB]{252,235,235}-0.11\% & 0.0299 & 0.0311 & \cellcolor[RGB]{246,207,207}-4.13\% & 0.0396 & 0.0410 & \cellcolor[RGB]{247,210,210}-3.51\% & 0.41\% \\
IQCaption360\textsuperscript{r}~\cite{yan2025iqcaption} & -0.2215 & -0.1690 & \cellcolor[RGB]{153,210,166}+23.68\% & -0.1777 & -0.1479 & \cellcolor[RGB]{166,216,178}+16.79\% & 0.4621 & 0.2014 & \cellcolor[RGB]{142,205,157}+56.42\% & 0.6035 & 0.2454 & \cellcolor[RGB]{142,205,157}+59.34\% & 2.08\% \\
\addlinespace[1pt]\rowcolor[gray]{0.92}\multicolumn{14}{@{}l}{\textbf{FGRestore (IQA)}~\cite{sheng2026fgresq} - Test/Total: 3,706 / 18,406} \\
FGResQ~\cite{sheng2026fgresq} & 0.4118 & 0.7847 & \cellcolor[RGB]{142,205,157}+90.55\% & 0.4070 & 0.8146 & \cellcolor[RGB]{142,205,157}+100.15\% & 0.1767 & 0.0827 & \cellcolor[RGB]{142,205,157}+53.20\% & 0.2120 & 0.1073 & \cellcolor[RGB]{142,205,157}+49.39\% & 1.28\% \\
\addlinespace[1pt]\rowcolor[gray]{0.92}\multicolumn{14}{@{}l}{\textbf{GFIQA (IQA)}~\cite{su2023gfiqa} - Test/Total: 4,001 / 19,998} \\
DSL-FIQA~\cite{chen2024dslfiqa} & 0.9508 & 0.9744 & \cellcolor[RGB]{211,236,216}+2.48\% & 0.9517 & 0.9753 & \cellcolor[RGB]{211,236,216}+2.48\% & 0.2015 & 0.0320 & \cellcolor[RGB]{142,205,157}+84.14\% & 0.2198 & 0.0411 & \cellcolor[RGB]{142,205,157}+81.29\% & 0.14\% \\
\addlinespace[1pt]\rowcolor[gray]{0.92}\multicolumn{14}{@{}l}{\textbf{CGFIQA (IQA)}~\cite{chen2024dslfiqa} - Test/Total: 7,855 / 39,312} \\
DSL-FIQA~\cite{chen2024dslfiqa} & 0.9843 & 0.9870 & \cellcolor[RGB]{230,244,232}+0.28\% & 0.9840 & 0.9877 & \cellcolor[RGB]{228,243,231}+0.37\% & 0.0837 & 0.0249 & \cellcolor[RGB]{142,205,157}+70.26\% & 0.0931 & 0.0323 & \cellcolor[RGB]{142,205,157}+65.25\% & 0.14\% \\
\bottomrule
\end{tabular*}
\par\vspace{3pt}
\begin{minipage}{\textwidth}\fontsize{7.5}{8.5}\selectfont
\textsuperscript{r}Retrained image assessment model.
\end{minipage}
\endgroup
\end{table*}
\noindent\textbf{Score Refinement.} \Cref{tab:complete-results-1,tab:complete-results-2} report original/refined metrics and relative gains for all 100 configurations. Equally averaging these configurations gives mean gains in PLCC, SRCC, MAE, and RMSE of 7.55\%, 7.10\%, 23.27\%, and 21.93\%, respectively, with medians of 1.70\%, 1.20\%, 12.80\%, and 10.01\%. The respective improvement counts are 88 (88.00\%), 85 (85.00\%), 90 (90.00\%), and 89 (89.00\%). Both correlations improve in 84 configurations (84.00\%), and all four metrics improve in 77 (77.00\%). \Cref{fig:qualitative-main} illustrates score refinement, with more qualitative comparisons in Appendix~\ref{app:qualitative-examples}.

\noindent\textbf{Model Size and Computation.} For joint-input dimension $D$, the three residual heads of hidden width 64 add $3(64D+129)$ trainable parameters and approximately $3(64D+64)$ dense multiply-accumulate operations per image, plus pooling and normalization. Across 100 configurations, the heads add 1.64M parameters on average. Relative overhead has a mean of 7.01\% and a median of 1.11\%, with 90 configurations (90.00\%) below 5\% and a range of 0.009\%--225.87\%. \Cref{tab:complete-results-1,tab:complete-results-2} report parameter overhead for each baseline.

\subsection{Ablation Study}
\noindent\textbf{Ablation Setup.} \Cref{tab:mean-ablation-90}(A-G) compares variants with complete \method (Full) across 21 datasets. Per-configuration changes are divided by the absolute Full metric and averaged equally, with error reductions counted positively. Component rows use three heads. Mean per-image dense head multiply-accumulate operations (MACs) exclude baseline computation, pooling, and normalization. Appendix~\ref{app:component-ablations} provides individual cases and per-metric counts. (A) \textit{Multi-Level Visual Features.} We remove all visual features, retain final features alone, or omit final, early, middle, or late features. Feature groups are defined in Appendix~\ref{app:feature-groups}. (B) \textit{Geometric Context and Score Pathways.} We remove geometry, the original-score input to the residual head, or both inputs while retaining the score skip connection. We also remove the skip connection alone or together with the head's score input, so that the heads predict scores directly. Frozen representations, score mappings, and the training objective remain fixed (Appendix~\ref{app:score-pathways}). (C) \textit{Training Objective.} We remove the correlation or residual-penalty term. (D) \textit{Regularization.} We remove hidden dropout or add feature-group dropout of 0.4. (E) \textit{Number of Residual Heads.} We compare $K=1,3,5,10$ (Appendix~\ref{app:head-count}). Single-head entries average five-seed metrics, whereas ensembles average predictions. (F) \textit{Residual-Head Architecture.} We compare linear heads with one- and two-hidden-layer multilayer perceptrons (MLPs). (G) \textit{Correction Bound.} We compare five fixed bounds and one adaptive bound on the refinement score scale (definitions in Appendix~\ref{app:correction-bounds}).

\begin{table}[!t]
\centering
\caption{\textbf{Ablation study result.} Full denotes the complete MetaRefine configuration. Relative improvements over Full are averaged equally. Positive values indicate higher correlation or lower error. \captionred{Red}/\captiongreen{green} denotes negative/positive changes.}
\label{tab:mean-ablation-90}
\begingroup\fontsize{7.0}{8.6}\selectfont
\setlength{\tabcolsep}{0.8pt}
\renewcommand{\arraystretch}{0.90}
\begin{tabular*}{\columnwidth}{@{\extracolsep{\fill}}lrrrr@{}}
\toprule
Variant & \multicolumn{1}{c}{PLCC} & \multicolumn{1}{c}{SRCC} & \multicolumn{1}{c}{MAE} & \multicolumn{1}{c@{}}{RMSE} \\
\midrule
\rowcolor[gray]{0.92}\multicolumn{5}{@{}l}{\textbf{Reference Configuration}} \\
Full & 0.000\% & 0.000\% & 0.000\% & 0.000\% \\
\rowcolor[gray]{0.92}\multicolumn{5}{@{}l}{\textbf{(A) Multi-Level Visual Features}} \\
w/o visual features & \cellcolor[RGB]{247,203,203}-3.638\% & \cellcolor[RGB]{247,202,202}-3.889\% & \cellcolor[RGB]{235,151,151}-55.656\% & \cellcolor[RGB]{235,151,151}-48.445\% \\
Final features only & \cellcolor[RGB]{250,217,217}-0.960\% & \cellcolor[RGB]{249,215,215}-1.265\% & \cellcolor[RGB]{245,197,197}-5.371\% & \cellcolor[RGB]{245,197,197}-5.436\% \\
w/o final features & \cellcolor[RGB]{251,223,223}-0.285\% & \cellcolor[RGB]{226,242,229}+0.267\% & \cellcolor[RGB]{244,192,192}-7.129\% & \cellcolor[RGB]{244,193,193}-6.751\% \\
w/o early features & \cellcolor[RGB]{251,224,224}-0.261\% & \cellcolor[RGB]{251,223,223}-0.288\% & \cellcolor[RGB]{252,226,226}-0.097\% & \cellcolor[RGB]{252,225,225}-0.177\% \\
w/o middle features & \cellcolor[RGB]{251,223,223}-0.309\% & \cellcolor[RGB]{251,222,222}-0.353\% & \cellcolor[RGB]{252,225,225}-0.174\% & \cellcolor[RGB]{251,224,224}-0.221\% \\
w/o late features & \cellcolor[RGB]{229,243,232}+0.123\% & \cellcolor[RGB]{251,224,224}-0.262\% & \cellcolor[RGB]{251,221,221}-0.458\% & \cellcolor[RGB]{250,220,220}-0.603\% \\
\rowcolor[gray]{0.92}\multicolumn{5}{@{}l}{\textbf{(B) Geometric Context and Score Pathways}} \\
w/o geometric metadata & \cellcolor[RGB]{252,229,229}-0.026\% & \cellcolor[RGB]{252,228,228}-0.033\% & \cellcolor[RGB]{252,228,228}-0.057\% & \cellcolor[RGB]{252,228,228}-0.041\% \\
w/o score input to head & \cellcolor[RGB]{252,228,228}-0.037\% & \cellcolor[RGB]{252,228,228}-0.035\% & \cellcolor[RGB]{235,246,237}+0.001\% & \cellcolor[RGB]{234,246,236}+0.004\% \\
w/o geometric metadata \& score input & \cellcolor[RGB]{252,227,227}-0.065\% & \cellcolor[RGB]{252,227,227}-0.061\% & \cellcolor[RGB]{252,227,227}-0.068\% & \cellcolor[RGB]{252,228,228}-0.053\% \\
w/o score skip & \cellcolor[RGB]{240,173,173}-15.751\% & \cellcolor[RGB]{241,176,176}-14.277\% & \cellcolor[RGB]{235,151,151}-170.260\% & \cellcolor[RGB]{235,151,151}-154.029\% \\
w/o score input \& skip & \cellcolor[RGB]{240,173,173}-15.723\% & \cellcolor[RGB]{241,176,176}-14.359\% & \cellcolor[RGB]{235,151,151}-171.385\% & \cellcolor[RGB]{235,151,151}-154.958\% \\
\rowcolor[gray]{0.92}\multicolumn{5}{@{}l}{\textbf{(C) Training Objective}} \\
w/o correlation loss & \cellcolor[RGB]{250,219,219}-0.622\% & \cellcolor[RGB]{250,219,219}-0.630\% & \cellcolor[RGB]{231,244,234}+0.048\% & \cellcolor[RGB]{234,245,236}+0.007\% \\
w/o residual penalty & \cellcolor[RGB]{230,244,233}+0.077\% & \cellcolor[RGB]{252,226,226}-0.121\% & \cellcolor[RGB]{228,243,231}+0.147\% & \cellcolor[RGB]{229,243,232}+0.132\% \\
\rowcolor[gray]{0.92}\multicolumn{5}{@{}l}{\textbf{(D) Regularization}} \\
w/o hidden dropout & \cellcolor[RGB]{223,241,226}+0.533\% & \cellcolor[RGB]{230,244,233}+0.085\% & \cellcolor[RGB]{251,223,223}-0.313\% & \cellcolor[RGB]{251,223,223}-0.325\% \\
Feature-group dropout (0.4) & \cellcolor[RGB]{252,227,227}-0.075\% & \cellcolor[RGB]{251,223,223}-0.283\% & \cellcolor[RGB]{249,215,215}-1.185\% & \cellcolor[RGB]{250,216,216}-1.088\% \\
\rowcolor[gray]{0.92}\multicolumn{5}{@{}l}{\textbf{(E) Number of Residual Heads}} \\
1 head (0.57M MACs) & \cellcolor[RGB]{252,226,226}-0.099\% & \cellcolor[RGB]{233,245,235}+0.015\% & \cellcolor[RGB]{252,225,225}-0.187\% & \cellcolor[RGB]{252,224,224}-0.199\% \\
3 heads (1.70M MACs, default) & 0.000\% & 0.000\% & 0.000\% & 0.000\% \\
5 heads (2.84M MACs) & \cellcolor[RGB]{252,228,228}-0.038\% & \cellcolor[RGB]{228,243,231}+0.179\% & \cellcolor[RGB]{233,245,235}+0.011\% & \cellcolor[RGB]{253,230,230}-0.008\% \\
10 heads (5.67M MACs) & \cellcolor[RGB]{231,244,234}+0.043\% & \cellcolor[RGB]{226,242,229}+0.272\% & \cellcolor[RGB]{231,244,234}+0.046\% & \cellcolor[RGB]{232,245,234}+0.034\% \\
\rowcolor[gray]{0.92}\multicolumn{5}{@{}l}{\textbf{(F) Residual-Head Architecture}} \\
Linear (0.03M MACs) & \cellcolor[RGB]{221,240,225}+0.727\% & \cellcolor[RGB]{216,238,221}+1.228\% & \cellcolor[RGB]{235,151,151}-37.458\% & \cellcolor[RGB]{235,152,152}-29.405\% \\
MLP (32, 0.85M MACs) & \cellcolor[RGB]{252,227,227}-0.069\% & \cellcolor[RGB]{226,242,229}+0.291\% & \cellcolor[RGB]{249,215,215}-1.158\% & \cellcolor[RGB]{250,216,216}-1.089\% \\
MLP (64, 1.70M MACs, default) & 0.000\% & 0.000\% & 0.000\% & 0.000\% \\
MLP (128, 3.40M MACs) & \cellcolor[RGB]{225,242,229}+0.337\% & \cellcolor[RGB]{227,242,230}+0.228\% & \cellcolor[RGB]{224,241,227}+0.450\% & \cellcolor[RGB]{224,241,228}+0.383\% \\
MLP (64--64, 1.71M MACs) & \cellcolor[RGB]{227,242,230}+0.243\% & \cellcolor[RGB]{220,240,224}+0.754\% & \cellcolor[RGB]{248,207,207}-2.766\% & \cellcolor[RGB]{248,208,208}-2.401\% \\
\rowcolor[gray]{0.92}\multicolumn{5}{@{}l}{\textbf{(G) Correction Bound}} \\
No bound (default) & 0.000\% & 0.000\% & 0.000\% & 0.000\% \\
0.1875 & \cellcolor[RGB]{248,209,209}-2.248\% & \cellcolor[RGB]{248,209,209}-2.294\% & \cellcolor[RGB]{235,151,151}-54.968\% & \cellcolor[RGB]{235,151,151}-48.182\% \\
0.375 & \cellcolor[RGB]{249,214,214}-1.384\% & \cellcolor[RGB]{249,214,214}-1.347\% & \cellcolor[RGB]{235,151,151}-45.274\% & \cellcolor[RGB]{235,151,151}-40.412\% \\
0.75 & \cellcolor[RGB]{250,219,219}-0.645\% & \cellcolor[RGB]{250,218,218}-0.752\% & \cellcolor[RGB]{235,151,151}-31.736\% & \cellcolor[RGB]{235,152,152}-29.118\% \\
1.5 & \cellcolor[RGB]{223,241,227}+0.483\% & \cellcolor[RGB]{223,241,226}+0.528\% & \cellcolor[RGB]{240,171,171}-16.758\% & \cellcolor[RGB]{240,172,172}-16.252\% \\
3.375 & \cellcolor[RGB]{211,235,216}+2.044\% & \cellcolor[RGB]{213,236,218}+1.725\% & \cellcolor[RGB]{244,192,192}-7.125\% & \cellcolor[RGB]{244,191,191}-7.511\% \\
Adaptive (90th percentile) & \cellcolor[RGB]{211,235,216}+2.025\% & \cellcolor[RGB]{213,236,218}+1.727\% & \cellcolor[RGB]{244,191,191}-7.436\% & \cellcolor[RGB]{244,190,190}-7.846\% \\
\bottomrule
\end{tabular*}
\endgroup
\end{table}

\noindent\textbf{Visual Evidence and Score Refinement.} Removing visual features substantially reduces performance. Removing geometric metadata or the head's original-score input also reduces performance. Removing the score skip connection reduces PLCC and SRCC by 15.75\% and 14.28\% on average and increases MAE and RMSE in every ablation configuration. Removing both score pathways gives similar results. Under the shared 85-update budget, these comparisons support learning corrections to the existing prediction and distinguish the skip connection's contribution from that of the score input within the head.

\noindent\textbf{Stability and Uncertainty.} We assess training stability through repeated training with ten random seeds and quantify uncertainty in ten- versus three-head performance using paired confidence intervals (Appendix~\ref{app:stability-uncertainty}).

\noindent\textbf{Efficiency.} We measure inference latency, memory usage, and training and feature-caching costs on an RTX~3090 (Appendices~\ref{app:efficiency-pipeline}-\ref{app:training-cache}).

\section{Conclusion}
\label{sec:conclusion}
We presented \method, a lightweight score refinement framework for pretrained image assessment models. It combines multi-level visual features and original scores with geometric metadata as auxiliary context, learning residual corrections while preserving the original scoring pipeline. Only the residual heads are trained, enabling the same refinement design to accommodate diverse assessment architectures. Across 100 configurations on 21 datasets, PLCC and SRCC increase by 7.55\% and 7.10\% on average, while MAE and RMSE decrease by 23.27\% and 21.93\%, respectively. These relative gains include further improvements to state-of-the-art IQA and IAA models, with a median parameter overhead of only 1.11\%. Ablations highlight the contributions of multi-level visual features, geometric metadata, and original-score pathways. These findings show that \method effectively reuses representations already learned by assessment models to further improve score prediction.
\clearpage
{\small
\setlength{\bibsep}{0pt}
\renewcommand{\bibfont}{\fontsize{9}{10.5}\selectfont}
\bibliographystyle{ieeenat_fullname}
\bibliography{references}
}
\clearpage
\appendix
\setcounter{table}{0}
\setcounter{figure}{0}
\renewcommand{\theHtable}{appendix.\arabic{table}}
\renewcommand{\theHfigure}{appendix.\arabic{figure}}
\twocolumn[{\centering\Large\bfseries Appendix\par\vspace{14pt}}]
\setcounter{topnumber}{4}
\setcounter{bottomnumber}{4}
\setcounter{totalnumber}{8}
\setcounter{dbltopnumber}{4}
\renewcommand{\topfraction}{0.92}
\renewcommand{\bottomfraction}{0.92}
\renewcommand{\dbltopfraction}{0.92}
\renewcommand{\textfraction}{0.08}
\renewcommand{\floatpagefraction}{0.9}
\renewcommand{\dblfloatpagefraction}{0.9}
\setlength{\dbltextfloatsep}{16pt plus 2pt minus 4pt}

\renewcommand{\thetable}{S\arabic{table}}
\renewcommand{\thefigure}{S\arabic{figure}}
\numberwithin{equation}{section}
\section{Further Method Details}
\label{app:method-details}
This section provides the aggregation and normalization conventions, residual-head training details, loss-function definitions, and supporting analyses for the method described in the main paper. Numerical settings and model-specific extraction locations are reported in Appendices~\ref{app:optimization}-\ref{app:feature-extraction}.

\subsection{Feature Aggregation}
\label{app:aggregation}
The fixed aggregation operator $\pi_\ell$ converts the visual representation at each selected extraction location into a feature vector. Spatial average pooling converts convolutional feature maps into feature vectors. For Transformers, we use average-pooled token features and class-token features when available. Aggregation may also use the model's frozen normalization or projection layers and combine its original crops, scales, or viewports. The final representation $\mathbf v_{\mathrm{out}}$ uses the corresponding image-level aggregation. These operations retain the model's original scoring computation.

\subsection{Score Scales and Input Normalization}
\label{app:normalization}
The fixed affine map $\mathcal T(s)=as+b$ specifies the refinement score scale in which residual corrections are learned. When no score conversion is needed, we set $a=1$ and $b=0$.

For the joint input $\mathbf u$, each coordinate $j$ has training-set mean $\mu_j$ and population standard deviation $\sigma_j$. A lower bound $\sigma_{\min}>0$ handles constant or nearly constant inputs, and a clipping threshold $z_{\max}>0$ limits standardized values. The normalization operator is:
\begin{equation}
 z_j=\clip\!\left(
 \frac{u_j-\mu_j}{\max(\sigma_j,\sigma_{\min})},
 -z_{\max},z_{\max}\right).
 \label{eq:normalization-detail}
\end{equation}
The stored training statistics are reused at inference. This normalization balances the scales of visual activations, geometric features, and scores before they enter the residual heads. Appendix~\ref{app:input-statistics} specifies the numerical thresholds and precision used to compute and store these statistics.

\subsection{Pair Sampling and Loss Functions}
\label{app:loss-definitions}
We sample pairs of distinct training images with a minimum label separation. The retained index pairs form the multiset $\mathcal P$. Its size $|\mathcal P|$ counts repeated pairs once per occurrence, and each occurrence receives equal weight in the pairwise objective. Appendix~\ref{app:pair-sampling} specifies the sampling budget, filtering threshold, and pair-resampling schedule.

The soft preference targets and predictions use the logistic sigmoid $\sigma(t)=1/(1+\exp(-t))$. Binary cross-entropy is defined as $\BCE(p,\hat p)=-p\log\hat p-(1-p)\log(1-\hat p)$. The Huber loss with transition $\kappa>0$ is defined as $H_\kappa(e)=e^2/2$ for $|e|\leq\kappa$ and $H_\kappa(e)=\kappa(|e|-\kappa/2)$ otherwise. The main paper uses $H_\kappa(e)/\kappa$ in the regression term. Numerical loss settings are reported in Appendix~\ref{app:loss-settings}.

\subsection{Residual-Correction Interpretation}
\label{app:residual-interpretation}
Let $r=\tilde y-q_0$ denote the frozen assessment model's prediction residual on the refinement score scale. With expectations taken over the image and ground-truth score distribution, the unconstrained correction that minimizes expected squared error over functions of $\mathbf z$ is $\delta^*(\mathbf z)=\E[r\mid\mathbf z]$, giving:
\begin{equation}
 \E[r^2]-\E[(r-\delta^*(\mathbf z))^2]
 =\E[\delta^*(\mathbf z)^2].
 \label{eq:residual_identity}
\end{equation}
Given $\mathbf z$, $r-\delta^*(\mathbf z)$ has zero conditional mean. Therefore, its expected cross term with $\delta^*(\mathbf z)$ vanishes, establishing the identity. This identity describes the population squared-error reduction associated with predictable residual variation. Geometric metadata can help when it explains residual variation beyond the visual features and original score.

\subsection{Trainable Parameter Count}
\label{app:parameter-count}
For input dimension $D$ and hidden width $d$, each residual head contains $Dd+d$ parameters in its input layer and $d+1$ parameters in its output layer. The added trainable parameter count for $K$ heads is:
\begin{equation}
 N_{\mathrm{add}}=K\left[d(D+1)+(d+1)\right].
 \label{eq:parameters}
\end{equation}
For a fixed number of residual heads, $N_{\mathrm{add}}$ scales with input dimension $D$ and hidden width $d$, without a separate dependence on the frozen assessment model's depth. Model-specific feature dimensions determine $D$. Appendix~\ref{app:inference} reports the dense computation and inference settings.
\section{Implementation Details}
\label{app:implementation}

\subsection{Refinement Architecture and Optimization}
\label{app:optimization}
The residual head uses the elementwise activation $\SiLU(t)=t/(1+\exp(-t))$. The hidden width is $d=64$, the ensemble size is $K=3$, the fixed output scale is $\gamma=0.375$, and hidden-layer dropout has probability $0.5$ during training only. The scalar output is multiplied by $\gamma$ without a bounding nonlinearity. During training, $\delta_i$ in the objective, including the residual penalty, is computed with inverted dropout ($p=0.5$) between SiLU and the output linear layer, scaling retained activations by $1/(1-p)$. The hidden layers of the three residual heads are independently initialized with seeds 42, 17, and 2026. All output-layer weights and biases are initialized to zero. Each residual head is trained for 85 full-batch updates using AdamW with learning rate $10^{-3}$, weight decay $0.02$, and gradient-norm clipping at 1. Each update uses the entire cached training set and updates only residual-head parameters. We use the final checkpoint of each residual head and average their corrections with equal weights. The two geometric metadata features contribute 384 weights across the three residual heads' input layers.

\subsection{Input Statistics and Score Scales}
\label{app:input-statistics}
Training means and second moments are accumulated in 64-bit floating-point precision (FP64). Population standard deviations are computed from these moments, with negative variances due to rounding set to zero. Means and standard deviations are stored in 32-bit floating-point precision (FP32). Standard deviations have a lower bound of $0.01$, and standardized inputs are clipped to $[-10,10]$. Features, normalized inputs, and residual-head computation use FP32. The input normalization statistics use only the training set for the residual heads. Configurations with evaluation scores on a normalized scale apply $\mathcal T(s)=1+9s$. The remaining evaluated configurations use the identity map. Any dataset-specific fixed score and label conversions precede this refinement mapping and are retained for both original and refined scores. At inference, FP32 residual corrections are converted to FP64 and divided by the slope $a$ of $\mathcal T$ (Appendix~\ref{app:normalization}) before addition to the original scores on the evaluation score scale. The refined scores are returned without clipping to the dataset's score range.

\subsection{Loss Settings}
\label{app:loss-settings}
The Huber transition is $\kappa=0.2$, and the smoothed absolute-error constant is $\epsilon=0.02$. Loss weights are $\lambda_{\mathrm{abs}}=0.05$, $\lambda_{\mathrm{corr}}=0.5$, $\lambda_{\mathrm{pair}}=0.05$, and $\lambda_{\mathrm{res}}=0.005$. The Pearson denominator, computed as the square root of the product of the centered sums of squares, is floored at $10^{-8}$. Pairwise targets use temperature $T_{\mathrm{rank}}=0.5$. Binary cross-entropy is evaluated directly from the predicted logits for numerical stability.

\subsection{Pair Sampling}
\label{app:pair-sampling}
We sample image pairs for the pairwise ranking loss at initialization and resample them after 20, 40, 60, and 80 parameter updates. At each sampling step, we draw 65,536 candidate pairs by sampling image indices uniformly with replacement from the full training set. Pairs with identical indices or an absolute label difference below $0.25$ on the refinement score scale are removed. The first 16,384 valid pairs are retained, or all valid pairs if fewer remain. Repeated pairs are allowed. The sampler requires at least 50 training images and 16 retained pairs. Sampling uses seed 620 initially and seed $620+t$ when resampling after $t$ completed parameter updates. All three residual heads use the same pair sampling schedule. Each retained pair has equal weight.

\subsection{Feature Extraction and Evaluation Context}
\label{app:feature-extraction}
Feature extraction locations are architecture-specific, typically covering three intermediate depths per selected visual branch. The Charm-AVA configuration aggregates all selected layer features. For example, TANet contributes features from three ResNet stages and three MobileNet stages. By contrast, LAR-IQA~\cite{avanaki2024lar} uses three stages from each of its two visual branches. MUSIQ additionally contributes its final patch-token mean alongside its final class-token feature and three intermediate class-token and patch-token summaries. Frozen normalization and projection are included where required by the original representations.

Spatial or token pooling produces feature vectors, followed by the prescribed aggregation across crops, scales, or viewports. This feature aggregation retains the original scoring procedure. For example, TReS supplies features averaged over its unflipped crops and retains the full original scoring computation. Models with context-dependent outputs retain their prescribed context during feature extraction. Model-specific reproduction and scoring settings are detailed in Appendix~\ref{app:retrained-models}.

\subsection{Feature Groups for Ablation}
\label{app:feature-groups}
For visual-feature ablations, the final group contains the model's final readout features and any designated final-stage summaries. For MUSIQ, this includes both the final class-token feature and final patch-token mean. The remaining extracted features form early, middle, and late groups according to their relative extraction depths. These groups follow architecture-specific locations rather than equal divisions of network depth. For example, CLIP-IQA+ (ViT-L/14) uses Transformer blocks 6, 12, and 18 for these three groups, while its final group contains the normalized scoring embedding. For models with multiple branches, features at corresponding relative depths are grouped together. The four groups occupy disjoint coordinates of the aggregated visual vector. Removing a feature group sets its normalized input coordinates to zero throughout both training and inference, retaining the input dimension and frozen model computation. The residual heads are retrained under the same optimization settings.

\subsection{Computation and Inference}
\label{app:inference}
Training reuses cached features from the frozen assessment model. At inference, all heads share features from one execution of the original scoring pipeline, reuse the training normalization statistics, and use all hidden activations with training-time dropout turned off. The three corrections are averaged equally. The dense residual heads require approximately $K(Dd+d)$ multiply-accumulate operations per image, in addition to pooling, frozen normalization or projection, input standardization, and elementwise operations.

The mean added parameter count of 1.64M averages all configurations in Tables~1 and 2 of the main paper, whereas the default head cost of 1.70M MACs averages the subset used for the ablations in Table~3. Both summaries weight configurations equally. The two configuration sets have different mean input dimensions $D$, accounting for the difference between these averages.
\section{Datasets and Evaluation Protocols}
\label{app:scope}

\subsection{Parameter Overhead}
\label{app:configurations}
Parameter overhead is the added trainable parameter count divided by the complete frozen assessment model's parameter count. Feature dimensions determine the added parameter count, while the relative overhead also depends on the original model's parameter count. For UNIQUE-mix, high-dimensional features are paired with a relatively small original model. Therefore, the added residual heads amount to 225.87\% of the original model's parameters.

\subsection{Data and Checkpoint Protocols}
\label{app:data-protocols}
\Cref{tab:data-protocols} gives the default data partitions and their principal exceptions. The assessment models include released checkpoints, cross-dataset checkpoints, and models retrained before refinement. We freeze each selected assessment model and optimize the residual heads using its designated training split. Model-specific reproduction and adaptation settings are detailed in Appendix~\ref{app:retrained-models}.

For Flickr-AES~\cite{ren2017personalized}, we use the public 4,737-image test list and group the remaining images by identical pixel content and assign each group to a single split, yielding 32,186 training and 3,576 validation images.

\begin{table*}[!tbp]
\centering
\caption{\textbf{Data partitions for the paired score-refinement comparisons.} Counts denote samples in the designated refinement training, validation, and evaluation partitions, including whole images and cropped regions for FLIVE. A dash means no separate validation partition is listed. Configuration-specific exceptions appear in the last column, and model reproduction and adaptation settings are detailed in Appendix~\ref{app:retrained-models}.}
\label{tab:data-protocols}
\begingroup\fontsize{8}{9.5}\selectfont
\setlength{\tabcolsep}{3pt}
\begin{tabular*}{\textwidth}{@{\extracolsep{\fill}}lrrrp{265pt}@{}}
\toprule
\rowcolor[gray]{0.92}Dataset & Train & Val. & Eval. & Configuration-specific settings \\
\midrule
\rowcolor[gray]{0.97}\cellcolor[gray]{0.92}KonIQ-10k & 7,058 & 1,000 & 2,015 & QCN and MANIQA: 8,058 training images, MANIQA uses an alternative author list. \\
\cellcolor[gray]{0.92}KADID-10k & 8,000 & -- & 2,125 & NLNet: author content split, 6,000/2,000/2,125, 17 evaluation contents. \\
\rowcolor[gray]{0.97}\cellcolor[gray]{0.92}SPAQ & 7,910 & 1,086 & 2,129 & QualiCLIP+ and NIMA IR-v2: 9,013/--/2,112, QCN: 8,996/--/2,129. \\
\cellcolor[gray]{0.92}LIVE Challenge & 814 & 117 & 231 & HyperIQA/HyperIQA+NT: 930/--/232, QualiCLIP+: author list, 931/--/231. \\
\rowcolor[gray]{0.97}\cellcolor[gray]{0.92}FLIVE & 121,080 & 30,856 & 7,304 & Default: whole images and cropped regions, DP-IQA and PaQ-2-PiQ: whole images only, 30,270/7,714/1,826. \\
\cellcolor[gray]{0.92}AVA & 235,579 & -- & 19,929 & AesCLIP: exposure-separated 248,259/--/7,249. \\
\rowcolor[gray]{0.97}\cellcolor[gray]{0.92}PARA GIAA & 28,268 & -- & 2,952 & PIAA-MIR and PIAA-ICI use one-hot trait encoding with their respective backbones. \\
\cellcolor[gray]{0.92}TAD66K & 52,255 & -- & 14,072 & Seven exact pixel duplicates are assigned to training, EAT/TANet retain batch context. \\
\rowcolor[gray]{0.97}\cellcolor[gray]{0.92}AADB & 8,458 & 372 & 1,000 & One train/evaluation pixel duplicate. \\
\cellcolor[gray]{0.92}BAID & 51,872 & 2,840 & 5,625 &  \\
\rowcolor[gray]{0.97}\cellcolor[gray]{0.92}UHD-IQA & 4,269 & 904 & 900 & Native configured inputs are retained for UIQA and both LAR-IQA variants. \\
\cellcolor[gray]{0.92}AGIQA-3K & 1,782 & 595 & 605 & Prompt and pixel groups, IP-IQA and AMFF-Net are retrained before refinement. \\
\rowcolor[gray]{0.97}\cellcolor[gray]{0.92}TID2013 & 1,800 & 600 & 600 & NLNet uses five evaluation reference contents, RankIQA uses reconstructed inference. \\
\cellcolor[gray]{0.92}LIVE & 460 & 159 & 160 & NLNet uses six evaluation reference contents, RankIQA uses reconstructed inference. DBCNN: 455/134/190. \\
\rowcolor[gray]{0.97}\cellcolor[gray]{0.92}CSIQ & 521 & 172 & 173 & NLNet author content split with six evaluation reference contents. \\
\cellcolor[gray]{0.92}CVIQ & 429 & -- & 99 & Assessor360 uses the author's validation images for evaluation, IQCaption360: 340/102/102. \\
\rowcolor[gray]{0.97}\cellcolor[gray]{0.92}FGRestore & 14,700 & -- & 3,706 & FGResQ retains the configured original scoring pipeline. \\
\cellcolor[gray]{0.92}GFIQA & 13,998 & 1,999 & 4,001 & DSL-FIQA retains 20 crops and its auxiliary face models. \\
\rowcolor[gray]{0.97}\cellcolor[gray]{0.92}CGFIQA & 27,527 & 3,930 & 7,855 & DSL-FIQA retains 20 crops and its auxiliary face models. \\
\cellcolor[gray]{0.92}LIVE-MD & 270 & 90 & 90 & DBCNN uses the configured content-grouped partition. \\
\rowcolor[gray]{0.97}\cellcolor[gray]{0.92}Flickr-AES & 32,186 & 3,576 & 4,737 & Aggregate MOS. PAM generic model, frozen BLG-PIAA prior, EAT fixed 16-image context. \\
\bottomrule
\end{tabular*}
\endgroup
\end{table*}

\subsection{Model Reproduction and Adaptation}
\label{app:retrained-models}
The retrained configurations use task-specific training protocols before the assessment model is frozen. IP-IQA and AMFF-Net use the AGIQA-3K partition with 1,782/595/605 training/validation/evaluation images. AMFF-Net is randomly initialized. IQCaption360 is adapted to CVIQ with 340/102/102 training/validation/evaluation images, including reference images. NIMA (MobileNet, technical) uses 2,640 training and 360 evaluation images from disjoint TID2013 reference groups, with three reference groups in the evaluation split. 

The evaluation groups number 2,015 for QPT V2 (IQA), 7,249 for QPT V2 (IAA), 17 for each of SIND, DFSS-IQA, and PAUQA, 900 for GLUQ, and 999 for TAVAR. Cross-dataset model names identify their training dataset where needed. QUASAR uses AVA-derived anchors without target-dataset fitting of the original scorer, while its residual heads are trained on the target training split.

EAT~\cite{he2023eat} and TANet retain their fixed evaluation batches, and QCN~\cite{shin2024qcn} retains its auxiliary inputs and comparison context. In Figure~2 of the main paper, QualiCLIP+ denotes a supervised adaptation of the original method~\cite{agnolucci2024qualiclip}. PIAA-ICI and PIAA-MIR denote the one-hot encoded Swin-Tiny variants described by Chen et al.~\cite{chen2025individual}. DSL-FIQA retains its 20 crops and auxiliary face models. RankIQA on LIVE and TID2013 uses imported Caffe VGG16 weights with a reconstructed inference implementation, averaging scores over 30 native-resolution $224\times224$ crops with fixed BGR channel means of $[104,117,123]$.

PAM~\cite{ren2017personalized}, BLG-PIAA~\cite{zhu2022blgpiaa}, and EAT~\cite{he2023eat} are evaluated against per-image aggregate MOS on Flickr-AES~\cite{ren2017personalized}. Scores are evaluated on the normalized $[0.2,1]$ MOS scale using the partitions specified in Appendix~\ref{app:data-protocols}. PAM retains its released generic scoring graph. BLG-PIAA uses its frozen ResNet-18 aesthetic prior without user-specific adaptation. EAT uses an independent fixed 224-pixel input protocol, 16-image evaluation batches, and a sigmoid score mapping established on training images. These scoring settings are shared by the frozen baselines and all refinement variants.

\subsection{Score Scales and Uncertainty}
\label{app:score-scales}
PLCC is computed directly from the predicted scores, without fitting a logistic mapping to the evaluation labels. The original and refined predictions use identical label conversions and evaluation score scales within each configuration. Therefore, error magnitudes are compared within each model-dataset configuration. Gains use unrounded metric values when available, so small changes can round to zero in the main tables. UNIQUE-mix uses its original output scale without MOS calibration. Signed correlations are retained, including the negative correlations for IQCaption360. Paired 95\% intervals use 2,000 resamples of the relevant pixel, content, or evaluation-context groups.
\section{Additional Ablation Studies}
\label{app:ablations}

\subsection{Experimental Setup}
\label{app:ablation-setup}
\Cref{tab:ablation-1,tab:ablation-2} detail TReS~\cite{golestaneh2022tres} (FLIVE), PIAA-MIR (Swin-T)~\cite{chen2025individual} (PARA), CLIP-IQA+ (L/14)~\cite{wang2023clipiqa} (KonIQ-10k), MUSIQ~\cite{ke2021musiq} (SPAQ), and TOPIQ (Swin)~\cite{chen2024topiq} (AVA). Table~3 of the main paper and \Cref{tab:expanded-ablation-90} aggregate the ablation results and compare variants directly with Full. We evaluate 14 component variants using seeds 42, 17, 2026, 7, and 123 with the same frozen model and data partitions, and summarize all four metrics by their single-head means, sample standard deviations, and relative changes from the Full single-head mean over the same seeds. Full denotes the complete MetaRefine configuration. Masked inputs retain the full input dimension. Removing the original-score input masks only its normalized head-input coordinate and retains the additive score skip.

We use fixed data partitions for these five model-dataset configurations. The assessment models, cached representations, score conversions, and 85-update training settings are held fixed. Seed group A contains 42, 17, 2026, 7, and 123, and group B contains 314, 2718, 1618, 577, and 811.

In the aggregate ablation comparisons, Full uses width-64 residual heads trained with Huber regression and $K=3$. The detailed single-head component and architecture studies retain the full input, objective, and head architecture, and compare metric means over matched seeds. Their Full reference is the corresponding single-head mean, as specified in each table caption.

\subsection{Input Features and Training Components}
\label{app:component-ablations}
We compare all component variants across the same 21 datasets as Table~3 of the main paper, using the complete \method configuration (Full) as the reference and weighting configurations equally. We summarize relative changes from Full and count the configurations where Full performs better (\Cref{tab:expanded-ablation-90}). Each variant uses seed group A (Appendix~\ref{app:ablation-setup}) with matched training data, cached representations, and optimization settings. The primary comparison averages predictions from the heads initialized with seeds 42, 17, and 2026.

The hidden-dropout ablation changes the training probability from $0.5$ to zero. Feature-group dropout is an additional variant that retains hidden dropout and independently zeros each early/middle/late input group with probability $0.4$ per image and training update (Appendix~\ref{app:feature-groups}), scaling retained groups by $1/0.6$. Final features, geometry, and the original-score input remain available. This random masking and scaling apply only during training. Full uses no feature-group dropout.

Hidden dropout and the residual penalty contribute to the balance between absolute score accuracy and image ordering (Table~3(C-D) of the main paper). Retaining hidden dropout favors absolute accuracy: removing it increases MAE and RMSE by 0.313\% and 0.325\% on average relative to Full, while improving PLCC and SRCC by 0.533\% and 0.085\%. The residual penalty applies a soft cost to large corrections and supports rank agreement: removing it reduces SRCC by 0.121\%, with gains of 0.077\%, 0.147\%, and 0.132\% in PLCC, MAE, and RMSE. These complementary effects motivate retaining both terms in the shared training recipe.

Full improves both correlations over every variant for TReS (FLIVE). Among the 14 reported comparisons, it improves all four metrics in 12 for TReS, 12 for PIAA-MIR, 12 for CLIP-IQA+, 12 for MUSIQ, and 9 for TOPIQ.

\begin{table*}[!tbp]
\centering
\caption{\textbf{Component comparisons.} Three-head variants versus the complete MetaRefine configuration (Full). In the upper panels, each cell gives the equally weighted mean relative improvement over Full and the number of configurations where Full is better. The final panel reports median changes for the three original-score pathway variants. Positive values indicate higher correlation or lower error. Negative/positive changes have red/green backgrounds.}
\label{tab:expanded-ablation-90}
\label{tab:score-paths-90}
\begingroup\fontsize{7}{9}\selectfont
\setlength{\tabcolsep}{3pt}
\begin{tabular*}{\textwidth}{@{\extracolsep{\fill}}lrrrr@{}}
\toprule
\rowcolor[gray]{0.92}Variant & PLCC $\uparrow$ & SRCC $\uparrow$ & MAE $\uparrow$ & RMSE $\uparrow$ \\
\midrule
\rowcolor[gray]{0.92}\multicolumn{5}{@{}l}{\textbf{(A) Multi-Level Visual Features}} \\
w/o visual features & \cellcolor[RGB]{247,203,203}-3.638\% (83) & \cellcolor[RGB]{247,202,202}-3.889\% (80) & \cellcolor[RGB]{235,151,151}-55.656\% (82) & \cellcolor[RGB]{235,151,151}-48.445\% (81) \\
Final features only & \cellcolor[RGB]{250,217,217}-0.960\% (75) & \cellcolor[RGB]{249,215,215}-1.265\% (73) & \cellcolor[RGB]{245,197,197}-5.371\% (73) & \cellcolor[RGB]{245,197,197}-5.436\% (77) \\
w/o final features & \cellcolor[RGB]{251,223,223}-0.285\% (63) & \cellcolor[RGB]{226,242,229}+0.267\% (61) & \cellcolor[RGB]{244,192,192}-7.129\% (58) & \cellcolor[RGB]{244,193,193}-6.751\% (61) \\
w/o early features & \cellcolor[RGB]{251,224,224}-0.261\% (58) & \cellcolor[RGB]{251,223,223}-0.288\% (57) & \cellcolor[RGB]{252,226,226}-0.097\% (60) & \cellcolor[RGB]{252,225,225}-0.177\% (61) \\
w/o middle features & \cellcolor[RGB]{251,223,223}-0.309\% (64) & \cellcolor[RGB]{251,222,222}-0.353\% (62) & \cellcolor[RGB]{252,225,225}-0.174\% (58) & \cellcolor[RGB]{251,224,224}-0.221\% (60) \\
w/o late features & \cellcolor[RGB]{229,243,232}+0.123\% (63) & \cellcolor[RGB]{251,224,224}-0.262\% (63) & \cellcolor[RGB]{251,221,221}-0.458\% (57) & \cellcolor[RGB]{250,220,220}-0.603\% (66) \\
\rowcolor[gray]{0.92}\multicolumn{5}{@{}l}{\textbf{(B) Geometric Context and Score Pathways}} \\
w/o geometric metadata & \cellcolor[RGB]{252,229,229}-0.026\% (39) & \cellcolor[RGB]{252,228,228}-0.033\% (37) & \cellcolor[RGB]{252,228,228}-0.057\% (42) & \cellcolor[RGB]{252,228,228}-0.041\% (40) \\
w/o score input to head & \cellcolor[RGB]{252,228,228}-0.037\% (51) & \cellcolor[RGB]{252,228,228}-0.035\% (46) & \cellcolor[RGB]{235,246,237}+0.001\% (52) & \cellcolor[RGB]{234,246,236}+0.004\% (55) \\
w/o geometric metadata \& score input & \cellcolor[RGB]{252,227,227}-0.065\% (64) & \cellcolor[RGB]{252,227,227}-0.061\% (58) & \cellcolor[RGB]{252,227,227}-0.068\% (64) & \cellcolor[RGB]{252,228,228}-0.053\% (66) \\
w/o score skip & \cellcolor[RGB]{240,173,173}-15.751\% (90) & \cellcolor[RGB]{241,176,176}-14.277\% (89) & \cellcolor[RGB]{235,151,151}-170.260\% (92) & \cellcolor[RGB]{235,151,151}-154.029\% (92) \\
w/o score input \& skip & \cellcolor[RGB]{240,173,173}-15.723\% (90) & \cellcolor[RGB]{241,176,176}-14.359\% (89) & \cellcolor[RGB]{235,151,151}-171.385\% (92) & \cellcolor[RGB]{235,151,151}-154.958\% (92) \\
\rowcolor[gray]{0.92}\multicolumn{5}{@{}l}{\textbf{(C) Training Objective}} \\
w/o correlation loss & \cellcolor[RGB]{250,219,219}-0.622\% (58) & \cellcolor[RGB]{250,219,219}-0.630\% (54) & \cellcolor[RGB]{231,244,234}+0.048\% (50) & \cellcolor[RGB]{234,245,236}+0.007\% (52) \\
w/o residual penalty & \cellcolor[RGB]{230,244,233}+0.077\% (63) & \cellcolor[RGB]{252,226,226}-0.121\% (60) & \cellcolor[RGB]{228,243,231}+0.147\% (44) & \cellcolor[RGB]{229,243,232}+0.132\% (49) \\
\rowcolor[gray]{0.92}\multicolumn{5}{@{}l}{\textbf{(D) Regularization}} \\
w/o hidden dropout & \cellcolor[RGB]{223,241,226}+0.533\% (57) & \cellcolor[RGB]{230,244,233}+0.085\% (60) & \cellcolor[RGB]{251,223,223}-0.313\% (52) & \cellcolor[RGB]{251,223,223}-0.325\% (57) \\
Feature-group dropout (0.4) & \cellcolor[RGB]{252,227,227}-0.075\% (49) & \cellcolor[RGB]{251,223,223}-0.283\% (47) & \cellcolor[RGB]{249,215,215}-1.185\% (50) & \cellcolor[RGB]{250,216,216}-1.088\% (46) \\
\rowcolor[gray]{0.92}\multicolumn{5}{@{}l}{\textbf{(F) Residual-Head Architecture}} \\
Linear (0.03M MACs) & \cellcolor[RGB]{221,240,225}+0.727\% (73) & \cellcolor[RGB]{216,238,221}+1.228\% (68) & \cellcolor[RGB]{235,151,151}-37.458\% (81) & \cellcolor[RGB]{235,152,152}-29.405\% (78) \\
MLP (32, 0.85M MACs) & \cellcolor[RGB]{252,227,227}-0.069\% (57) & \cellcolor[RGB]{226,242,229}+0.291\% (54) & \cellcolor[RGB]{249,215,215}-1.158\% (53) & \cellcolor[RGB]{250,216,216}-1.089\% (58) \\
MLP (64, 1.70M MACs, default) & 0.000\% (0) & 0.000\% (0) & 0.000\% (0) & 0.000\% (0) \\
MLP (128, 3.40M MACs) & \cellcolor[RGB]{225,242,229}+0.337\% (31) & \cellcolor[RGB]{227,242,230}+0.228\% (32) & \cellcolor[RGB]{224,241,227}+0.450\% (36) & \cellcolor[RGB]{224,241,228}+0.383\% (35) \\
MLP (64--64, 1.71M MACs) & \cellcolor[RGB]{227,242,230}+0.243\% (35) & \cellcolor[RGB]{220,240,224}+0.754\% (28) & \cellcolor[RGB]{248,207,207}-2.766\% (46) & \cellcolor[RGB]{248,208,208}-2.401\% (48) \\
\rowcolor[gray]{0.92}\multicolumn{5}{@{}l}{\textbf{(G) Correction Bound}} \\
No bound (default) & 0.000\% (0) & 0.000\% (0) & 0.000\% (0) & 0.000\% (0) \\
0.1875 & \cellcolor[RGB]{248,209,209}-2.248\% (73) & \cellcolor[RGB]{248,209,209}-2.294\% (68) & \cellcolor[RGB]{235,151,151}-54.968\% (77) & \cellcolor[RGB]{235,151,151}-48.182\% (73) \\
0.375 & \cellcolor[RGB]{249,214,214}-1.384\% (71) & \cellcolor[RGB]{249,214,214}-1.347\% (64) & \cellcolor[RGB]{235,151,151}-45.274\% (69) & \cellcolor[RGB]{235,151,151}-40.412\% (67) \\
0.75 & \cellcolor[RGB]{250,219,219}-0.645\% (69) & \cellcolor[RGB]{250,218,218}-0.752\% (69) & \cellcolor[RGB]{235,151,151}-31.736\% (71) & \cellcolor[RGB]{235,152,152}-29.118\% (70) \\
1.5 & \cellcolor[RGB]{223,241,227}+0.483\% (65) & \cellcolor[RGB]{223,241,226}+0.528\% (61) & \cellcolor[RGB]{240,171,171}-16.758\% (64) & \cellcolor[RGB]{240,172,172}-16.252\% (66) \\
3.375 & \cellcolor[RGB]{211,235,216}+2.044\% (59) & \cellcolor[RGB]{213,236,218}+1.725\% (54) & \cellcolor[RGB]{244,192,192}-7.125\% (55) & \cellcolor[RGB]{244,191,191}-7.511\% (57) \\
Adaptive (90th percentile) & \cellcolor[RGB]{211,235,216}+2.025\% (65) & \cellcolor[RGB]{213,236,218}+1.727\% (62) & \cellcolor[RGB]{244,191,191}-7.436\% (57) & \cellcolor[RGB]{244,190,190}-7.846\% (62) \\
\midrule
\rowcolor[gray]{0.92}\multicolumn{5}{@{}l}{\textbf{Median Changes for Original-Score Pathways}} \\
w/o head-score input & \cellcolor[RGB]{253,231,231}-0.001\% & \cellcolor[RGB]{235,246,237}+0.000\% & \cellcolor[RGB]{253,230,230}-0.003\% & \cellcolor[RGB]{253,230,230}-0.005\% \\
w/o score skip & \cellcolor[RGB]{241,176,176}-14.176\% & \cellcolor[RGB]{241,178,178}-13.374\% & \cellcolor[RGB]{235,151,151}-120.739\% & \cellcolor[RGB]{235,151,151}-107.610\% \\
w/o head-score input \& score skip & \cellcolor[RGB]{241,176,176}-14.098\% & \cellcolor[RGB]{241,178,178}-13.238\% & \cellcolor[RGB]{235,151,151}-121.178\% & \cellcolor[RGB]{235,151,151}-107.218\% \\
\bottomrule
\end{tabular*}
\endgroup
\end{table*}

\begin{table*}[!tbp]
\centering
\caption{\textbf{Component ablations, part 1 of 2.} Full denotes \method. Single-head means $\pm$ sample standard deviations use seed group A (Appendix~\ref{app:ablation-setup}). Changes compare each mean with the Full mean over the same seeds. Positive/green denotes improvement and negative/red decline.}
\label{tab:ablation-1}
\begingroup\fontsize{7.5}{9}\selectfont
\setlength{\tabcolsep}{1pt}
\begin{tabular*}{\textwidth}{@{\extracolsep{\fill}}l*{4}{r>{\fontsize{7}{9}\selectfont}r}@{}}
\toprule
Model (Dataset) & \multicolumn{2}{c}{PLCC {\fontsize{7.5}{9}\selectfont$\uparrow$}} & \multicolumn{2}{c}{SRCC {\fontsize{7.5}{9}\selectfont$\uparrow$}} & \multicolumn{2}{c}{MAE {\fontsize{7.5}{9}\selectfont$\downarrow$}} & \multicolumn{2}{c}{RMSE {\fontsize{7.5}{9}\selectfont$\downarrow$}} \\
\cmidrule(lr){2-3}\cmidrule(lr){4-5}\cmidrule(lr){6-7}\cmidrule(lr){8-9}
Variant & Mean $\pm$ SD & Change & Mean $\pm$ SD & Change & Mean $\pm$ SD & Change & Mean $\pm$ SD & Change \\
\midrule
\addlinespace[2pt]\rowcolor[gray]{0.92}\multicolumn{9}{@{}l}{\rule{0pt}{11pt}\makebox[105pt][l]{\textbf{TReS}}\hspace{12pt}\textbf{(FLIVE)}} \\
Frozen assessment model & 0.6398 &  & 0.6293 &  & 0.1317 &  & 0.1470 &  \\
Full & 0.7468\,\textpm\,0.0008 & \cellcolor[RGB]{255,255,255}0.00\% & 0.7104\,\textpm\,0.0007 & \cellcolor[RGB]{255,255,255}0.00\% & 0.0444\,\textpm\,0.0001 & \cellcolor[RGB]{255,255,255}0.00\% & 0.0588\,\textpm\,0.0001 & \cellcolor[RGB]{255,255,255}0.00\% \\
Final features only & 0.7221\,\textpm\,0.0006 & \cellcolor[RGB]{247,210,210}-3.30\% & 0.6804\,\textpm\,0.0011 & \cellcolor[RGB]{246,207,207}-4.21\% & 0.0463\,\textpm\,0.0001 & \cellcolor[RGB]{246,206,206}-4.28\% & 0.0611\,\textpm\,0.0001 & \cellcolor[RGB]{246,208,208}-3.95\% \\
Without final features & 0.7424\,\textpm\,0.0012 & \cellcolor[RGB]{250,228,228}-0.58\% & 0.7079\,\textpm\,0.0015 & \cellcolor[RGB]{251,230,230}-0.34\% & 0.0450\,\textpm\,0.0001 & \cellcolor[RGB]{249,222,222}-1.29\% & 0.0595\,\textpm\,0.0001 & \cellcolor[RGB]{249,222,222}-1.19\% \\
Without early features & 0.7466\,\textpm\,0.0012 & \cellcolor[RGB]{252,237,237}-0.03\% & 0.7099\,\textpm\,0.0012 & \cellcolor[RGB]{252,236,236}-0.06\% & 0.0445\,\textpm\,0.0000 & \cellcolor[RGB]{252,235,235}-0.10\% & 0.0589\,\textpm\,0.0001 & \cellcolor[RGB]{252,235,235}-0.09\% \\
Without middle features & 0.7433\,\textpm\,0.0021 & \cellcolor[RGB]{251,229,229}-0.47\% & 0.7062\,\textpm\,0.0029 & \cellcolor[RGB]{250,228,228}-0.58\% & 0.0447\,\textpm\,0.0001 & \cellcolor[RGB]{251,228,228}-0.56\% & 0.0591\,\textpm\,0.0001 & \cellcolor[RGB]{251,228,228}-0.51\% \\
Without late features & 0.7378\,\textpm\,0.0010 & \cellcolor[RGB]{249,222,222}-1.20\% & 0.7002\,\textpm\,0.0018 & \cellcolor[RGB]{249,221,221}-1.44\% & 0.0452\,\textpm\,0.0001 & \cellcolor[RGB]{249,219,219}-1.69\% & 0.0597\,\textpm\,0.0001 & \cellcolor[RGB]{249,220,220}-1.49\% \\
Without geometric metadata & 0.7446\,\textpm\,0.0010 & \cellcolor[RGB]{251,231,231}-0.28\% & 0.7069\,\textpm\,0.0006 & \cellcolor[RGB]{251,229,229}-0.49\% & 0.0446\,\textpm\,0.0001 & \cellcolor[RGB]{251,229,229}-0.42\% & 0.0590\,\textpm\,0.0001 & \cellcolor[RGB]{251,230,230}-0.37\% \\
Without original-score input & 0.7466\,\textpm\,0.0008 & \cellcolor[RGB]{252,237,237}-0.03\% & 0.7100\,\textpm\,0.0004 & \cellcolor[RGB]{252,236,236}-0.05\% & 0.0444\,\textpm\,0.0001 & \cellcolor[RGB]{235,246,237}+0.05\% & 0.0588\,\textpm\,0.0001 & \cellcolor[RGB]{237,247,238}+0.02\% \\
w/o geometric metadata \& score input & 0.7445\,\textpm\,0.0008 & \cellcolor[RGB]{251,231,231}-0.30\% & 0.7068\,\textpm\,0.0004 & \cellcolor[RGB]{251,228,228}-0.51\% & 0.0446\,\textpm\,0.0001 & \cellcolor[RGB]{251,229,229}-0.46\% & 0.0590\,\textpm\,0.0001 & \cellcolor[RGB]{251,230,230}-0.40\% \\
w/o visual features & 0.6308\,\textpm\,0.0066 & \cellcolor[RGB]{240,176,176}-15.53\% & 0.6274\,\textpm\,0.0073 & \cellcolor[RGB]{242,184,184}-11.68\% & 0.1106\,\textpm\,0.0007 & \cellcolor[RGB]{235,151,151}-148.77\% & 0.1292\,\textpm\,0.0005 & \cellcolor[RGB]{235,151,151}-119.73\% \\
Without correlation loss & 0.7337\,\textpm\,0.0024 & \cellcolor[RGB]{249,218,218}-1.76\% & 0.6982\,\textpm\,0.0025 & \cellcolor[RGB]{249,219,219}-1.72\% & 0.0451\,\textpm\,0.0002 & \cellcolor[RGB]{249,220,220}-1.46\% & 0.0600\,\textpm\,0.0002 & \cellcolor[RGB]{248,217,217}-2.06\% \\
Without residual penalty & 0.7466\,\textpm\,0.0009 & \cellcolor[RGB]{252,237,237}-0.03\% & 0.7100\,\textpm\,0.0007 & \cellcolor[RGB]{252,237,237}-0.04\% & 0.0445\,\textpm\,0.0001 & \cellcolor[RGB]{252,235,235}-0.09\% & 0.0589\,\textpm\,0.0001 & \cellcolor[RGB]{252,236,236}-0.07\% \\
Bounded control (0.375) & 0.6426\,\textpm\,0.0007 & \cellcolor[RGB]{241,179,179}-13.95\% & 0.6295\,\textpm\,0.0000 & \cellcolor[RGB]{242,185,185}-11.39\% & 0.0932\,\textpm\,0.0000 & \cellcolor[RGB]{235,151,151}-109.74\% & 0.1117\,\textpm\,0.0000 & \cellcolor[RGB]{235,151,151}-89.87\% \\
Without hidden dropout & 0.7463\,\textpm\,0.0019 & \cellcolor[RGB]{252,236,236}-0.07\% & 0.7071\,\textpm\,0.0023 & \cellcolor[RGB]{251,229,229}-0.46\% & 0.0444\,\textpm\,0.0001 & \cellcolor[RGB]{235,246,237}+0.05\% & 0.0588\,\textpm\,0.0002 & \cellcolor[RGB]{252,236,236}-0.05\% \\
Feature-group dropout (0.4) & 0.7433\,\textpm\,0.0017 & \cellcolor[RGB]{251,229,229}-0.47\% & 0.7043\,\textpm\,0.0021 & \cellcolor[RGB]{250,225,225}-0.85\% & 0.0446\,\textpm\,0.0000 & \cellcolor[RGB]{251,229,229}-0.45\% & 0.0591\,\textpm\,0.0001 & \cellcolor[RGB]{251,228,228}-0.53\% \\
\addlinespace[2pt]\rowcolor[gray]{0.92}\multicolumn{9}{@{}l}{\rule{0pt}{11pt}\makebox[105pt][l]{\textbf{PIAA-MIR (Swin-T)}}\hspace{12pt}\textbf{(PARA GIAA)}} \\
Frozen assessment model & 0.7931 &  & 0.7236 &  & 0.0702 &  & 0.0925 &  \\
Full & 0.8976\,\textpm\,0.0002 & \cellcolor[RGB]{255,255,255}0.00\% & 0.8561\,\textpm\,0.0003 & \cellcolor[RGB]{255,255,255}0.00\% & 0.0446\,\textpm\,0.0000 & \cellcolor[RGB]{255,255,255}0.00\% & 0.0589\,\textpm\,0.0001 & \cellcolor[RGB]{255,255,255}0.00\% \\
Final features only & 0.8823\,\textpm\,0.0003 & \cellcolor[RGB]{249,219,219}-1.70\% & 0.8360\,\textpm\,0.0003 & \cellcolor[RGB]{248,215,215}-2.35\% & 0.0477\,\textpm\,0.0001 & \cellcolor[RGB]{244,197,197}-6.97\% & 0.0631\,\textpm\,0.0001 & \cellcolor[RGB]{244,197,197}-7.10\% \\
Without final features & 0.8922\,\textpm\,0.0002 & \cellcolor[RGB]{250,227,227}-0.60\% & 0.8531\,\textpm\,0.0002 & \cellcolor[RGB]{251,230,230}-0.35\% & 0.0459\,\textpm\,0.0001 & \cellcolor[RGB]{248,213,213}-2.75\% & 0.0604\,\textpm\,0.0001 & \cellcolor[RGB]{248,214,214}-2.51\% \\
Without early features & 0.8971\,\textpm\,0.0002 & \cellcolor[RGB]{252,236,236}-0.05\% & 0.8555\,\textpm\,0.0004 & \cellcolor[RGB]{252,236,236}-0.07\% & 0.0448\,\textpm\,0.0000 & \cellcolor[RGB]{251,231,231}-0.28\% & 0.0591\,\textpm\,0.0001 & \cellcolor[RGB]{251,232,232}-0.25\% \\
Without middle features & 0.8952\,\textpm\,0.0002 & \cellcolor[RGB]{251,232,232}-0.27\% & 0.8522\,\textpm\,0.0002 & \cellcolor[RGB]{251,229,229}-0.46\% & 0.0452\,\textpm\,0.0000 & \cellcolor[RGB]{249,222,222}-1.20\% & 0.0596\,\textpm\,0.0000 & \cellcolor[RGB]{249,222,222}-1.17\% \\
Without late features & 0.8913\,\textpm\,0.0002 & \cellcolor[RGB]{250,226,226}-0.70\% & 0.8488\,\textpm\,0.0004 & \cellcolor[RGB]{250,225,225}-0.86\% & 0.0459\,\textpm\,0.0001 & \cellcolor[RGB]{247,213,213}-2.81\% & 0.0607\,\textpm\,0.0001 & \cellcolor[RGB]{247,212,212}-2.95\% \\
Without geometric metadata & 0.8973\,\textpm\,0.0002 & \cellcolor[RGB]{252,237,237}-0.04\% & 0.8556\,\textpm\,0.0003 & \cellcolor[RGB]{252,236,236}-0.06\% & 0.0448\,\textpm\,0.0000 & \cellcolor[RGB]{251,232,232}-0.26\% & 0.0590\,\textpm\,0.0001 & \cellcolor[RGB]{252,234,234}-0.16\% \\
Without original-score input & 0.8976\,\textpm\,0.0001 & \cellcolor[RGB]{253,239,239}-0.01\% & 0.8561\,\textpm\,0.0003 & \cellcolor[RGB]{255,255,255}0.00\% & 0.0446\,\textpm\,0.0000 & \cellcolor[RGB]{252,237,237}-0.02\% & 0.0589\,\textpm\,0.0000 & \cellcolor[RGB]{252,237,237}-0.03\% \\
w/o geometric metadata \& score input & 0.8972\,\textpm\,0.0002 & \cellcolor[RGB]{252,237,237}-0.04\% & 0.8556\,\textpm\,0.0003 & \cellcolor[RGB]{252,236,236}-0.06\% & 0.0448\,\textpm\,0.0000 & \cellcolor[RGB]{251,231,231}-0.28\% & 0.0590\,\textpm\,0.0001 & \cellcolor[RGB]{252,233,233}-0.18\% \\
w/o visual features & 0.7968\,\textpm\,0.0002 & \cellcolor[RGB]{242,186,186}-11.23\% & 0.7249\,\textpm\,0.0002 & \cellcolor[RGB]{240,176,176}-15.33\% & 0.0638\,\textpm\,0.0001 & \cellcolor[RGB]{235,151,151}-42.85\% & 0.0828\,\textpm\,0.0001 & \cellcolor[RGB]{235,151,151}-40.46\% \\
Without correlation loss & 0.8961\,\textpm\,0.0001 & \cellcolor[RGB]{252,233,233}-0.16\% & 0.8543\,\textpm\,0.0003 & \cellcolor[RGB]{251,232,232}-0.21\% & 0.0449\,\textpm\,0.0000 & \cellcolor[RGB]{250,227,227}-0.64\% & 0.0593\,\textpm\,0.0001 & \cellcolor[RGB]{250,227,227}-0.65\% \\
Without residual penalty & 0.8977\,\textpm\,0.0002 & \cellcolor[RGB]{238,247,239}+0.01\% & 0.8561\,\textpm\,0.0003 & \cellcolor[RGB]{255,255,255}0.00\% & 0.0446\,\textpm\,0.0000 & \cellcolor[RGB]{235,246,236}+0.06\% & 0.0589\,\textpm\,0.0001 & \cellcolor[RGB]{235,246,237}+0.05\% \\
Bounded control (0.375) & 0.8610\,\textpm\,0.0003 & \cellcolor[RGB]{246,207,207}-4.08\% & 0.8330\,\textpm\,0.0003 & \cellcolor[RGB]{248,213,213}-2.70\% & 0.0556\,\textpm\,0.0000 & \cellcolor[RGB]{237,160,160}-24.46\% & 0.0751\,\textpm\,0.0000 & \cellcolor[RGB]{236,155,155}-27.54\% \\
Without hidden dropout & 0.8997\,\textpm\,0.0001 & \cellcolor[RGB]{230,244,233}+0.24\% & 0.8588\,\textpm\,0.0001 & \cellcolor[RGB]{229,244,232}+0.31\% & 0.0440\,\textpm\,0.0000 & \cellcolor[RGB]{218,239,222}+1.41\% & 0.0583\,\textpm\,0.0000 & \cellcolor[RGB]{220,240,224}+1.11\% \\
Feature-group dropout (0.4) & 0.8938\,\textpm\,0.0003 & \cellcolor[RGB]{251,229,229}-0.43\% & 0.8512\,\textpm\,0.0004 & \cellcolor[RGB]{251,228,228}-0.57\% & 0.0454\,\textpm\,0.0001 & \cellcolor[RGB]{249,218,218}-1.77\% & 0.0600\,\textpm\,0.0001 & \cellcolor[RGB]{249,218,218}-1.85\% \\
\addlinespace[2pt]\rowcolor[gray]{0.92}\multicolumn{9}{@{}l}{\rule{0pt}{11pt}\makebox[105pt][l]{\textbf{CLIP-IQA+ (L/14)}}\hspace{12pt}\textbf{(KonIQ-10k)}} \\
Frozen assessment model & 0.8747 &  & 0.8610 &  & 0.0602 &  & 0.0770 &  \\
Full & 0.9277\,\textpm\,0.0006 & \cellcolor[RGB]{255,255,255}0.00\% & 0.9104\,\textpm\,0.0007 & \cellcolor[RGB]{255,255,255}0.00\% & 0.0446\,\textpm\,0.0002 & \cellcolor[RGB]{255,255,255}0.00\% & 0.0581\,\textpm\,0.0002 & \cellcolor[RGB]{255,255,255}0.00\% \\
Final features only & 0.8908\,\textpm\,0.0003 & \cellcolor[RGB]{246,208,208}-3.97\% & 0.8722\,\textpm\,0.0005 & \cellcolor[RGB]{246,207,207}-4.19\% & 0.0539\,\textpm\,0.0001 & \cellcolor[RGB]{238,166,166}-20.65\% & 0.0702\,\textpm\,0.0001 & \cellcolor[RGB]{238,166,166}-20.94\% \\
Without final features & 0.9066\,\textpm\,0.0009 & \cellcolor[RGB]{248,216,216}-2.27\% & 0.8815\,\textpm\,0.0016 & \cellcolor[RGB]{247,211,211}-3.17\% & 0.0519\,\textpm\,0.0001 & \cellcolor[RGB]{240,174,174}-16.27\% & 0.0664\,\textpm\,0.0001 & \cellcolor[RGB]{241,179,179}-14.30\% \\
Without early features & 0.9271\,\textpm\,0.0007 & \cellcolor[RGB]{252,236,236}-0.07\% & 0.9093\,\textpm\,0.0008 & \cellcolor[RGB]{252,235,235}-0.11\% & 0.0449\,\textpm\,0.0002 & \cellcolor[RGB]{250,227,227}-0.63\% & 0.0582\,\textpm\,0.0003 & \cellcolor[RGB]{252,234,234}-0.16\% \\
Without middle features & 0.9243\,\textpm\,0.0006 & \cellcolor[RGB]{251,230,230}-0.37\% & 0.9054\,\textpm\,0.0007 & \cellcolor[RGB]{251,228,228}-0.54\% & 0.0456\,\textpm\,0.0002 & \cellcolor[RGB]{248,216,216}-2.25\% & 0.0593\,\textpm\,0.0002 & \cellcolor[RGB]{248,217,217}-2.06\% \\
Without late features & 0.9293\,\textpm\,0.0007 & \cellcolor[RGB]{232,245,234}+0.18\% & 0.9139\,\textpm\,0.0008 & \cellcolor[RGB]{228,243,231}+0.39\% & 0.0436\,\textpm\,0.0002 & \cellcolor[RGB]{212,236,217}+2.33\% & 0.0572\,\textpm\,0.0002 & \cellcolor[RGB]{217,238,221}+1.58\% \\
Without geometric metadata & 0.9277\,\textpm\,0.0006 & \cellcolor[RGB]{255,255,255}0.00\% & 0.9104\,\textpm\,0.0007 & \cellcolor[RGB]{255,255,255}0.00\% & 0.0446\,\textpm\,0.0002 & \cellcolor[RGB]{255,255,255}0.00\% & 0.0581\,\textpm\,0.0002 & \cellcolor[RGB]{255,255,255}0.00\% \\
Without original-score input & 0.9268\,\textpm\,0.0004 & \cellcolor[RGB]{252,235,235}-0.09\% & 0.9092\,\textpm\,0.0005 & \cellcolor[RGB]{252,234,234}-0.13\% & 0.0450\,\textpm\,0.0001 & \cellcolor[RGB]{250,226,226}-0.72\% & 0.0584\,\textpm\,0.0002 & \cellcolor[RGB]{250,227,227}-0.63\% \\
w/o geometric metadata \& score input & 0.9268\,\textpm\,0.0004 & \cellcolor[RGB]{252,235,235}-0.09\% & 0.9092\,\textpm\,0.0005 & \cellcolor[RGB]{252,234,234}-0.13\% & 0.0450\,\textpm\,0.0001 & \cellcolor[RGB]{250,226,226}-0.72\% & 0.0584\,\textpm\,0.0002 & \cellcolor[RGB]{250,227,227}-0.63\% \\
w/o visual features & 0.8747\,\textpm\,0.0001 & \cellcolor[RGB]{245,201,201}-5.71\% & 0.8610\,\textpm\,0.0000 & \cellcolor[RGB]{245,202,202}-5.42\% & 0.0579\,\textpm\,0.0000 & \cellcolor[RGB]{235,152,152}-29.61\% & 0.0748\,\textpm\,0.0000 & \cellcolor[RGB]{235,153,153}-28.76\% \\
Without correlation loss & 0.9268\,\textpm\,0.0003 & \cellcolor[RGB]{252,235,235}-0.10\% & 0.9094\,\textpm\,0.0002 & \cellcolor[RGB]{252,235,235}-0.10\% & 0.0449\,\textpm\,0.0002 & \cellcolor[RGB]{250,227,227}-0.66\% & 0.0584\,\textpm\,0.0002 & \cellcolor[RGB]{251,228,228}-0.58\% \\
Without residual penalty & 0.9277\,\textpm\,0.0006 & \cellcolor[RGB]{255,255,255}0.00\% & 0.9103\,\textpm\,0.0007 & \cellcolor[RGB]{253,238,238}-0.01\% & 0.0446\,\textpm\,0.0002 & \cellcolor[RGB]{253,238,238}-0.01\% & 0.0581\,\textpm\,0.0002 & \cellcolor[RGB]{253,238,238}-0.01\% \\
Bounded control (0.375) & 0.9086\,\textpm\,0.0005 & \cellcolor[RGB]{248,217,217}-2.06\% & 0.8921\,\textpm\,0.0005 & \cellcolor[RGB]{248,217,217}-2.01\% & 0.0500\,\textpm\,0.0001 & \cellcolor[RGB]{242,184,184}-11.91\% & 0.0649\,\textpm\,0.0001 & \cellcolor[RGB]{242,184,184}-11.67\% \\
Without hidden dropout & 0.9214\,\textpm\,0.0016 & \cellcolor[RGB]{250,227,227}-0.68\% & 0.9027\,\textpm\,0.0019 & \cellcolor[RGB]{250,225,225}-0.84\% & 0.0467\,\textpm\,0.0004 & \cellcolor[RGB]{246,205,205}-4.62\% & 0.0605\,\textpm\,0.0004 & \cellcolor[RGB]{246,207,207}-4.19\% \\
Feature-group dropout (0.4) & 0.9164\,\textpm\,0.0007 & \cellcolor[RGB]{249,222,222}-1.21\% & 0.8966\,\textpm\,0.0007 & \cellcolor[RGB]{249,220,220}-1.51\% & 0.0476\,\textpm\,0.0002 & \cellcolor[RGB]{245,198,198}-6.65\% & 0.0618\,\textpm\,0.0002 & \cellcolor[RGB]{245,199,199}-6.34\% \\
\bottomrule
\end{tabular*}
\endgroup
\end{table*}

\begin{table*}[!tbp]
\centering
\caption{\textbf{Component ablations, part 2 of 2.} Full denotes \method. Single-head means $\pm$ sample standard deviations use seed group A (Appendix~\ref{app:ablation-setup}). Changes compare each mean with the Full mean over the same seeds. Positive/green denotes improvement and negative/red decline.}
\label{tab:ablation-2}
\begingroup\fontsize{7.5}{9}\selectfont
\setlength{\tabcolsep}{1pt}
\begin{tabular*}{\textwidth}{@{\extracolsep{\fill}}l*{4}{r>{\fontsize{7}{9}\selectfont}r}@{}}
\toprule
Model (Dataset) & \multicolumn{2}{c}{PLCC {\fontsize{7.5}{9}\selectfont$\uparrow$}} & \multicolumn{2}{c}{SRCC {\fontsize{7.5}{9}\selectfont$\uparrow$}} & \multicolumn{2}{c}{MAE {\fontsize{7.5}{9}\selectfont$\downarrow$}} & \multicolumn{2}{c}{RMSE {\fontsize{7.5}{9}\selectfont$\downarrow$}} \\
\cmidrule(lr){2-3}\cmidrule(lr){4-5}\cmidrule(lr){6-7}\cmidrule(lr){8-9}
Variant & Mean $\pm$ SD & Change & Mean $\pm$ SD & Change & Mean $\pm$ SD & Change & Mean $\pm$ SD & Change \\
\midrule
\addlinespace[2pt]\rowcolor[gray]{0.92}\multicolumn{9}{@{}l}{\rule{0pt}{11pt}\makebox[105pt][l]{\textbf{MUSIQ}}\hspace{12pt}\textbf{(SPAQ)}} \\
Frozen assessment model & 0.6954 &  & 0.6984 &  & 0.1824 &  & 0.2210 &  \\
Full & 0.9052\,\textpm\,0.0005 & \cellcolor[RGB]{255,255,255}0.00\% & 0.9005\,\textpm\,0.0005 & \cellcolor[RGB]{255,255,255}0.00\% & 0.0696\,\textpm\,0.0001 & \cellcolor[RGB]{255,255,255}0.00\% & 0.0893\,\textpm\,0.0002 & \cellcolor[RGB]{255,255,255}0.00\% \\
Final features only & 0.8690\,\textpm\,0.0007 & \cellcolor[RGB]{246,207,207}-4.01\% & 0.8674\,\textpm\,0.0007 & \cellcolor[RGB]{247,209,209}-3.68\% & 0.0818\,\textpm\,0.0003 & \cellcolor[RGB]{239,172,172}-17.49\% & 0.1060\,\textpm\,0.0003 & \cellcolor[RGB]{239,170,170}-18.69\% \\
Without final features & 0.8921\,\textpm\,0.0003 & \cellcolor[RGB]{249,220,220}-1.45\% & 0.8887\,\textpm\,0.0001 & \cellcolor[RGB]{249,221,221}-1.31\% & 0.0746\,\textpm\,0.0003 & \cellcolor[RGB]{244,196,196}-7.23\% & 0.0958\,\textpm\,0.0003 & \cellcolor[RGB]{244,196,196}-7.30\% \\
Without early features & 0.9021\,\textpm\,0.0005 & \cellcolor[RGB]{251,230,230}-0.35\% & 0.8976\,\textpm\,0.0005 & \cellcolor[RGB]{251,231,231}-0.32\% & 0.0705\,\textpm\,0.0002 & \cellcolor[RGB]{249,222,222}-1.23\% & 0.0908\,\textpm\,0.0002 & \cellcolor[RGB]{249,219,219}-1.70\% \\
Without middle features & 0.9013\,\textpm\,0.0004 & \cellcolor[RGB]{251,229,229}-0.44\% & 0.8966\,\textpm\,0.0004 & \cellcolor[RGB]{251,229,229}-0.44\% & 0.0707\,\textpm\,0.0001 & \cellcolor[RGB]{249,219,219}-1.63\% & 0.0912\,\textpm\,0.0001 & \cellcolor[RGB]{248,216,216}-2.10\% \\
Without late features & 0.9032\,\textpm\,0.0004 & \cellcolor[RGB]{251,232,232}-0.22\% & 0.8979\,\textpm\,0.0004 & \cellcolor[RGB]{251,231,231}-0.29\% & 0.0702\,\textpm\,0.0003 & \cellcolor[RGB]{250,225,225}-0.85\% & 0.0904\,\textpm\,0.0003 & \cellcolor[RGB]{249,222,222}-1.25\% \\
Without geometric metadata & 0.9052\,\textpm\,0.0005 & \cellcolor[RGB]{255,255,255}0.00\% & 0.9005\,\textpm\,0.0005 & \cellcolor[RGB]{255,255,255}0.00\% & 0.0696\,\textpm\,0.0001 & \cellcolor[RGB]{253,238,238}-0.02\% & 0.0893\,\textpm\,0.0002 & \cellcolor[RGB]{253,238,238}-0.01\% \\
Without original-score input & 0.9052\,\textpm\,0.0005 & \cellcolor[RGB]{255,255,255}0.00\% & 0.9005\,\textpm\,0.0006 & \cellcolor[RGB]{255,255,255}0.00\% & 0.0696\,\textpm\,0.0001 & \cellcolor[RGB]{253,239,239}-0.01\% & 0.0893\,\textpm\,0.0002 & \cellcolor[RGB]{253,239,239}-0.01\% \\
w/o geometric metadata \& score input & 0.9052\,\textpm\,0.0005 & \cellcolor[RGB]{255,255,255}0.00\% & 0.9004\,\textpm\,0.0005 & \cellcolor[RGB]{253,239,239}-0.01\% & 0.0696\,\textpm\,0.0001 & \cellcolor[RGB]{252,238,238}-0.02\% & 0.0893\,\textpm\,0.0002 & \cellcolor[RGB]{253,238,238}-0.02\% \\
w/o visual features & 0.7024\,\textpm\,0.0003 & \cellcolor[RGB]{237,163,163}-22.41\% & 0.7053\,\textpm\,0.0003 & \cellcolor[RGB]{238,164,164}-21.68\% & 0.1659\,\textpm\,0.0004 & \cellcolor[RGB]{235,151,151}-138.36\% & 0.2035\,\textpm\,0.0004 & \cellcolor[RGB]{235,151,151}-127.84\% \\
Without correlation loss & 0.9034\,\textpm\,0.0008 & \cellcolor[RGB]{252,233,233}-0.20\% & 0.8990\,\textpm\,0.0007 & \cellcolor[RGB]{252,233,233}-0.17\% & 0.0699\,\textpm\,0.0002 & \cellcolor[RGB]{251,230,230}-0.39\% & 0.0896\,\textpm\,0.0001 & \cellcolor[RGB]{251,232,232}-0.27\% \\
Without residual penalty & 0.9052\,\textpm\,0.0005 & \cellcolor[RGB]{255,255,255}0.00\% & 0.9005\,\textpm\,0.0006 & \cellcolor[RGB]{255,255,255}0.00\% & 0.0695\,\textpm\,0.0001 & \cellcolor[RGB]{231,245,233}+0.19\% & 0.0892\,\textpm\,0.0002 & \cellcolor[RGB]{232,245,234}+0.18\% \\
Bounded control (0.375) & 0.7550\,\textpm\,0.0020 & \cellcolor[RGB]{240,174,174}-16.59\% & 0.7604\,\textpm\,0.0019 & \cellcolor[RGB]{240,176,176}-15.55\% & 0.1525\,\textpm\,0.0001 & \cellcolor[RGB]{235,151,151}-119.15\% & 0.1906\,\textpm\,0.0001 & \cellcolor[RGB]{235,151,151}-113.41\% \\
Without hidden dropout & 0.9076\,\textpm\,0.0001 & \cellcolor[RGB]{230,244,232}+0.26\% & 0.9034\,\textpm\,0.0002 & \cellcolor[RGB]{229,244,231}+0.32\% & 0.0682\,\textpm\,0.0001 & \cellcolor[RGB]{213,237,218}+2.08\% & 0.0877\,\textpm\,0.0001 & \cellcolor[RGB]{215,238,220}+1.78\% \\
Feature-group dropout (0.4) & 0.8962\,\textpm\,0.0031 & \cellcolor[RGB]{250,224,224}-0.99\% & 0.8926\,\textpm\,0.0029 & \cellcolor[RGB]{250,225,225}-0.87\% & 0.0740\,\textpm\,0.0006 & \cellcolor[RGB]{245,199,199}-6.31\% & 0.0944\,\textpm\,0.0006 & \cellcolor[RGB]{245,201,201}-5.74\% \\
\addlinespace[2pt]\rowcolor[gray]{0.92}\multicolumn{9}{@{}l}{\rule{0pt}{11pt}\makebox[105pt][l]{\textbf{TOPIQ (Swin)}}\hspace{12pt}\textbf{(AVA)}} \\
Frozen assessment model & 0.7901 &  & 0.7910 &  & 0.3566 &  & 0.4614 &  \\
Full & 0.7939\,\textpm\,0.0001 & \cellcolor[RGB]{255,255,255}0.00\% & 0.7940\,\textpm\,0.0001 & \cellcolor[RGB]{255,255,255}0.00\% & 0.3525\,\textpm\,0.0001 & \cellcolor[RGB]{255,255,255}0.00\% & 0.4570\,\textpm\,0.0001 & \cellcolor[RGB]{255,255,255}0.00\% \\
Final features only & 0.7917\,\textpm\,0.0000 & \cellcolor[RGB]{251,231,231}-0.28\% & 0.7915\,\textpm\,0.0000 & \cellcolor[RGB]{251,231,231}-0.32\% & 0.3542\,\textpm\,0.0000 & \cellcolor[RGB]{251,228,228}-0.51\% & 0.4592\,\textpm\,0.0000 & \cellcolor[RGB]{251,229,229}-0.48\% \\
Without final features & 0.7940\,\textpm\,0.0001 & \cellcolor[RGB]{237,247,238}+0.01\% & 0.7947\,\textpm\,0.0001 & \cellcolor[RGB]{234,246,236}+0.08\% & 0.3524\,\textpm\,0.0000 & \cellcolor[RGB]{237,247,239}+0.01\% & 0.4569\,\textpm\,0.0001 & \cellcolor[RGB]{236,247,238}+0.02\% \\
Without early features & 0.7939\,\textpm\,0.0000 & \cellcolor[RGB]{253,239,239}-0.01\% & 0.7940\,\textpm\,0.0000 & \cellcolor[RGB]{255,255,255}0.00\% & 0.3525\,\textpm\,0.0000 & \cellcolor[RGB]{255,255,255}0.00\% & 0.4570\,\textpm\,0.0000 & \cellcolor[RGB]{253,238,238}-0.01\% \\
Without middle features & 0.7938\,\textpm\,0.0001 & \cellcolor[RGB]{253,238,238}-0.02\% & 0.7939\,\textpm\,0.0001 & \cellcolor[RGB]{253,238,238}-0.02\% & 0.3526\,\textpm\,0.0000 & \cellcolor[RGB]{252,237,237}-0.04\% & 0.4571\,\textpm\,0.0001 & \cellcolor[RGB]{252,237,237}-0.03\% \\
Without late features & 0.7924\,\textpm\,0.0000 & \cellcolor[RGB]{252,233,233}-0.19\% & 0.7922\,\textpm\,0.0000 & \cellcolor[RGB]{251,232,232}-0.23\% & 0.3536\,\textpm\,0.0000 & \cellcolor[RGB]{251,231,231}-0.31\% & 0.4585\,\textpm\,0.0000 & \cellcolor[RGB]{251,231,231}-0.33\% \\
Without geometric metadata & 0.7937\,\textpm\,0.0000 & \cellcolor[RGB]{252,237,237}-0.03\% & 0.7938\,\textpm\,0.0000 & \cellcolor[RGB]{252,237,237}-0.03\% & 0.3527\,\textpm\,0.0000 & \cellcolor[RGB]{252,236,236}-0.06\% & 0.4572\,\textpm\,0.0000 & \cellcolor[RGB]{252,237,237}-0.04\% \\
Without original-score input & 0.7939\,\textpm\,0.0001 & \cellcolor[RGB]{255,255,255}0.00\% & 0.7940\,\textpm\,0.0001 & \cellcolor[RGB]{255,255,255}0.00\% & 0.3525\,\textpm\,0.0001 & \cellcolor[RGB]{255,255,255}0.00\% & 0.4570\,\textpm\,0.0001 & \cellcolor[RGB]{255,255,255}0.00\% \\
w/o geometric metadata \& score input & 0.7937\,\textpm\,0.0000 & \cellcolor[RGB]{252,237,237}-0.03\% & 0.7938\,\textpm\,0.0000 & \cellcolor[RGB]{252,237,237}-0.03\% & 0.3527\,\textpm\,0.0000 & \cellcolor[RGB]{252,236,236}-0.06\% & 0.4572\,\textpm\,0.0000 & \cellcolor[RGB]{252,237,237}-0.04\% \\
w/o visual features & 0.7902\,\textpm\,0.0000 & \cellcolor[RGB]{251,229,229}-0.48\% & 0.7910\,\textpm\,0.0000 & \cellcolor[RGB]{251,230,230}-0.39\% & 0.3558\,\textpm\,0.0000 & \cellcolor[RGB]{250,224,224}-0.95\% & 0.4607\,\textpm\,0.0000 & \cellcolor[RGB]{250,225,225}-0.81\% \\
Without correlation loss & 0.7942\,\textpm\,0.0001 & \cellcolor[RGB]{236,247,237}+0.04\% & 0.7943\,\textpm\,0.0001 & \cellcolor[RGB]{236,247,237}+0.04\% & 0.3522\,\textpm\,0.0001 & \cellcolor[RGB]{234,246,236}+0.07\% & 0.4567\,\textpm\,0.0000 & \cellcolor[RGB]{235,246,236}+0.06\% \\
Without residual penalty & 0.7939\,\textpm\,0.0001 & \cellcolor[RGB]{255,255,255}0.00\% & 0.7940\,\textpm\,0.0001 & \cellcolor[RGB]{255,255,255}0.00\% & 0.3524\,\textpm\,0.0001 & \cellcolor[RGB]{255,255,255}0.00\% & 0.4570\,\textpm\,0.0001 & \cellcolor[RGB]{255,255,255}0.00\% \\
Bounded control (0.375) & 0.7939\,\textpm\,0.0001 & \cellcolor[RGB]{255,255,255}0.00\% & 0.7940\,\textpm\,0.0001 & \cellcolor[RGB]{255,255,255}0.00\% & 0.3525\,\textpm\,0.0001 & \cellcolor[RGB]{255,255,255}0.00\% & 0.4570\,\textpm\,0.0001 & \cellcolor[RGB]{255,255,255}0.00\% \\
Without hidden dropout & 0.7941\,\textpm\,0.0002 & \cellcolor[RGB]{237,247,238}+0.02\% & 0.7943\,\textpm\,0.0002 & \cellcolor[RGB]{236,247,238}+0.03\% & 0.3524\,\textpm\,0.0002 & \cellcolor[RGB]{237,247,238}+0.01\% & 0.4569\,\textpm\,0.0002 & \cellcolor[RGB]{236,247,238}+0.02\% \\
Feature-group dropout (0.4) & 0.7934\,\textpm\,0.0000 & \cellcolor[RGB]{252,236,236}-0.07\% & 0.7934\,\textpm\,0.0000 & \cellcolor[RGB]{252,235,235}-0.08\% & 0.3528\,\textpm\,0.0000 & \cellcolor[RGB]{252,235,235}-0.11\% & 0.4575\,\textpm\,0.0000 & \cellcolor[RGB]{252,235,235}-0.11\% \\
\bottomrule
\end{tabular*}
\endgroup
\end{table*}

\subsection{Original-Score Pathways}
\label{app:score-pathways}
The original score serves as both a head input and an additive skip. Let $g(\mathbf z)$ denote the averaged, scaled head output. Full predicts $q=q_0+g(\mathbf z)$ on the refinement score scale. Removing the head-score input masks only its normalized input coordinate and retains the skip. Removing the score skip predicts $q=g(\mathbf z)$ while retaining the head-score input. Removing both pathways predicts $q=g(\mathbf z)$ with the normalized original-score input coordinate also masked. All variants retain the same training objective, including $\lambda_{\mathrm{res}}\,n^{-1}\sum_i(q_i-q_{0,i})^2$, so the residual penalty remains referenced to the original score. The two no-skip variants use seed group A (Appendix~\ref{app:ablation-setup}), zero-initialized output layers, and 85 updates. Their initial prediction is zero on the refinement score scale, whereas the skip variants initially predict $q_0$.

\Cref{tab:score-paths-90} reports the mean and median relative changes from Full and the numbers of configurations where Full outperforms each of the three pathway-removal variants. Removing only the head-score input produces mean relative changes from $-0.037\%$ to $0.004\%$. Removing the skip reduces mean PLCC and SRCC by 15.751\% and 14.277\%, while increasing MAE and RMSE by 170.260\% and 154.029\%. Both no-skip variants increase the two error metrics in every ablation configuration. These comparisons show the value of retaining the existing prediction under the shared short training budget.

\subsection{Number of Residual Heads}
\label{app:head-count}
\Cref{tab:ensemble} reports the mean and sample standard deviation of the five single-head results, followed by metrics computed from three-, five-, and ten-head averaged predictions. Relative changes compare each single-head mean metric or ensemble metric with the default three-head Full. Green and red cells indicate improvement and decline, respectively. The three-head ensemble uses seeds 42, 17, and 2026. The five-head ensemble uses seed group A (42, 17, 2026, 7, and 123), and the ten-head ensemble uses groups A and B (Appendix~\ref{app:ablation-setup}). The single-head comparison uses the same five seeds as the component ablations. Ten-seed single-head statistics appear in \Cref{tab:head-architecture-1,tab:head-architecture-2}. All heads share the frozen assessment model's forward computation and cached training features. Only residual-head parameters are replicated. The reported parameter counts therefore separate the residual-head cost from the shared assessment model.

\begin{table*}[!tbp]
\centering
\caption{\textbf{Single-head variability and residual-head ensembles.} All refined rows use \method. One-head metrics are means $\pm$ sample standard deviations over five seeds. Ensembles average predictions. Changes are relative to the default three-head Full. Added parameters count residual heads. The frozen model is shared.}
\label{tab:ensemble}
\begingroup\fontsize{7.5}{9}\selectfont
\setlength{\tabcolsep}{1pt}
\begin{tabular*}{\textwidth}{@{\extracolsep{\fill}}l*{4}{r>{\fontsize{7}{9}\selectfont}r}r@{}}
\toprule
Model (Dataset) & \multicolumn{2}{c}{PLCC {\fontsize{7.5}{9}\selectfont$\uparrow$}} & \multicolumn{2}{c}{SRCC {\fontsize{7.5}{9}\selectfont$\uparrow$}} & \multicolumn{2}{c}{MAE {\fontsize{7.5}{9}\selectfont$\downarrow$}} & \multicolumn{2}{c}{RMSE {\fontsize{7.5}{9}\selectfont$\downarrow$}} & Added \\
\cmidrule(lr){2-3}\cmidrule(lr){4-5}\cmidrule(lr){6-7}\cmidrule(lr){8-9}
Heads & Score & Change & Score & Change & Score & Change & Score & Change & params \\
\midrule
\addlinespace[2pt]\rowcolor[gray]{0.92}\multicolumn{10}{@{}l}{\rule{0pt}{11pt}\makebox[105pt][l]{\textbf{TReS}}\hspace{12pt}\textbf{(FLIVE)}} \\
Base & 0.6398 &  & 0.6293 &  & 0.1317 &  & 0.1470 &  & 0 \\
1 & 0.7468\,\textpm\,0.0008 & \cellcolor[RGB]{252,236,236}-0.05\% & 0.7104\,\textpm\,0.0007 & \cellcolor[RGB]{252,236,236}-0.07\% & 0.0444\,\textpm\,0.0001 & \cellcolor[RGB]{252,235,235}-0.08\% & 0.0588\,\textpm\,0.0001 & \cellcolor[RGB]{252,235,235}-0.09\% & 377,153 \\
3 (Full) & 0.7472 & \cellcolor[RGB]{255,255,255}0.00\% & 0.7109 & \cellcolor[RGB]{255,255,255}0.00\% & 0.0444 & \cellcolor[RGB]{255,255,255}0.00\% & 0.0588 & \cellcolor[RGB]{255,255,255}0.00\% & 1,131,459 \\
5 & 0.7471 & \cellcolor[RGB]{253,239,239}-0.01\% & 0.7107 & \cellcolor[RGB]{253,238,238}-0.02\% & 0.0444 & \cellcolor[RGB]{255,255,255}0.00\% & 0.0588 & \cellcolor[RGB]{253,238,238}-0.01\% & 1,885,765 \\
10 & 0.7467 & \cellcolor[RGB]{252,236,236}-0.06\% & 0.7099 & \cellcolor[RGB]{252,234,234}-0.14\% & 0.0444 & \cellcolor[RGB]{235,246,237}+0.05\% & 0.0588 & \cellcolor[RGB]{237,247,238}+0.01\% & 3,771,530 \\
\addlinespace[2pt]\rowcolor[gray]{0.92}\multicolumn{10}{@{}l}{\rule{0pt}{11pt}\makebox[105pt][l]{\textbf{PIAA-MIR (Swin-T)}}\hspace{12pt}\textbf{(PARA GIAA)}} \\
Base & 0.7931 &  & 0.7236 &  & 0.0702 &  & 0.0925 &  & 0 \\
1 & 0.8976\,\textpm\,0.0002 & \cellcolor[RGB]{253,239,239}-0.01\% & 0.8561\,\textpm\,0.0003 & \cellcolor[RGB]{255,255,255}0.00\% & 0.0446\,\textpm\,0.0000 & \cellcolor[RGB]{252,236,236}-0.05\% & 0.0589\,\textpm\,0.0001 & \cellcolor[RGB]{252,237,237}-0.04\% & 92,481 \\
3 (Full) & 0.8977 & \cellcolor[RGB]{255,255,255}0.00\% & 0.8561 & \cellcolor[RGB]{255,255,255}0.00\% & 0.0446 & \cellcolor[RGB]{255,255,255}0.00\% & 0.0589 & \cellcolor[RGB]{255,255,255}0.00\% & 277,443 \\
5 & 0.8977 & \cellcolor[RGB]{255,255,255}0.00\% & 0.8562 & \cellcolor[RGB]{238,247,239}+0.01\% & 0.0446 & \cellcolor[RGB]{255,255,255}0.00\% & 0.0589 & \cellcolor[RGB]{255,255,255}0.00\% & 462,405 \\
10 & 0.8976 & \cellcolor[RGB]{253,239,239}-0.01\% & 0.8562 & \cellcolor[RGB]{237,247,239}+0.01\% & 0.0446 & \cellcolor[RGB]{252,237,237}-0.04\% & 0.0589 & \cellcolor[RGB]{252,237,237}-0.04\% & 924,810 \\
\addlinespace[2pt]\rowcolor[gray]{0.92}\multicolumn{10}{@{}l}{\rule{0pt}{11pt}\makebox[105pt][l]{\textbf{CLIP-IQA+ (L/14)}}\hspace{12pt}\textbf{(KonIQ-10k)}} \\
Base & 0.8747 &  & 0.8610 &  & 0.0602 &  & 0.0770 &  & 0 \\
1 & 0.9277\,\textpm\,0.0006 & \cellcolor[RGB]{252,235,235}-0.08\% & 0.9104\,\textpm\,0.0007 & \cellcolor[RGB]{252,235,235}-0.11\% & 0.0446\,\textpm\,0.0002 & \cellcolor[RGB]{251,228,228}-0.54\% & 0.0581\,\textpm\,0.0002 & \cellcolor[RGB]{251,229,229}-0.46\% & 344,385 \\
3 (Full) & 0.9285 & \cellcolor[RGB]{255,255,255}0.00\% & 0.9114 & \cellcolor[RGB]{255,255,255}0.00\% & 0.0444 & \cellcolor[RGB]{255,255,255}0.00\% & 0.0578 & \cellcolor[RGB]{255,255,255}0.00\% & 1,033,155 \\
5 & 0.9285 & \cellcolor[RGB]{238,247,239}+0.01\% & 0.9114 & \cellcolor[RGB]{255,255,255}0.00\% & 0.0443 & \cellcolor[RGB]{231,244,233}+0.21\% & 0.0577 & \cellcolor[RGB]{232,245,234}+0.16\% & 1,721,925 \\
10 & 0.9286 & \cellcolor[RGB]{237,247,238}+0.02\% & 0.9116 & \cellcolor[RGB]{236,247,237}+0.03\% & 0.0443 & \cellcolor[RGB]{230,244,232}+0.25\% & 0.0577 & \cellcolor[RGB]{230,244,233}+0.24\% & 3,443,850 \\
\addlinespace[2pt]\rowcolor[gray]{0.92}\multicolumn{10}{@{}l}{\rule{0pt}{11pt}\makebox[105pt][l]{\textbf{MUSIQ}}\hspace{12pt}\textbf{(SPAQ)}} \\
Base & 0.6954 &  & 0.6984 &  & 0.1824 &  & 0.2210 &  & 0 \\
1 & 0.9052\,\textpm\,0.0005 & \cellcolor[RGB]{252,236,236}-0.05\% & 0.9005\,\textpm\,0.0005 & \cellcolor[RGB]{252,237,237}-0.04\% & 0.0696\,\textpm\,0.0001 & \cellcolor[RGB]{252,234,234}-0.13\% & 0.0893\,\textpm\,0.0002 & \cellcolor[RGB]{252,236,236}-0.07\% & 196,929 \\
3 (Full) & 0.9057 & \cellcolor[RGB]{255,255,255}0.00\% & 0.9009 & \cellcolor[RGB]{255,255,255}0.00\% & 0.0695 & \cellcolor[RGB]{255,255,255}0.00\% & 0.0893 & \cellcolor[RGB]{255,255,255}0.00\% & 590,787 \\
5 & 0.9056 & \cellcolor[RGB]{253,238,238}-0.01\% & 0.9008 & \cellcolor[RGB]{255,255,255}0.00\% & 0.0695 & \cellcolor[RGB]{234,246,235}+0.09\% & 0.0892 & \cellcolor[RGB]{233,245,235}+0.11\% & 984,645 \\
10 & 0.9058 & \cellcolor[RGB]{237,247,239}+0.01\% & 0.9011 & \cellcolor[RGB]{236,247,238}+0.03\% & 0.0693 & \cellcolor[RGB]{230,244,232}+0.28\% & 0.0891 & \cellcolor[RGB]{231,245,234}+0.18\% & 1,969,290 \\
\addlinespace[2pt]\rowcolor[gray]{0.92}\multicolumn{10}{@{}l}{\rule{0pt}{11pt}\makebox[105pt][l]{\textbf{TOPIQ (Swin)}}\hspace{12pt}\textbf{(AVA)}} \\
Base & 0.7901 &  & 0.7910 &  & 0.3566 &  & 0.4614 &  & 0 \\
1 & 0.7939\,\textpm\,0.0001 & \cellcolor[RGB]{255,255,255}0.00\% & 0.7940\,\textpm\,0.0001 & \cellcolor[RGB]{253,239,239}-0.01\% & 0.3525\,\textpm\,0.0001 & \cellcolor[RGB]{253,238,238}-0.01\% & 0.4570\,\textpm\,0.0001 & \cellcolor[RGB]{253,239,239}-0.01\% & 147,777 \\
3 (Full) & 0.7940 & \cellcolor[RGB]{255,255,255}0.00\% & 0.7941 & \cellcolor[RGB]{255,255,255}0.00\% & 0.3524 & \cellcolor[RGB]{255,255,255}0.00\% & 0.4569 & \cellcolor[RGB]{255,255,255}0.00\% & 443,331 \\
5 & 0.7940 & \cellcolor[RGB]{255,255,255}0.00\% & 0.7941 & \cellcolor[RGB]{255,255,255}0.00\% & 0.3524 & \cellcolor[RGB]{255,255,255}0.00\% & 0.4570 & \cellcolor[RGB]{255,255,255}0.00\% & 738,885 \\
10 & 0.7940 & \cellcolor[RGB]{255,255,255}0.00\% & 0.7941 & \cellcolor[RGB]{255,255,255}0.00\% & 0.3524 & \cellcolor[RGB]{255,255,255}0.00\% & 0.4570 & \cellcolor[RGB]{255,255,255}0.00\% & 1,477,770 \\
\bottomrule
\end{tabular*}
\endgroup
\end{table*}

\subsection{Residual-Head Architecture}
\label{app:head-architecture}
\Cref{tab:head-architecture-1,tab:head-architecture-2} compare a linear head, single-hidden-layer MLPs of widths 32, 64, and 128, and a two-hidden-layer MLP with widths 64-64. All architectures use unbounded output. The linear head has no hidden activation or dropout. MLPs apply SiLU after each hidden linear layer, followed by dropout with probability $0.5$ during training only. All output layers are zero-initialized. Each row reports the mean and sample standard deviation of single-head metrics across seed groups A and B, with parameters counted for one head. The default unbounded output is $\gamma\eta$, with $\gamma=0.375$ and raw scalar head output $\eta$. The factor $\gamma$ fixes the correction scale without restricting the correction magnitude. The full head is the unbounded width-64 MLP.

The default width of 64 provides an efficient balance between predictive performance and head computation. In the three-head comparison in Table~3(F) of the main paper, it uses 1.70M dense MACs per image, half the 3.40M required by width 128, with approximately half the added parameters. Doubling the width gives mean relative improvements of 0.337\%, 0.228\%, 0.450\%, and 0.383\% in PLCC, SRCC, MAE, and RMSE, respectively. Therefore, width 64 retains performance close to the wider head at substantially lower computational and parameter cost.

\begin{table*}[!tbp]
\centering
\caption{\textbf{Residual-head architecture, part 1 of 2.} Single-head means $\pm$ sample standard deviations over ten seeds. Changes are relative to the Full mean over the same seeds. Parameters are for one head. Full is the unbounded 64-unit MLP. All rows use Huber regression.}
\label{tab:head-architecture-1}
\begingroup\fontsize{7}{9}\selectfont
\setlength{\tabcolsep}{2pt}
\begin{tabular*}{\textwidth}{@{\extracolsep{\fill}}lrrrrrrrrr@{}}
\toprule
Variant & \multicolumn{2}{c}{PLCC $\uparrow$} & \multicolumn{2}{c}{SRCC $\uparrow$} & \multicolumn{2}{c}{MAE $\downarrow$} & \multicolumn{2}{c}{RMSE $\downarrow$} & Added params \\
\midrule
\rowcolor[gray]{0.92}\multicolumn{10}{@{}l}{\textbf{TReS}\hspace{2em}\textbf{(FLIVE)}} \\
Linear, unbounded & 0.7590\,\textpm\,0.0000 & \cellcolor[RGB]{216,238,220}+1.70\% & 0.7320\,\textpm\,0.0000 & \cellcolor[RGB]{207,234,213}+3.18\% & 0.1251\,\textpm\,0.0000 & \cellcolor[RGB]{235,151,151}-181.50\% & 0.1377\,\textpm\,0.0000 & \cellcolor[RGB]{235,151,151}-134.04\% & 5,892 \\
MLP (32), unbounded & 0.7387\,\textpm\,0.0016 & \cellcolor[RGB]{250,224,224}-1.03\% & 0.7025\,\textpm\,0.0020 & \cellcolor[RGB]{250,224,224}-0.97\% & 0.0454\,\textpm\,0.0001 & \cellcolor[RGB]{248,217,217}-2.08\% & 0.0599\,\textpm\,0.0002 & \cellcolor[RGB]{249,218,218}-1.84\% & 188,577 \\
MLP (64), Full & 0.7463\,\textpm\,0.0010 & \cellcolor[RGB]{255,255,255}0.00\% & 0.7094\,\textpm\,0.0016 & \cellcolor[RGB]{255,255,255}0.00\% & 0.0444\,\textpm\,0.0000 & \cellcolor[RGB]{255,255,255}0.00\% & 0.0588\,\textpm\,0.0001 & \cellcolor[RGB]{255,255,255}0.00\% & 377,153 \\
MLP (128), unbounded & 0.7479\,\textpm\,0.0062 & \cellcolor[RGB]{231,244,233}+0.21\% & 0.7096\,\textpm\,0.0085 & \cellcolor[RGB]{236,247,238}+0.02\% & 0.0442\,\textpm\,0.0005 & \cellcolor[RGB]{227,243,230}+0.44\% & 0.0587\,\textpm\,0.0007 & \cellcolor[RGB]{230,244,233}+0.23\% & 754,305 \\
MLP (64--64), unbounded & 0.7398\,\textpm\,0.0016 & \cellcolor[RGB]{250,225,225}-0.87\% & 0.7129\,\textpm\,0.0010 & \cellcolor[RGB]{227,243,229}+0.49\% & 0.0531\,\textpm\,0.0010 & \cellcolor[RGB]{238,168,168}-19.50\% & 0.0686\,\textpm\,0.0011 & \cellcolor[RGB]{240,174,174}-16.55\% & 381,313 \\
\rowcolor[gray]{0.92}\multicolumn{10}{@{}l}{\textbf{PIAA-MIR (Swin-T)}\hspace{2em}\textbf{(PARA GIAA)}} \\
Linear, unbounded & 0.8899\,\textpm\,0.0000 & \cellcolor[RGB]{250,225,225}-0.85\% & 0.8504\,\textpm\,0.0000 & \cellcolor[RGB]{250,227,227}-0.66\% & 0.0518\,\textpm\,0.0000 & \cellcolor[RGB]{240,175,175}-15.99\% & 0.0677\,\textpm\,0.0000 & \cellcolor[RGB]{240,177,177}-14.81\% & 1,444 \\
MLP (32), unbounded & 0.8931\,\textpm\,0.0001 & \cellcolor[RGB]{251,229,229}-0.50\% & 0.8503\,\textpm\,0.0003 & \cellcolor[RGB]{250,227,227}-0.66\% & 0.0458\,\textpm\,0.0000 & \cellcolor[RGB]{248,214,214}-2.46\% & 0.0603\,\textpm\,0.0001 & \cellcolor[RGB]{248,216,216}-2.25\% & 46,241 \\
MLP (64), Full & 0.8975\,\textpm\,0.0002 & \cellcolor[RGB]{255,255,255}0.00\% & 0.8560\,\textpm\,0.0003 & \cellcolor[RGB]{255,255,255}0.00\% & 0.0447\,\textpm\,0.0001 & \cellcolor[RGB]{255,255,255}0.00\% & 0.0589\,\textpm\,0.0001 & \cellcolor[RGB]{255,255,255}0.00\% & 92,481 \\
MLP (128), unbounded & 0.9007\,\textpm\,0.0001 & \cellcolor[RGB]{228,243,231}+0.36\% & 0.8599\,\textpm\,0.0003 & \cellcolor[RGB]{227,243,230}+0.45\% & 0.0438\,\textpm\,0.0000 & \cellcolor[RGB]{215,237,219}+1.85\% & 0.0580\,\textpm\,0.0000 & \cellcolor[RGB]{216,238,221}+1.63\% & 184,961 \\
MLP (64--64), unbounded & 0.8993\,\textpm\,0.0004 & \cellcolor[RGB]{231,245,233}+0.19\% & 0.8580\,\textpm\,0.0003 & \cellcolor[RGB]{230,244,233}+0.23\% & 0.0444\,\textpm\,0.0001 & \cellcolor[RGB]{227,243,229}+0.49\% & 0.0586\,\textpm\,0.0001 & \cellcolor[RGB]{226,242,229}+0.57\% & 96,641 \\
\rowcolor[gray]{0.92}\multicolumn{10}{@{}l}{\textbf{CLIP-IQA+ (L/14)}\hspace{2em}\textbf{(KonIQ-10k)}} \\
Linear, unbounded & 0.9261\,\textpm\,0.0000 & \cellcolor[RGB]{252,233,233}-0.16\% & 0.9092\,\textpm\,0.0000 & \cellcolor[RGB]{252,234,234}-0.12\% & 0.0457\,\textpm\,0.0000 & \cellcolor[RGB]{248,216,216}-2.25\% & 0.0588\,\textpm\,0.0000 & \cellcolor[RGB]{249,222,222}-1.26\% & 5,380 \\
MLP (32), unbounded & 0.9248\,\textpm\,0.0006 & \cellcolor[RGB]{251,231,231}-0.30\% & 0.9065\,\textpm\,0.0007 & \cellcolor[RGB]{251,230,230}-0.41\% & 0.0456\,\textpm\,0.0002 & \cellcolor[RGB]{248,217,217}-2.02\% & 0.0591\,\textpm\,0.0003 & \cellcolor[RGB]{249,219,219}-1.66\% & 172,193 \\
MLP (64), Full & 0.9276\,\textpm\,0.0005 & \cellcolor[RGB]{255,255,255}0.00\% & 0.9103\,\textpm\,0.0007 & \cellcolor[RGB]{255,255,255}0.00\% & 0.0447\,\textpm\,0.0002 & \cellcolor[RGB]{255,255,255}0.00\% & 0.0581\,\textpm\,0.0002 & \cellcolor[RGB]{255,255,255}0.00\% & 344,385 \\
MLP (128), unbounded & 0.9282\,\textpm\,0.0006 & \cellcolor[RGB]{234,246,236}+0.07\% & 0.9109\,\textpm\,0.0010 & \cellcolor[RGB]{234,246,236}+0.07\% & 0.0444\,\textpm\,0.0003 & \cellcolor[RGB]{224,241,227}+0.69\% & 0.0577\,\textpm\,0.0003 & \cellcolor[RGB]{224,241,227}+0.71\% & 688,769 \\
MLP (64--64), unbounded & 0.9274\,\textpm\,0.0007 & \cellcolor[RGB]{252,237,237}-0.02\% & 0.9105\,\textpm\,0.0013 & \cellcolor[RGB]{236,247,238}+0.03\% & 0.0448\,\textpm\,0.0002 & \cellcolor[RGB]{251,232,232}-0.22\% & 0.0579\,\textpm\,0.0002 & \cellcolor[RGB]{229,244,231}+0.33\% & 348,545 \\
\bottomrule
\end{tabular*}
\endgroup
\end{table*}

\begin{table*}[!tbp]
\centering
\caption{\textbf{Residual-head architecture, part 2 of 2.} Single-head means $\pm$ sample standard deviations over ten seeds. Changes are relative to the Full mean over the same seeds. Parameters are for one head. Full is the unbounded 64-unit MLP. All rows use Huber regression.}
\label{tab:head-architecture-2}
\begingroup\fontsize{7}{9}\selectfont
\setlength{\tabcolsep}{2pt}
\begin{tabular*}{\textwidth}{@{\extracolsep{\fill}}lrrrrrrrrr@{}}
\toprule
Variant & \multicolumn{2}{c}{PLCC $\uparrow$} & \multicolumn{2}{c}{SRCC $\uparrow$} & \multicolumn{2}{c}{MAE $\downarrow$} & \multicolumn{2}{c}{RMSE $\downarrow$} & Added params \\
\midrule
\rowcolor[gray]{0.92}\multicolumn{10}{@{}l}{\textbf{MUSIQ}\hspace{2em}\textbf{(SPAQ)}} \\
Linear, unbounded & 0.9007\,\textpm\,0.0000 & \cellcolor[RGB]{251,228,228}-0.53\% & 0.8966\,\textpm\,0.0000 & \cellcolor[RGB]{251,229,229}-0.46\% & 0.1642\,\textpm\,0.0000 & \cellcolor[RGB]{235,151,151}-136.26\% & 0.1857\,\textpm\,0.0000 & \cellcolor[RGB]{235,151,151}-107.99\% & 3,076 \\
MLP (32), unbounded & 0.9003\,\textpm\,0.0011 & \cellcolor[RGB]{251,228,228}-0.57\% & 0.8952\,\textpm\,0.0010 & \cellcolor[RGB]{250,227,227}-0.61\% & 0.0719\,\textpm\,0.0002 & \cellcolor[RGB]{247,210,210}-3.47\% & 0.0926\,\textpm\,0.0003 & \cellcolor[RGB]{247,208,208}-3.76\% & 98,465 \\
MLP (64), Full & 0.9054\,\textpm\,0.0004 & \cellcolor[RGB]{255,255,255}0.00\% & 0.9008\,\textpm\,0.0005 & \cellcolor[RGB]{255,255,255}0.00\% & 0.0695\,\textpm\,0.0002 & \cellcolor[RGB]{255,255,255}0.00\% & 0.0893\,\textpm\,0.0002 & \cellcolor[RGB]{255,255,255}0.00\% & 196,929 \\
MLP (128), unbounded & 0.9078\,\textpm\,0.0003 & \cellcolor[RGB]{230,244,232}+0.26\% & 0.9035\,\textpm\,0.0002 & \cellcolor[RGB]{229,244,232}+0.31\% & 0.0687\,\textpm\,0.0001 & \cellcolor[RGB]{220,240,224}+1.16\% & 0.0880\,\textpm\,0.0001 & \cellcolor[RGB]{218,238,222}+1.47\% & 393,857 \\
MLP (64--64), unbounded & 0.9054\,\textpm\,0.0005 & \cellcolor[RGB]{253,239,239}-0.01\% & 0.9004\,\textpm\,0.0006 & \cellcolor[RGB]{252,237,237}-0.04\% & 0.0727\,\textpm\,0.0010 & \cellcolor[RGB]{246,205,205}-4.58\% & 0.0928\,\textpm\,0.0011 & \cellcolor[RGB]{247,208,208}-3.90\% & 201,089 \\
\rowcolor[gray]{0.92}\multicolumn{10}{@{}l}{\textbf{TOPIQ (Swin)}\hspace{2em}\textbf{(AVA)}} \\
Linear, unbounded & 0.7931\,\textpm\,0.0000 & \cellcolor[RGB]{252,235,235}-0.10\% & 0.7932\,\textpm\,0.0000 & \cellcolor[RGB]{252,235,235}-0.11\% & 0.3530\,\textpm\,0.0000 & \cellcolor[RGB]{252,233,233}-0.17\% & 0.4578\,\textpm\,0.0000 & \cellcolor[RGB]{252,233,233}-0.18\% & 2,308 \\
MLP (32), unbounded & 0.7936\,\textpm\,0.0000 & \cellcolor[RGB]{252,237,237}-0.04\% & 0.7937\,\textpm\,0.0000 & \cellcolor[RGB]{252,237,237}-0.04\% & 0.3527\,\textpm\,0.0000 & \cellcolor[RGB]{252,235,235}-0.08\% & 0.4573\,\textpm\,0.0000 & \cellcolor[RGB]{252,236,236}-0.07\% & 73,889 \\
MLP (64), Full & 0.7939\,\textpm\,0.0001 & \cellcolor[RGB]{255,255,255}0.00\% & 0.7941\,\textpm\,0.0001 & \cellcolor[RGB]{255,255,255}0.00\% & 0.3524\,\textpm\,0.0001 & \cellcolor[RGB]{255,255,255}0.00\% & 0.4570\,\textpm\,0.0001 & \cellcolor[RGB]{255,255,255}0.00\% & 147,777 \\
MLP (128), unbounded & 0.7942\,\textpm\,0.0002 & \cellcolor[RGB]{236,247,237}+0.03\% & 0.7943\,\textpm\,0.0002 & \cellcolor[RGB]{236,246,237}+0.04\% & 0.3522\,\textpm\,0.0003 & \cellcolor[RGB]{235,246,236}+0.06\% & 0.4567\,\textpm\,0.0002 & \cellcolor[RGB]{235,246,236}+0.05\% & 295,553 \\
MLP (64--64), unbounded & 0.7938\,\textpm\,0.0001 & \cellcolor[RGB]{253,238,238}-0.02\% & 0.7939\,\textpm\,0.0001 & \cellcolor[RGB]{252,237,237}-0.02\% & 0.3526\,\textpm\,0.0001 & \cellcolor[RGB]{252,237,237}-0.03\% & 0.4571\,\textpm\,0.0001 & \cellcolor[RGB]{252,237,237}-0.03\% & 151,937 \\
\bottomrule
\end{tabular*}
\endgroup
\end{table*}

\subsection{Correction-Bound Sensitivity}
\label{app:correction-bounds}
We compare the unbounded full head with fixed bounds $\tau\in\{0.1875,0.375,0.75,1.5,3.375\}$ and an adaptive bound obtained from the 90th percentile of the frozen model's absolute training errors on the refinement score scale, clipped to $[0.375,3.375]$. To vary $\tau$ while keeping the correction derivative with respect to $\eta$ equal to $\gamma$ at $\eta=0$, bounded corrections use $\tau\tanh(\gamma\eta/\tau)$ with $\gamma=0.375$. The unbounded correction $\gamma\eta$ is the limit as $\tau\to\infty$. The adaptive bound is computed once per configuration and shared by all images in that configuration.

\Cref{tab:bound-loss-90} evaluates seven residual-output settings with Huber regression under the same evaluation settings as Table~3 of the main paper. The settings comprise the five fixed bounds, the adaptive bound, and unbounded output. Each setting uses five initializations, the same frozen representations, and 85 updates. The primary three-head ensemble uses seeds 42, 17, and 2026. All other objective terms remain fixed. The table reports configuration-wise relative changes from Full, summarized by their equally weighted mean and median.

For the bounded Huber control with $\tau=0.375$, mean relative changes from Full in PLCC, SRCC, MAE, and RMSE are $-1.38\%$, $-1.35\%$, $-45.27\%$, and $-40.41\%$, with medians of $-0.30\%$, $-0.20\%$, $-3.46\%$, and $-3.06\%$. Negative values indicate lower correlations or higher errors than Full. Full outperforms this control in 71, 64, 69, and 67 configurations, respectively, and on all four metrics in 56 configurations. The control outperforms Full on all four metrics in 15 configurations.

\begin{table*}[!tbp]
\centering
\caption{\textbf{Residual-bound comparison under the same evaluation settings as Table~3 of the main paper.} Three-head results with Huber regression. Each cell gives the equally weighted mean (median) relative change from Full. Positive means improvement for every metric. Green/red shading indicates mean improvement/decline. Full wins counts configurations where Full outperforms the variant on all four metrics. Full is bold.}
\label{tab:bound-loss-90}
\begingroup\fontsize{7}{9}\selectfont
\setlength{\tabcolsep}{3pt}
\begin{tabular*}{\textwidth}{@{\extracolsep{\fill}}llrrrrr@{}}
\toprule
\rowcolor[gray]{0.92}Bound & Regression & PLCC $\uparrow$ & SRCC $\uparrow$ & MAE $\downarrow$ & RMSE $\downarrow$ & Full wins \\
\midrule
0.1875 & Huber & \cellcolor[RGB]{248,216,216}-2.25\% (-0.55\%) & \cellcolor[RGB]{248,215,215}-2.29\% (-0.40\%) & \cellcolor[RGB]{235,151,151}-54.97\% (-8.01\%) & \cellcolor[RGB]{235,151,151}-48.18\% (-7.08\%) & 61 \\
0.375 & Huber & \cellcolor[RGB]{249,221,221}-1.38\% (-0.30\%) & \cellcolor[RGB]{249,221,221}-1.35\% (-0.20\%) & \cellcolor[RGB]{235,151,151}-45.27\% (-3.46\%) & \cellcolor[RGB]{235,151,151}-40.41\% (-3.06\%) & 56 \\
0.75 & Huber & \cellcolor[RGB]{250,227,227}-0.65\% (-0.17\%) & \cellcolor[RGB]{250,226,226}-0.75\% (-0.05\%) & \cellcolor[RGB]{235,151,151}-31.74\% (-1.02\%) & \cellcolor[RGB]{235,152,152}-29.12\% (-0.55\%) & 56 \\
1.5 & Huber & \cellcolor[RGB]{227,243,229}+0.48\% (-0.03\%) & \cellcolor[RGB]{226,242,229}+0.53\% (-0.01\%) & \cellcolor[RGB]{240,173,173}-16.76\% (-0.08\%) & \cellcolor[RGB]{240,174,174}-16.25\% (-0.13\%) & 47 \\
3.375 & Huber & \cellcolor[RGB]{214,237,218}+2.04\% (0.00\%) & \cellcolor[RGB]{216,238,220}+1.73\% (0.00\%) & \cellcolor[RGB]{244,197,197}-7.13\% (0.00\%) & \cellcolor[RGB]{244,195,195}-7.51\% (-0.01\%) & 35 \\
Adaptive (90th percentile) & Huber & \cellcolor[RGB]{214,237,218}+2.02\% (-0.03\%) & \cellcolor[RGB]{216,238,220}+1.73\% (-0.02\%) & \cellcolor[RGB]{244,196,196}-7.44\% (-0.06\%) & \cellcolor[RGB]{244,194,194}-7.85\% (-0.08\%) & 41 \\
\textbf{No bound (Full)} & \textbf{Huber} & \textbf{\cellcolor[RGB]{255,255,255}0.00\% (0.00\%)} & \textbf{\cellcolor[RGB]{255,255,255}0.00\% (0.00\%)} & \textbf{\cellcolor[RGB]{255,255,255}0.00\% (0.00\%)} & \textbf{\cellcolor[RGB]{255,255,255}0.00\% (0.00\%)} & \textbf{0} \\
\bottomrule
\end{tabular*}
\endgroup
\end{table*}

\subsection{Stability and Uncertainty}
\label{app:stability-uncertainty}
The Full rows in \Cref{tab:head-architecture-1,tab:head-architecture-2} report single-head metric means and sample standard deviations across all ten seeds in groups A and B (Appendix~\ref{app:ablation-setup}), quantifying initialization variability. Multi-head metrics are computed from averaged predictions.

\Cref{tab:paired-uncertainty} compares the ten-head ensemble with the default three-head Full. Images with identical pixel content form one sampling group. Each interval uses 2,000 paired resamples with candidate and Full predictions evaluated on the same sampled units: 1,826 reference contents for TReS (FLIVE), 2,952 pixel groups for PIAA-MIR (PARA), 2,015 pixel groups for CLIP-IQA+ (KonIQ-10k), 2,114 pixel groups for MUSIQ (SPAQ), and 19,929 images for TOPIQ (AVA). For each metric, the table reports a 95\% bootstrap percentile interval for the absolute change, without adjustment for multiple comparisons: candidate minus Full for PLCC/SRCC and Full minus candidate for MAE/RMSE. Relative percentages use the observed Full metric as the denominator. These intervals measure evaluation-sample uncertainty, distinct from the initialization variability in \Cref{tab:head-architecture-1,tab:head-architecture-2}.

\begin{table*}[!tbp]
\centering
\caption{\textbf{Paired uncertainty of the head-count comparison.} Ten-head relative changes and 95\% bootstrap confidence intervals for absolute changes from the default three-head Full, computed separately for each metric. Positive values mean improvement for every metric. Intervals use 2,000 paired resamples, without adjustment for multiple comparisons. Independent units: 1,826 reference-ID groups (TReS), 2,952 groups of images with identical pixels (PIAA-MIR), 2,015 groups of images with identical pixels (CLIP-IQA+), 2,114 groups of images with identical pixels (MUSIQ), and 19,929 images (TOPIQ).}
\label{tab:paired-uncertainty}
\begingroup\fontsize{7}{9}\selectfont
\setlength{\tabcolsep}{2pt}
\begin{tabular*}{\textwidth}{@{\extracolsep{\fill}}lrrrrrrrr@{}}
\toprule
Variant & \multicolumn{2}{c}{PLCC $\uparrow$} & \multicolumn{2}{c}{SRCC $\uparrow$} & \multicolumn{2}{c}{MAE $\downarrow$} & \multicolumn{2}{c}{RMSE $\downarrow$} \\
Heads & Change & 95\% interval & Change & 95\% interval & Change & 95\% interval & Change & 95\% interval \\
\midrule
\rowcolor[gray]{0.92}\multicolumn{9}{@{}l}{\textbf{TReS}\hspace{2em}\textbf{(FLIVE)}} \\
10 & \cellcolor[RGB]{252,236,236}-0.06\% & [-0.00068, -0.00015] & \cellcolor[RGB]{252,234,234}-0.14\% & [-0.00132, -0.00068] & \cellcolor[RGB]{235,246,237}+0.05\% & [-0.00001, +0.00005] & \cellcolor[RGB]{237,247,238}+0.01\% & [-0.00002, +0.00004] \\
\rowcolor[gray]{0.92}\multicolumn{9}{@{}l}{\textbf{PIAA-MIR (Swin-T)}\hspace{2em}\textbf{(PARA GIAA)}} \\
10 & \cellcolor[RGB]{253,239,239}-0.01\% & [-0.00019, +0.00009] & \cellcolor[RGB]{237,247,239}+0.01\% & [-0.00018, +0.00032] & \cellcolor[RGB]{252,237,237}-0.04\% & [-0.00005, +0.00001] & \cellcolor[RGB]{252,237,237}-0.04\% & [-0.00006, +0.00001] \\
\rowcolor[gray]{0.92}\multicolumn{9}{@{}l}{\textbf{CLIP-IQA+ (L/14)}\hspace{2em}\textbf{(KonIQ-10k)}} \\
10 & \cellcolor[RGB]{237,247,238}+0.02\% & [-0.00023, +0.00053] & \cellcolor[RGB]{236,247,237}+0.03\% & [-0.00035, +0.00085] & \cellcolor[RGB]{230,244,232}+0.25\% & [-0.00004, +0.00025] & \cellcolor[RGB]{230,244,233}+0.24\% & [-0.00001, +0.00029] \\
\rowcolor[gray]{0.92}\multicolumn{9}{@{}l}{\textbf{MUSIQ}\hspace{2em}\textbf{(SPAQ)}} \\
10 & \cellcolor[RGB]{237,247,239}+0.01\% & [-0.00017, +0.00035] & \cellcolor[RGB]{236,247,238}+0.03\% & [-0.00008, +0.00056] & \cellcolor[RGB]{230,244,232}+0.28\% & [+0.00008, +0.00031] & \cellcolor[RGB]{231,245,234}+0.18\% & [+0.00004, +0.00029] \\
\rowcolor[gray]{0.92}\multicolumn{9}{@{}l}{\textbf{TOPIQ (Swin)}\hspace{2em}\textbf{(AVA)}} \\
10 & \cellcolor[RGB]{255,255,255}0.00\% & [-0.00005, +0.00001] & \cellcolor[RGB]{255,255,255}0.00\% & [-0.00006, +0.00001] & \cellcolor[RGB]{255,255,255}0.00\% & [-0.00004, +0.00002] & \cellcolor[RGB]{255,255,255}0.00\% & [-0.00004, +0.00002] \\
\bottomrule
\end{tabular*}
\endgroup
\end{table*}

\subsection{Performance Decreases and Metric Trade-offs}
\label{app:performance-decreases}
The decreases in Tables~1 and 2 of the main paper include distinct patterns. For WaDIQaM-NR on KonIQ-10k, MAE decreases from 0.8095 to 0.2740, while SRCC decreases from 0.8046 to 0.3557. SIND on KADID-10k shows the same direction of change: MAE decreases from 1.5902 to 0.2740, while SRCC decreases from 0.8606 to 0.7634. These results separate absolute score accuracy from ranking accuracy. Corrections can bring predictions closer to the target scores while changing the ordering of images. A positive affine rescaling alone preserves both PLCC and SRCC, so the correlation decreases reflect changes beyond global score rescaling.

Charm on BAID shows the reverse pattern: PLCC rises from 0.3833 to 0.3944 and SRCC from 0.3731 to 0.3778, while MAE rises from 0.0463 to 0.0723 and RMSE from 0.0633 to 0.0880. Better correlation can coexist with larger absolute errors because correlation measures agreement in variation or ordering rather than numerical proximity to the labels. LAR-IQA KAN on UHD-IQA instead declines on all four metrics. In this case, the learned corrections worsen both absolute score accuracy and correlation on the evaluation images.

The ablations also expose this tension between correlation and absolute error. In Table~3 of the main paper, linear heads improve mean PLCC and SRCC relative to Full but increase MAE and RMSE. A correction bound of 3.375 produces the same directional trade-off (\Cref{tab:bound-loss-90}). These comparisons show that the head architecture and correction parameterization affect the balance between correlation and absolute score accuracy.
\section{Efficiency and Cache Costs}
\label{app:efficiency}

\begin{table*}[!tbp]
\centering
\caption{\textbf{End-to-end inference efficiency on an RTX 3090.} Batch size 1, 24 timed image calls per setting. $K=0$ denotes the frozen model without refinement. Memory is peak PyTorch allocated/reserved memory, including the assessment model and residual heads on the GPU. Green/red cells indicate lower/higher cost or higher/lower throughput relative to each model's $K=0$. Saturation reflects the relative change.}
\label{tab:inference-efficiency}
\begingroup\fontsize{8}{9}\selectfont
\setlength{\tabcolsep}{2pt}
\begin{tabular*}{\textwidth}{@{\extracolsep{\fill}}lrrrr@{}}
\toprule
\rowcolor[gray]{0.92}Heads & Median ms/image {\fontsize{7.5}{9}\selectfont$\downarrow$} & Images/s {\fontsize{7.5}{9}\selectfont$\uparrow$} & Allocated MiB {\fontsize{7.5}{9}\selectfont$\downarrow$} & Reserved MiB {\fontsize{7.5}{9}\selectfont$\downarrow$} \\
\midrule
\rowcolor[gray]{0.92}\multicolumn{5}{@{}l}{\textbf{Charm}\hspace{2em}\textbf{(AVA)}} \\
0 & \cellcolor[gray]{0.97}72.94 & \cellcolor[gray]{0.97}13.36 & \cellcolor[gray]{0.97}1223.93 & \cellcolor[gray]{0.97}1250.00 \\
1 & \cellcolor[RGB]{245,207,207}74.83 & \cellcolor[RGB]{246,209,209}13.03 & \cellcolor[RGB]{250,225,225}1231.95 & \cellcolor[RGB]{230,243,232}1248.00 \\
3 & \cellcolor[RGB]{244,203,203}75.40 & \cellcolor[RGB]{249,224,224}13.26 & \cellcolor[RGB]{249,221,221}1235.96 & \cellcolor[RGB]{248,218,218}1266.00 \\
5 & \cellcolor[RGB]{244,202,202}75.49 & \cellcolor[RGB]{246,211,211}13.07 & \cellcolor[RGB]{248,217,217}1239.96 & \cellcolor[RGB]{248,216,216}1268.00 \\
10 & \cellcolor[RGB]{242,192,192}76.86 & \cellcolor[RGB]{242,196,196}12.74 & \cellcolor[RGB]{246,211,211}1249.97 & \cellcolor[RGB]{248,216,216}1268.00 \\
\rowcolor[gray]{0.92}\multicolumn{5}{@{}l}{\textbf{CLIP-IQA+ (L/14)}\hspace{2em}\textbf{(KonIQ-10k)}} \\
0 & \cellcolor[gray]{0.97}82.08 & \cellcolor[gray]{0.97}12.19 & \cellcolor[gray]{0.97}1721.91 & \cellcolor[gray]{0.97}1790.00 \\
1 & \cellcolor[RGB]{246,209,209}84.05 & \cellcolor[RGB]{247,214,214}11.98 & \cellcolor[RGB]{252,235,235}1723.93 & \cellcolor[gray]{0.97}1790.00 \\
3 & \cellcolor[RGB]{245,207,207}84.31 & \cellcolor[RGB]{246,209,209}11.91 & \cellcolor[RGB]{251,229,229}1727.94 & \cellcolor[gray]{0.97}1790.00 \\
5 & \cellcolor[RGB]{243,200,200}85.23 & \cellcolor[RGB]{244,202,202}11.77 & \cellcolor[RGB]{250,226,226}1731.94 & \cellcolor[gray]{0.97}1790.00 \\
10 & \cellcolor[RGB]{241,190,190}86.88 & \cellcolor[RGB]{241,191,191}11.50 & \cellcolor[RGB]{248,219,219}1741.95 & \cellcolor[gray]{0.97}1790.00 \\
\rowcolor[gray]{0.92}\multicolumn{5}{@{}l}{\textbf{NLNet}\hspace{2em}\textbf{(KADID-10k)}} \\
0 & \cellcolor[gray]{0.97}871.07 & \cellcolor[gray]{0.97}1.15 & \cellcolor[gray]{0.97}212.35 & \cellcolor[gray]{0.97}290.00 \\
1 & \cellcolor[RGB]{203,230,209}850.22 & \cellcolor[RGB]{198,228,205}1.18 & \cellcolor[RGB]{252,233,233}212.70 & \cellcolor[gray]{0.97}290.00 \\
3 & \cellcolor[RGB]{190,224,197}832.63 & \cellcolor[RGB]{194,226,201}1.19 & \cellcolor[RGB]{250,227,227}213.39 & \cellcolor[gray]{0.97}290.00 \\
5 & \cellcolor[RGB]{199,229,205}845.56 & \cellcolor[RGB]{251,231,231}1.15 & \cellcolor[RGB]{249,223,223}214.09 & \cellcolor[RGB]{250,224,224}292.00 \\
10 & \cellcolor[RGB]{186,222,193}824.99 & \cellcolor[RGB]{200,229,206}1.18 & \cellcolor[RGB]{247,215,215}215.82 & \cellcolor[RGB]{248,217,217}294.00 \\
\rowcolor[gray]{0.92}\multicolumn{5}{@{}l}{\textbf{MUSIQ}\hspace{2em}\textbf{(SPAQ)}} \\
0 & \cellcolor[gray]{0.97}2141.81 & \cellcolor[gray]{0.97}0.50 & \cellcolor[gray]{0.97}3277.91 & \cellcolor[gray]{0.97}6074.00 \\
1 & \cellcolor[RGB]{229,243,232}2138.16 & \cellcolor[RGB]{232,244,234}0.50 & \cellcolor[RGB]{253,239,239}3278.66 & \cellcolor[gray]{0.97}6074.00 \\
3 & \cellcolor[RGB]{242,195,195}2244.82 & \cellcolor[RGB]{250,225,225}0.49 & \cellcolor[RGB]{253,236,236}3280.17 & \cellcolor[RGB]{253,238,238}6076.00 \\
5 & \cellcolor[RGB]{248,219,219}2165.35 & \cellcolor[RGB]{251,230,230}0.49 & \cellcolor[RGB]{252,235,235}3281.67 & \cellcolor[RGB]{253,236,236}6078.00 \\
10 & \cellcolor[RGB]{251,231,231}2147.71 & \cellcolor[RGB]{251,229,229}0.49 & \cellcolor[RGB]{251,232,232}3285.43 & \cellcolor[RGB]{252,234,234}6082.00 \\
\rowcolor[gray]{0.92}\multicolumn{5}{@{}l}{\textbf{FGResQ}\hspace{2em}\textbf{(FGRestore)}} \\
0 & \cellcolor[gray]{0.97}30.70 & \cellcolor[gray]{0.97}32.33 & \cellcolor[gray]{0.97}688.46 & \cellcolor[gray]{0.97}770.00 \\
1 & \cellcolor[RGB]{239,183,183}33.01 & \cellcolor[RGB]{242,192,192}30.60 & \cellcolor[RGB]{251,228,228}691.31 & \cellcolor[RGB]{245,207,207}790.00 \\
3 & \cellcolor[RGB]{240,185,185}32.87 & \cellcolor[RGB]{242,194,194}30.71 & \cellcolor[RGB]{248,218,218}697.25 & \cellcolor[RGB]{245,207,207}790.00 \\
5 & \cellcolor[RGB]{239,181,181}33.16 & \cellcolor[RGB]{243,198,198}30.95 & \cellcolor[RGB]{246,211,211}702.88 & \cellcolor[RGB]{245,207,207}790.00 \\
10 & \cellcolor[RGB]{230,146,146}37.67 & \cellcolor[RGB]{230,146,146}25.33 & \cellcolor[RGB]{243,198,198}717.26 & \cellcolor[RGB]{242,193,193}810.00 \\
\bottomrule
\end{tabular*}
\endgroup
\end{table*}
\begin{table*}[!tbp]
\centering
\caption{\textbf{Residual-head training and cache costs.} One full-batch 85-update head fit, cache reading, and feature normalization with GPU transfer are timed separately. Cache size counts the stored files. Training-feature size counts only the visual feature values used by the full head, at four bytes per value.}
\label{tab:training-cache}
\begingroup\fontsize{7}{9}\selectfont
\setlength{\tabcolsep}{2pt}
\begin{tabular*}{\textwidth}{@{\extracolsep{\fill}}lrrrrrrr@{}}
\toprule
\rowcolor[gray]{0.92}Model (Dataset) & Train images & Fit (s) & Load (s) & Norm./GPU (s) & Cache GiB & Train feat. GiB & Extract/write 8 (s) \\
\midrule
\rowcolor[gray]{0.97}\cellcolor[gray]{0.92}Charm (AVA) & 235579 & 5.287 & 0.196 & 57.464 & 10.808 & 7.189 & 0.608/0.003 \\
\cellcolor[gray]{0.92}CLIP-IQA+ (L/14) (KonIQ-10k) & 7058 & 0.843 & 1.147 & 0.278 & 0.205 & 0.141 & 0.669/0.003 \\
\rowcolor[gray]{0.97}\cellcolor[gray]{0.92}NLNet (KADID-10k) & 6000 & 0.963 & 0.357 & 0.051 & 0.056 & 0.031 & 6.682/0.004 \\
\cellcolor[gray]{0.92}MUSIQ (SPAQ) & 7910 & 0.790 & 0.472 & 0.136 & 0.131 & 0.091 & 16.184/0.003 \\
\rowcolor[gray]{0.97}\cellcolor[gray]{0.92}FGResQ (FGRestore) & 14700 & 0.878 & 3.874 & 1.043 & 0.796 & 0.631 & 0.241/0.003 \\
\bottomrule
\end{tabular*}
\endgroup
\end{table*}

\subsection{Complete Inference Pipeline}
\label{app:efficiency-pipeline}
\Cref{tab:inference-efficiency} compares the frozen model with one-, three-, five-, and ten-head refinement on an NVIDIA GeForce RTX~3090 with 24\,GB memory. We measure efficiency for Charm (AVA), CLIP-IQA+ (KonIQ-10k), NLNet (KADID-10k), MUSIQ (SPAQ), and FGResQ (FGRestore). Batch size is one. Timing includes image reading and decoding, native preprocessing, device transfer, original model inference, required feature extraction, and score transfer to CPU. The original model's crops, scales, or viewports are preserved. All residual heads reuse that scoring pipeline. Baseline measurements compute original scores without extracting the additional features used by the residual heads. Refinement timing also includes CPU feature export, per-head prediction, and the associated CPU/GPU data transfers. We use no compilation, mixed precision, head fusion, or multi-GPU execution.

\subsection{Timing and Memory Protocol}
\label{app:timing-memory}
We use eight fixed training images and three randomly interleaved measurement rounds. Each baseline or residual-head-count setting receives two warm-up calls per round followed by eight timed calls, for 24 timed calls per setting. We report median latency and images per second computed as the number of calls divided by total elapsed time. The images have already been read before timing, allowing subsequent reads to use the operating system's file cache. PyTorch measures peak allocated memory (occupied by tensors) and peak reserved memory (held by its memory allocator), including the assessment model and residual heads on the GPU. These memory measurements exclude driver context and other processes.

\subsection{Training and Caching}
\label{app:training-cache}
We measure training time for one single-head run with seed 314, using the full training cache and 85 updates (\Cref{tab:training-cache}). The timing includes pair generation and optimization. Cache loading is timed separately from feature normalization and transfer to the GPU. Stored cache size counts the retained feature-cache files and may include representations unused by the full head. Training-feature storage is calculated as the number of training images times the visual feature dimension times four bytes per FP32 value, counting only features used by the full residual head. For Charm, the initial loading stage opens the feature files and reads their metadata. Feature values are read on access. Their reading time is included in feature normalization and GPU transfer. Cache-write timing includes writing to the operating system's file cache, without waiting for all data to reach the disk.
\section{Qualitative Score Refinement}
\label{app:qualitative}

\subsection{Cross-Dataset Qualitative Comparisons}
\label{app:qualitative-examples}

\begin{figure*}[!tbp]
\centering
\includegraphics[page=1,width=\textwidth]{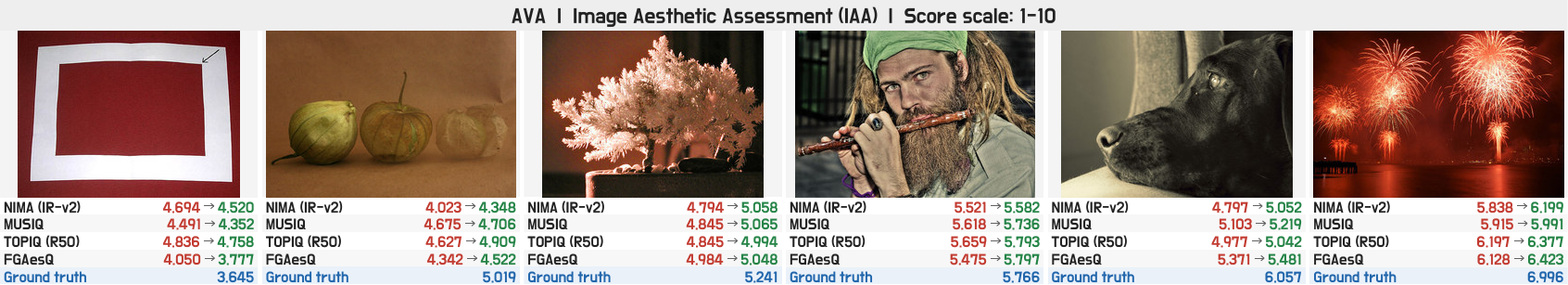}
\caption{\textbf{Qualitative score refinement on AVA.} Original scores (red) and scores after \method (green) are shown together for each assessment model. Ground truth is blue. Scores use the AVA 1--10 scale.}
\label{fig:qualitative-ava}
\end{figure*}

\begin{figure*}[!tbp]
\centering
\includegraphics[page=2,width=\textwidth]{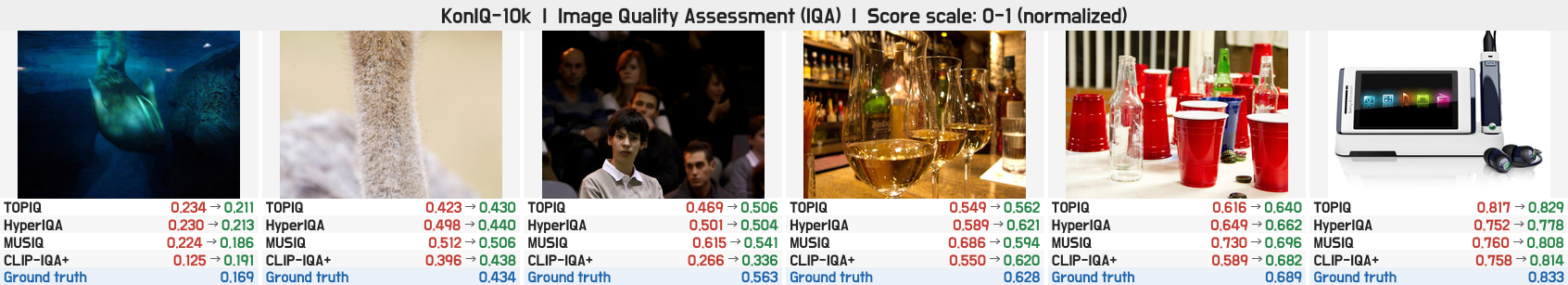}
\caption{\textbf{Qualitative score refinement on KonIQ-10k.} Scores use the normalized 0--1 evaluation scale. Colors and paired-score notation follow \Cref{fig:qualitative-ava}.}
\label{fig:qualitative-koniq}
\end{figure*}

\begin{figure*}[!tbp]
\centering
\includegraphics[page=3,width=\textwidth]{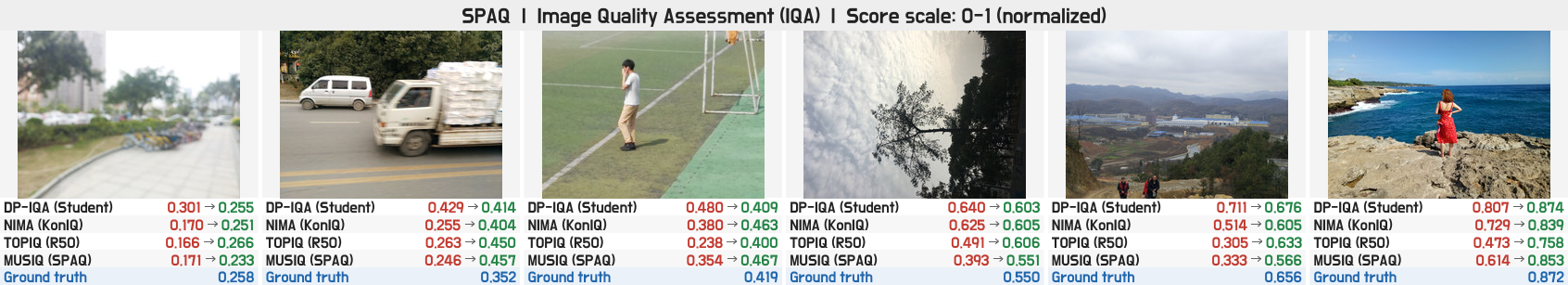}
\caption{\textbf{Qualitative score refinement on SPAQ.} Scores use the normalized 0--1 evaluation scale. Each model retains its own original scoring pipeline. Colors and notation follow \Cref{fig:qualitative-ava}.}
\label{fig:qualitative-spaq}
\end{figure*}

\begin{figure*}[!tbp]
\centering
\includegraphics[page=4,width=\textwidth]{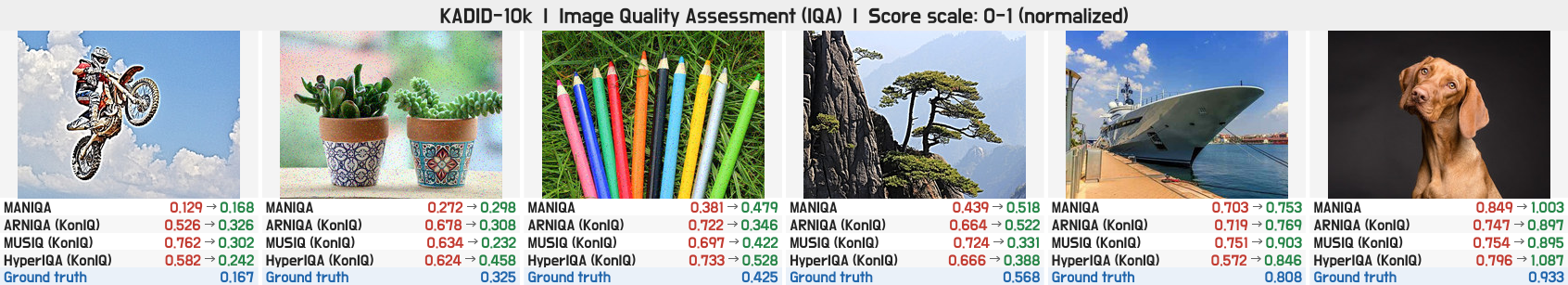}
\caption{\textbf{Qualitative score refinement on KADID-10k.} Examples span multiple contents and distortion types on the normalized 0--1 scale. Colors and notation follow \Cref{fig:qualitative-ava}.}
\label{fig:qualitative-kadid}
\end{figure*}

\begin{figure*}[!tbp]
\centering
\includegraphics[page=5,width=\textwidth]{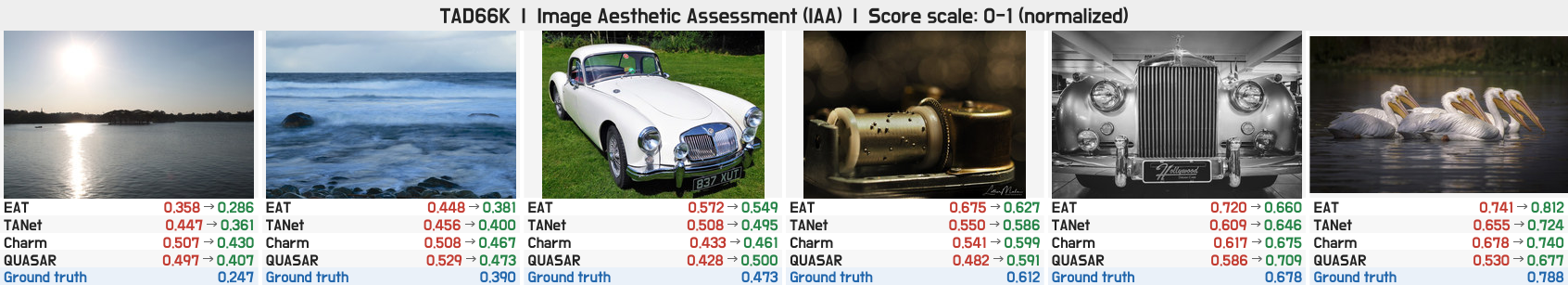}
\caption{\textbf{Qualitative score refinement on TAD66K.} All scores are displayed on the normalized 0--1 scale. Charm's original and refined scores are both divided by 10 to match this scale. Colors and notation follow \Cref{fig:qualitative-ava}.}
\label{fig:qualitative-tad66k}
\end{figure*}

\begin{figure*}[!tbp]
\centering
\includegraphics[page=6,width=\textwidth]{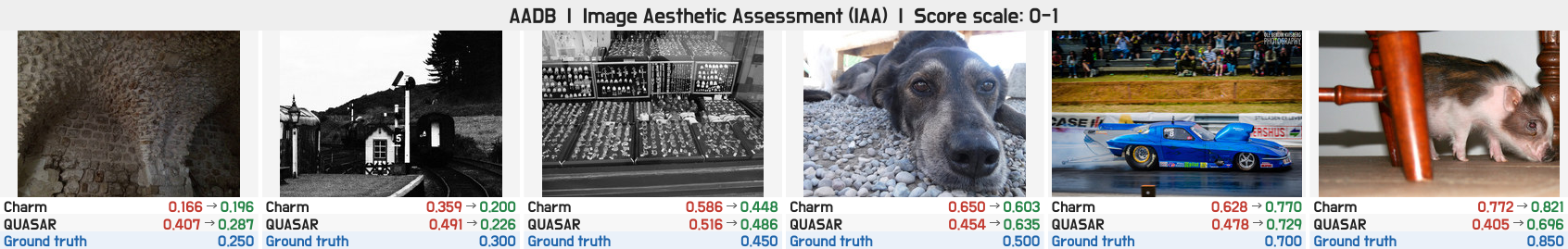}
\caption{\textbf{Qualitative score refinement on AADB.} Scores use the dataset's 0--1 scale. Colors and paired-score notation follow \Cref{fig:qualitative-ava}.}
\label{fig:qualitative-aadb}
\end{figure*}

\Cref{fig:qualitative-ava,fig:qualitative-koniq,fig:qualitative-spaq,fig:qualitative-kadid,fig:qualitative-tad66k,fig:qualitative-aadb} visualize six illustrative image-level comparisons per dataset on AVA, KonIQ-10k, SPAQ, KADID-10k, TAD66K, and AADB. Each method is paired with its own three-head \method refinement on the same image. Red and green values denote the original and refined scores, respectively, and blue values denote ground truth. The dataset, assessment task, and score scale are indicated above each grid. These qualitative examples complement the complete quantitative results in the main paper.

\end{document}